\documentclass[10pt,twocolumn,letterpaper]{article}

\usepackage{Styles/cvpr}              

\usepackage{graphicx}
\usepackage{amsmath}
\usepackage{amssymb}
\usepackage{booktabs}
\usepackage{graphicx}
\usepackage{url}
\usepackage{multirow}
\usepackage{xcolor}
\usepackage[accsupp]{axessibility}

\newcommand{\minus}{\scalebox{0.75}[1.0]{$-$}}

\usepackage[pagebackref,breaklinks,colorlinks]{hyperref}

\usepackage[capitalize]{cleveref}
\crefname{section}{Sec.}{Secs.}
\Crefname{section}{Section}{Sections}
\Crefname{table}{Table}{Tables}
\crefname{table}{Tab.}{Tabs.}

\def\cvprPaperID{5870} 
\def\confName{CVPR}
\def\confYear{2023}

\begin{document}

\title{Progressive Transformation Learning for Leveraging Virtual Images in Training}

\author{%
  Yi-Ting Shen$^{\star,1}$~~~~~~~~~~Hyungtae Lee$^{\star,2}$~~~~~~~~~~Heesung Kwon$^{2}$~~~~~~~~~~Shuvra S. Bhattacharyya$^{1}$ \\\vspace{-1.1em}\\
  $^{\star}$ {\normalsize equal contribution}\\\vspace{-0.9em}\\
  $^{1}$ University of Maryland, College Park~~~~~~~~~~$^{2}$ DEVCOM Army Research Laboratory\\
  \small{Code: \url{https://gitlab.umiacs.umd.edu/dspcad/ptl-release}}
  }
\maketitle

\begin{abstract}

To effectively interrogate UAV-based images for detecting objects of interest, such as humans, it is essential to acquire large-scale UAV-based datasets that include human instances with various poses captured from widely varying viewing angles. As a viable alternative to laborious and costly data curation, we introduce Progressive Transformation Learning (PTL), which gradually augments a training dataset by adding transformed virtual images with enhanced realism. Generally, a virtual2real transformation generator in the conditional GAN framework suffers from quality degradation when a large domain gap exists between real and virtual images. To deal with the domain gap, PTL takes a novel approach that progressively iterates the following three steps: 1) select a subset from a pool of virtual images according to the domain gap, 2) transform the selected virtual images to enhance realism, and 3) add the transformed virtual images to the training set while removing them from the pool. In PTL, accurately quantifying the domain gap is critical. To do that, we theoretically demonstrate that the feature representation space of a given object detector can be modeled as a multivariate Gaussian distribution from which the Mahalanobis distance between a virtual object and the Gaussian distribution of each object category in the representation space can be readily computed. Experiments show that PTL results in a substantial performance increase over the baseline, especially in the small data and the cross-domain regime.

\end{abstract}

\section{Introduction}
\label{sec:intro}

Training an object detector usually requires a large-scale training image set so that the detector can acquire the ability to detect objects' diverse appearances. This desire for a large-scale training set is bound to be greater for object categories with more diverse appearances, such as the human category whose appearances vary greatly depending on its pose or viewing angles. Moreover, a person's appearance becomes more varied in images captured by an unmanned aerial vehicle (UAV), leading to a wide variety of camera viewing angles compared to ground-based cameras, making the desire for a large-scale training set even greater. In this paper, we aim to satisfy this desire, especially when the availability of UAV-based images to train a human detector is scarce, where this desire is more pressing. 

\begin{figure}[t]
\centering
\includegraphics[trim=5mm 5mm 5mm 5mm,clip,width=.8\linewidth]{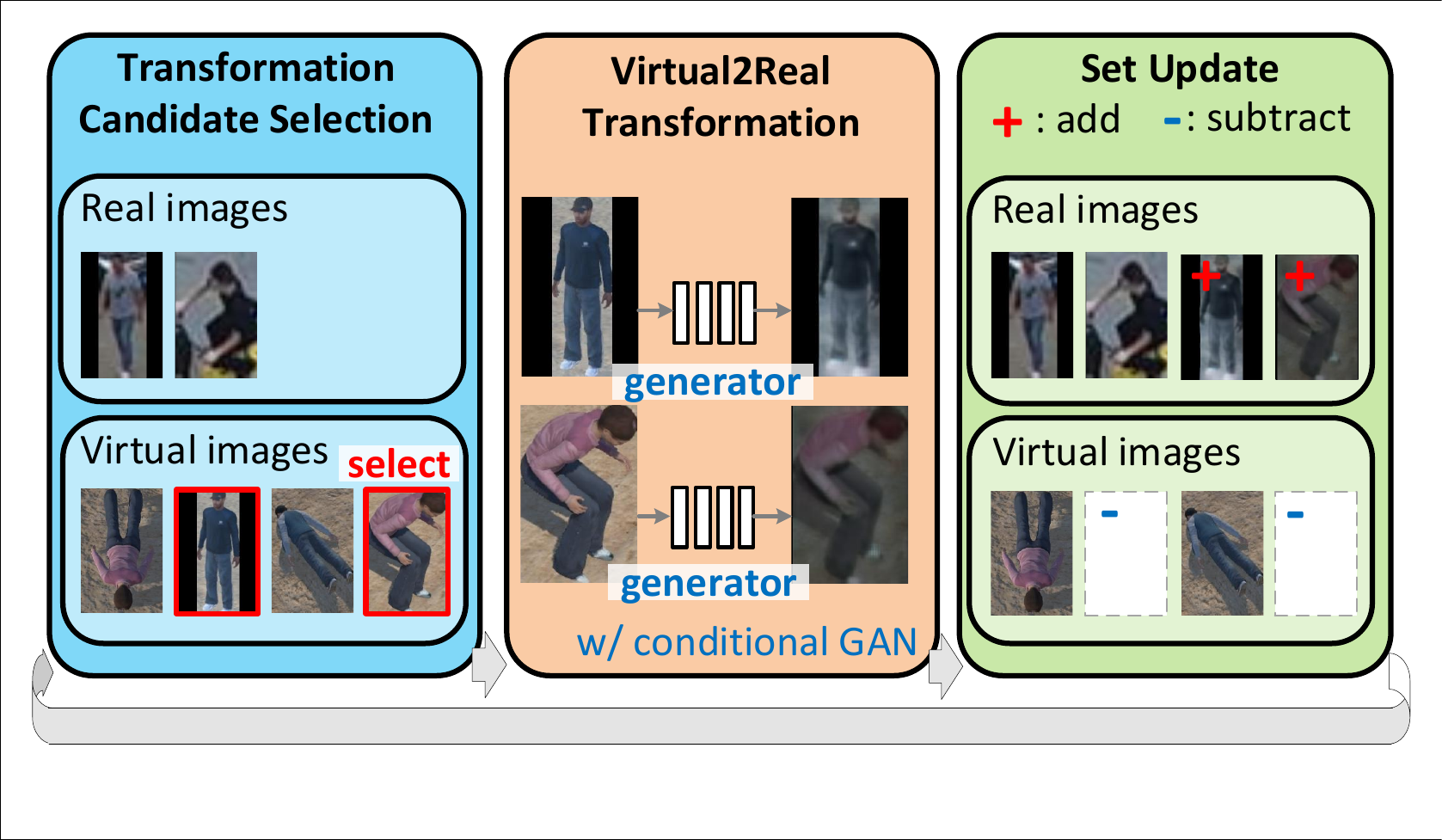}
\vspace{-.6cm}
\caption{{\bf Overview of Progressive Transformation Learning.}}
\label{fig:overview}
\end{figure}

As an intuitive way to expand the training set, one might consider synthesizing virtual images to imitate real-world images by controlling the optical and physical conditions in a virtual environment. Virtual images are particularly useful for UAV-based object detection since abundant object instances can be rendered with varying UAV locations and camera viewing angles along with ground-truth information (e.g., bounding boxes, segmentation masks) that comes free of charge. Therefore, a large-scale virtual image set covering diverse appearances of human subjects that are rarely shown in existing UAV-based object detection benchmarks~\cite{PZhuTPAMI2021,MBerkatainCVPRW2017,ICGlink} can be conveniently acquired by controlling entities and parameters in a virtual environment, such as poses, camera viewing angles, and illumination conditions.

To make virtual images usable for training real-world object detection models, recent works~\cite{JHoffmanICML2018,JMuCVPR2020,CLiCVPR2021,HQiuICCV2021} transform virtual images to look realistic. They commonly use the virtual2real generator~\cite{MMirzaArXiv2014} trained with the conditional GAN framework to transform images in the source domain to have the visual properties of images in the target domain. Here, virtual and real images are treated as the source and target domains, respectively. However, the large discrepancy in visual appearance between the two domains, referred to as the ``domain gap'', result in the degraded transformation quality of the generator. In fact, the aforementioned works using virtual images validate their methods where the domain gap is not large (e.g., digit detection~\cite{JHoffmanICML2018}) or when additional information is available (e.g., animal pose estimation with additional keypoint information~\cite{JMuCVPR2020,CLiCVPR2021,HQiuICCV2021}). In our case, real and virtual humans in UAV-based images inevitably have a large domain gap due to the wide variety of human appearances. 

To address the large domain gap, one critical question inherent in our task is \emph{how to measure accurately the domain gap.} Consequently, we estimate the domain gap in the representation space of a human detector trained on the real images. The representation space of the detector is learned such that test samples, which have significantly different properties than training samples from the perspective of the detector, are located away from the training samples. In this paper, we show that the feature distribution of object entities belonging to a certain category, such as the human category, in the representation space can be modeled with a multivariate Gaussian distribution if the following two conditions are met: i) the detector uses the sigmoid function to normalize the final output and ii) the representation space is constructed using the output of the penultimate layer of the detector. This idea is inspired by~\cite{KLeeNeurIPS2018}, which shows that softmax-based classifiers can be modeled as multivariate Gaussian distributions. In this paper, we show that the proposition is also applicable to sigmoid-based classifiers, which are widely used by object detectors. Based on this modeling, when the two aforementioned conditions are met, the human category in the representation space can be represented by two parameters (i.e., mean and covariance) of a multivariate Gaussian distribution that can be computed on the training images. With the empirically calculated mean and covariance, the domain gap from a single virtual human image to real human images (i.e., the training set) can be measured using the Mahalanobis distance~\cite{PMahalanobisNISI1936}.

To add virtual images to the training set to include more diverse appearances of objects while preventing the transformation quality degradation caused by large domain gaps, we introduce \emph{Progressive Transformation Learning (PTL)} (Figure~\ref{fig:overview}). PTL progressively expands the training set by adding virtual images through iterating the three steps: 1) transformation candidate selection, 2) virtual2real transformation, and 3) set update. When selecting transformation candidates from a virtual image pool, we use weighted random sampling, which gives higher weights to images with smaller domain gaps. The weight takes an exponential term with one hyperparameter controlling the ratio between images with smaller domain gaps and images with more diverse appearances. Then, the virtual2real transformation generator is trained via the conditional GAN, taking the selected transformation candidates as the ``source'' and the images in the training set as the ``target''. After transforming the transformation candidates by applying the virtual2real transformation generator, the training set is expanded with the transformed candidates while the original candidates are excluded from the pool of virtual images.

The main contribution of this paper is that we have validated the utility of virtual images in augmenting training data via PTL coupled with carefully designed comprehensive experiments. We first use the task of low-shot learning, where adequately expanding datasets has notable effects. Specifically, PTL provides better accuracy on three UAV-view human detection benchmarks than other previous works that leverage virtual images in training, as well as methods that only use real images. Then, we validate PTL on the cross-domain detection task where training and test sets are from distinct domains and virtual images can serve as a bridge between these two sets. 
The experimental results indicate that a high-performance human detection model can be effectively learned via PTL, even with the significant lack of real-world training data.

\section{Related Works}
\label{sec:related_works}

\noindent{\bf Leveraging virtual images in training.} In this section, we have listed previous works that demonstrate how virtual images can play a role in a variety of real-world applications when used for training. In fact, virtual images are desirable for model training as large-scale labeled datasets can be built virtually almost free of charge. Unfortunately, when virtual images are used without proper care, it is shown that the performance improvement is limited due to the domain gap between the virtual images and the real test images. Generally, previous works leveraging virtual images during model training can be summarized into three approaches according to how they exploit the advantages of virtual images and address the challenges of using virtual images.

The most intuitive and widely used approach of using virtual images is to pre-train a model on virtual images and fine-tune the pre-trained model on real images acquired in the same domain as the test images~\cite{AHandaCVPR2016,DKimCVPR2016,AGaidonCVPR2016,MFabbriECCV2018,GVarolCVPR2018,JMuCVPR2020,SMashraCVPR2022,XGuoCVPR2022,ZJinCVPR2022,KBaekCVPR2022}. This approach aims to avoid the domain gap by fine-tuning the model on real images acquired under the same conditions and environments of test images. While the first approach seeks to use the representative capability learned from \emph{large-scale} virtual datasets, the second approach seeks to exploit additional information that can be easily \emph{labeled} on virtual images. For example, \cite{QLiuCVPR2022} annotates part segmentation maps when acquiring virtual vehicle images, and uses the part segmentation results from the pre-trained model on the virtual images during the fine-tuning process. Similarly, \cite{QYanCVPR2022} uses depth and semantic information labeled when acquiring virtual images. 

The third approach directly builds training batches consisting of both virtual images and real images. \cite{GRosCVPR2016} and \cite{SRichterECCV2016} adopt the most naive approach to build a batch by randomly selecting a fixed number of images from each of the real and virtual image sets. However, even if the number of virtual images is several orders of magnitude greater than the number of real training images, this approach does not provide remarkably better accuracy and may even provide worse accuracy than its counterparts using only real images for training. In this case, the effect of using virtual images during model training does not appear as expected because the large domain gap between the real (test) images and the virtual images is not adequately addressed. In this paper, we also use virtual images directly for model training while considering appropriately reducing this domain gap.\smallskip

\noindent{\bf Progressive learning.} Progressive learning is a machine learning strategy that continuously trains a model from easy to hard tasks, primarily for the purpose of training stabilization or fast optimization. One of the most common applications for progressive learning is incrementally increasing the network capacity to improve network capability. The most intuitive approach in this category is to gradually increase the network size (e.g., depth or width) to ease the training difficulty of very deep networks~\cite{SFahlmanNeurIPS1989,YBengioNeurIPS2006,LSmithCVPR2016,TChenICLR2016,TWeiICML2016,LGongICML2019,CLiCVPR2022}. Conversely, \cite{MZhangNeurIPS2020} uses progressive learning in the direction of reducing the network size for fast training. Progressive learning is also used in GAN frameworks to enhance the generator's ability to transform input images of larger resolution~\cite{TKarrasICLR2018}. Curriculum learning~\cite{YBengioICML2009,VSpitkovskyNeurIPSW2009,FKhanNeurIPS2011,AGravesICML2017}, which  continuously raises the level of training from easy to difficult samples, also falls into this category.

Progressive learning is also used to deal with the scalability of datasets that are incompletely labeled. In a semi-supervised learning task, \cite{GWangICCV2017,CWangICLR2022,TCaoECCV2022} adopt progressive learning by gradually increasing the number of unlabeled data used for training. Self-learning in~\cite{MKumarNeurIPS2010,PKaulCVPR2022} uses progressive learning by repeating the two steps, assigning labels depending on the current detector and updating the current detector with these labels.

Our method can be seen as similar to the second approach in that it also intend to expand the training dataset. However, our method uses progressive learning to reliably add realistically transformed virtual images to the training dataset by avoiding quality degradation of the transformation, which has never been attempted before using the progressive learning strategy.

\section{Method}
\label{sec:method}

\subsection{Measuring the Domain Gap between Real and Virtual Images}
\label{ssec:domain_gap_measurement}

\noindent{\bf Modeling training set with multivariate Gaussian distribution.} The purpose of adopting progressive transformation learning, which progressively expands the training set with a subset of the realistically transformed virtual images instead of expanding it with the full set at once, is to avoid the significant domain gap between the real and virtual images when training the transformation generator. Here, the domain gap is measured in the representation space of the detector, which is learned so that two samples with different properties from the perspective of the detector are separated far from each other. 

\begin{figure*}[t]
\centering
\includegraphics[trim=5mm 5mm 5mm 5mm,clip,width=0.85\linewidth]{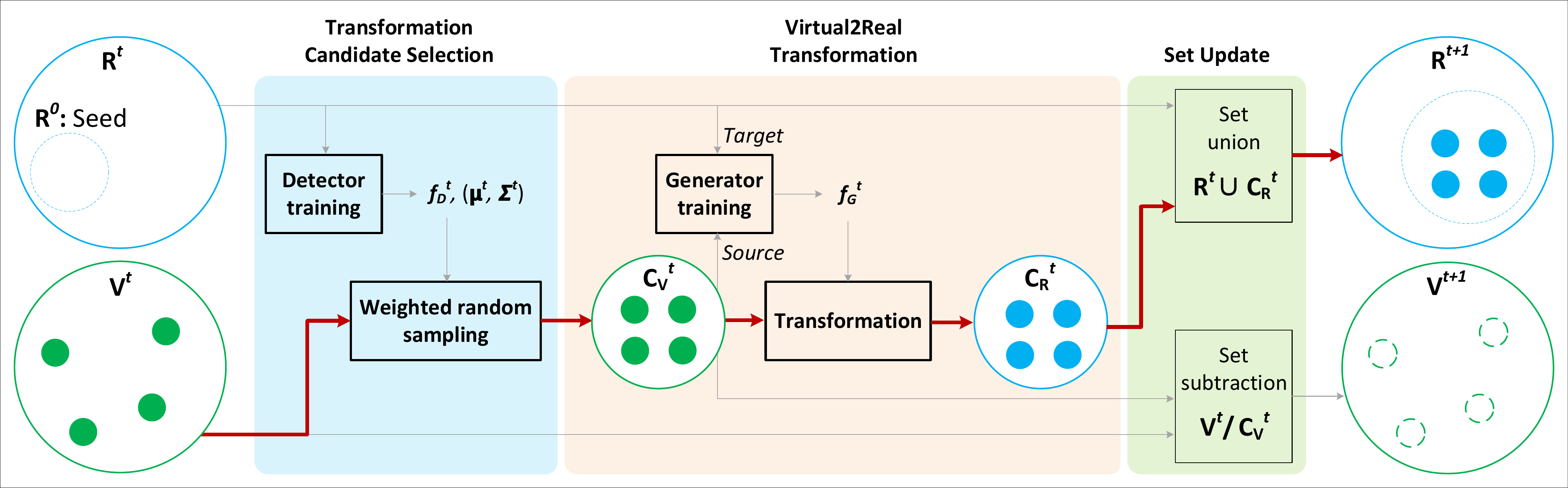}
\vspace{-0.2cm}
\caption{{\bf Progressive Transformation Learning (PTL) pipeline.} The red arrow indicates the processing flow of the virtual images selected to be added to the training set.} 
\label{fig:flowchart}
\end{figure*}

In general, the representation space of the detector refers to the space formed by the output of the penultimate layer of the detector since all layers of the detector except for the last layer can be transferred for different downstream tasks~\cite{KHeCVPR2020,TChenICML2020,JGrillNeurIPS2020}. In~\cite{KLeeNeurIPS2018}, it is shown that for the softmax-based classifier, the distribution of each category in the representation space can be modeled as a multivariate Gaussian distribution. Object detector generally uses the sigmoid function (i.e., $f_{\mathit{sigmoid}}({\bf x}) = 1/(1+\exp(\minus {\bf w}_c^T{\bf x}\minus b_c)$ for the category $c$), which does not consider outputs for other categories, instead of the softmax function (i.e., $f_{\mathit{softmax}}({\bf x})=\exp({\bf w}_c^T{\bf x}+ b_c)/\sum_{c'}{\exp({\bf w}_{c'}^T{\bf x}+ b_{c'})}$ for the category $c$) that competes for outputs for all categories to normalize the model output to [0 1]. This is because, unlike classification, the detection task must take into account that two or more co-located objects may be active on a single output. In the supplementary material, we show that even for the sigmoid-based detector, the distribution of each category in the representation space can also be modeled as a multivariate Gaussian distribution. 

Specifically, let ${\bf x}\in\mathcal{X}$ and $y=\{y_c\}_{c=1,\cdots,C}\in\mathcal{Y}, y_c\in\{0,1\}$ be an input and its label, respectively. Then, the representation space of the sigmoid-based detector can be expressed as follows:
\begin{equation}
    P(f({\bf x})|y_c=1)\sim\mathcal{N}(f({\bf x})|{\bf \mu}_c,{\bf \Sigma}_c),
\end{equation}
where $f(\cdot)$ denotes the output of the penultimate layer of the detector. $\mu_c$ and $\Sigma_c$ are the mean and the covariance of the multivariate Gaussian distribution for the category $c$. ${\bf \mu}_c$ and ${\bf \Sigma}_c$ can be calculated over the entire set of training images as follows:
\begin{eqnarray}
    {\bf \mu}_c &=& \frac{1}{|\text{D}_c|}\sum_{{\bf x}\in\text{D}_c}{f({\bf x})},\nonumber\\
    {\bf \Sigma}_c&=&\frac{1}{|\text{D}_c|}\sum_{{\bf x}\in\text{D}_c}{\left(f({\bf x})-{\bf \mu}_c\right)\left(f({\bf x})-{\bf \mu}_c\right)^\top},~\label{eq:gaussian_params}
\end{eqnarray}
where $\text{D}_c$ is the set of instances for the category $c$. Practically, any detection whose IoU with the groundtruth of the category $c$ is greater than 0.5 belongs to $\text{D}_c$.\smallskip

\noindent{\bf Measuring domain gap.} After ${\bf \mu}_c$ and ${\bf \Sigma}_c$ are empirically calculated to represent $\text{D}_c$, the domain gap between a new image ${\bf x}_{\mathit{new}}$ and $\text{D}_c$ can be measured using the Mahalanobis distance, as follows:
\begin{equation}
    d({\bf x}_{\mathit{new}}) = \left(f({\bf x}_{\mathit{new}})-{\bf \mu}_c\right)^\top{\bf \Sigma}_c^{\minus 1}\left(f({\bf x}_{\mathit{new}})-{\bf \mu}_c\right).
\end{equation}

This measurement of the domain gap is highly dependent on the detector's ability to detect objects in the image. It is commonly known that the detection capability of a detector is greatly affected by the image size as well as the object appearance in the image. To mitigate the effect of image size on measuring the domain gap, the Mahalanobis distance for ${\bf x}_{new}$ is calculated at multiple image scales, and the minimum distance is used as the domain gap, as follows:
\begin{equation}
    d({\bf x}_{\mathit{new}}) = \min_{s\in S}(\{d({\bf x}^s_{\mathit{new}})\}),\label{eq:mahalanobis}
\end{equation}
where ${\bf x}^s_{\mathit{new}}$ is the resized image of ${\bf x}_{\mathit{new}}$ to be $s$$\times$$s$. $S$ is the set of resizing factors. In our experiments, we use $S=\{128, 256, 384, 512\}$.

\subsection{Progressive Transformation Learning}
\label{ssec:PTL}

Our objective is to expand the training set consisting of real images by adding virtual images which are transformed to intimate real images. The virtual2real transformation can be performed by a generator trained by treating virtual images and real images as ``source'' and ``target'', respectively, in the conditional GAN framework. Inevitably, the transformation quality of the trained generator is degraded when the domain gap between the source domain and the target domain is large. To prevent the degraded transformation quality due to the large domain gap, we introduce \emph{Progressive Transformation Learning (PTL)}, which progressively and iteratively expands the training set with a subset of virtual images carefully selected to avoid the large domain gap. PTL goes through three steps for each iteration (Fig~\ref{fig:flowchart}): i) sampling a subset of virtual images from a virtual image pool by giving heavier weights to images close to the current training set (Transformation candidate selection), ii) transforming the selected images to be realistic (Virtual2real transformation), and iii) adding the transformed images to the training set while excluding the selected images from the virtual image pool (Set update). 
The details of each step are described next.\smallskip

\noindent{\bf Transformation candidate selection.} When selecting transformation candidates, we must consider two conflicting claims simultaneously: i) to suit the purpose of PTL, virtual images with a small domain gap should be selected, but ii) to suit the purpose of expanding the training set, virtual images with appearances that rarely appear in the training set, which usually implies a large domain gap, should also be selected.

To jointly consider these two claims, we use weighted random sampling. The sampling weight takes the exponential term which gives higher weights to virtual images with smaller domain gaps, while introducing one hyper-parameter $\tau$ to control the amplitude of the weights, as follows:
\begin{equation}
    w({\bf x})=\exp{\left(\minus\frac{d({\bf x})}{\tau}\right)},\label{eq:weight}
\end{equation}
where $d({\bf x})$ is the Mahalanobis distance, which is used to measure the domain gap of ${\bf x}$ from the current training set (eq~\ref{eq:mahalanobis}). Intuitively, using a small $\tau$ allows a more frequent selection of images with smaller domain gaps by giving them larger weights than using a large $\tau$. (We use $\tau$=5.0 throughout all experiments.)

In practice, transformation candidates are selected from virtual image pool through the following four steps: i) training the human detector $f_{D}^t$ on the current training set of real images ${\bf \text{R}}^t$, ii) calculating ${\bf \mu}^t$ and ${\bf \Sigma}^t$ on ${\bf \text{R}}^t$ as in  eq.~\ref{eq:gaussian_params}, iii) calculating weights for each image in the current set of virtual images ${\bf \text{V}}^t$ as in eq.~\ref{eq:weight}, and iv) applying weighted random sampling to ${\bf \text{V}}^t$ to select a pre-defined number $n$ of transformation candidates. (We use $n$=100 throughout all experiments.)\smallskip

\noindent{\bf Virtual2Real transformation.} In line with the goal of this paper to obtain a human detector that can identify humans with diverse appearances captured by a UAV, we design the virtual2real transformation to focus on the person region rather than the background of the selected virtual images. To do so, we crop the person region in the virtual image, apply the transformation only to this region, and segment the transformed person back to the original background. For accurate segmentation, the pixel-wise segmentation mask is required. Obtaining such pixel-wise segmentation mask at no cost is another benefit of using virtual images.

The conditional GAN framework~\cite{MMirzaArXiv2014}, in which the generator is trained to transform a given input image from source styles into target styles, is widely used to transform virtual images to look like real images~\cite{PIsolaCVPR2017,CLiCVPR2021}. Among many variants of conditional GANs, we use CycleGAN~\cite{JZhuICCV2017} in which the generator is trained to minimize the reconstruction error between the input image and the reconstructed image transformed back to the original style of the input image after the initial transformation to the target style. It is shown in~\cite{MMirzaArXiv2014} that the transformation with CycleGAN is likely to maintain the original object pose while changing detailed styles such as patterns (e.g., transforming a white horse into a zebra in the same pose). We intend to borrow this characteristic of CycleGAN to transform virtual images in the direction that makes the detailed styles realistic while maintaining the overall human appearances, which depend on various viewing angles or human poses.

In practice, the virtual2real transformation generator $f_{G}^t$ is trained using the CycleGAN framework by treating the selected transformation candidates ${\bf \text{C}_\text{V}}^t$ and the current set of real images ${\bf \text{R}}^t$ as ``source'' and ``target'', respectively. Then, ${\bf \text{C}_\text{V}}^t$ are transformed to realistic transformed images ${\bf \text{C}_\text{R}}^t$ by applying the virtual2real transformation generator.\smallskip

\noindent{\bf Set update.} After the transformed images ${\bf \text{C}_{\text R}}^t$ are acquired from the selected transformation candidates ${\bf \text{C}_{\text V}}^t$, PTL updates the current real image set ${\bf \text{R}}^t$ and the current virtual image set ${\bf \text{V}}^t$ as follows:
\begin{equation}
{\bf \text{R}}^{t+1}={\bf \text{R}}^t\cup{\bf \text{C}_{\text R}}^t ~~~~~~ {\bf \text{V}}^{t+1}={\bf \text{V}}^t/{\bf \text{C}_{\text V}}^t
\end{equation} 
When this progressive learning is terminated, the final human detection model can be acquired by training on the final set of real images. 

In practice, the first two steps of PTL are applied to the tightly cropped image region of human region, but in the `set update' step, the entire image including the human region and the background is added to the training set for training the human detector. More precisely, when training the virtual2real transformation generator, the tightly cropped image region around each human from the training images are used as the ``target''.

\section{Experiments}
\label{sec:experiments}

\noindent{\bf Datasets and evaluation metrics.} We perform experiments on three real UAV-based datasets, VisDrone~\cite{PZhuTPAMI2021}, Okutama-Action~\cite{MBerkatainCVPRW2017}, and ICG~\cite{ICGlink}, and one virtual dataset, \textit{Archangel-Synthetic}~\cite{YShenArXiv2022}, all including human instances. \textit{Archangel-Synthetic} consists of various virtual characters with different poses across a range of altitudes and circle radii with different camera pitch angles (i.e., 17.3K images with eight characters, three poses, ten altitudes, six circle radii, and twelve camera pitch angles). Each image in \textit{Archangel-Synthetic} accompanies metadata about the above imaging conditions, allowing us to analyze how the feature distributions of virtual characters evolve with respect to these imaging conditions when PTL progresses. 
We use AP@.5 and AP@[.5:.95] as evaluation metrics for all experiments.\smallskip

\noindent{\bf Detector.} For the detector, we use RetinaNet~\cite{TLinICCV2017} that uses the feature pyramid network (FPN) to provide a rich multi-scale feature pyramid and processes features at all scale levels with the same subnetwork responsible for the final classification and bounding box regression. It is important to use the same subnetwork across all scale levels since the domain gap for each virtual image should be measured in a shared representation space regardless of the image size. Most other object detectors using FPN (e.g., SSD~\cite{WLiuECCV2016} and v4 or later versions of YOLO~\cite{ABochkovskiyArXiv2020,GJocher2020,CLiArXiv2022,CWangArXiv2022}) use different subnetworks at each scale level. However, PTL is not structurally limited to RetinaNet as it can be used with any detector with minor modifications, such as adding one shared layer across all scale levels. 

\subsection{Properties of Progressive Learning}
\label{ssec:progressive_learning}

\begin{figure*}[t]
\centering
\resizebox{0.81\linewidth}{!}{%
\setlength{\tabcolsep}{6.0pt}
\renewcommand{\arraystretch}{1.2}
\begin{tabular}{ccccc}
\scriptsize{{\bf iter. 1}} & \scriptsize{{\bf iter. 2}} & \scriptsize{{\bf iter. 3}} & \scriptsize{{\bf iter. 4}} & \scriptsize{{\bf iter. 5}}\\
\includegraphics[trim=0mm 0mm 0mm 0mm,clip,width=0.185\linewidth]{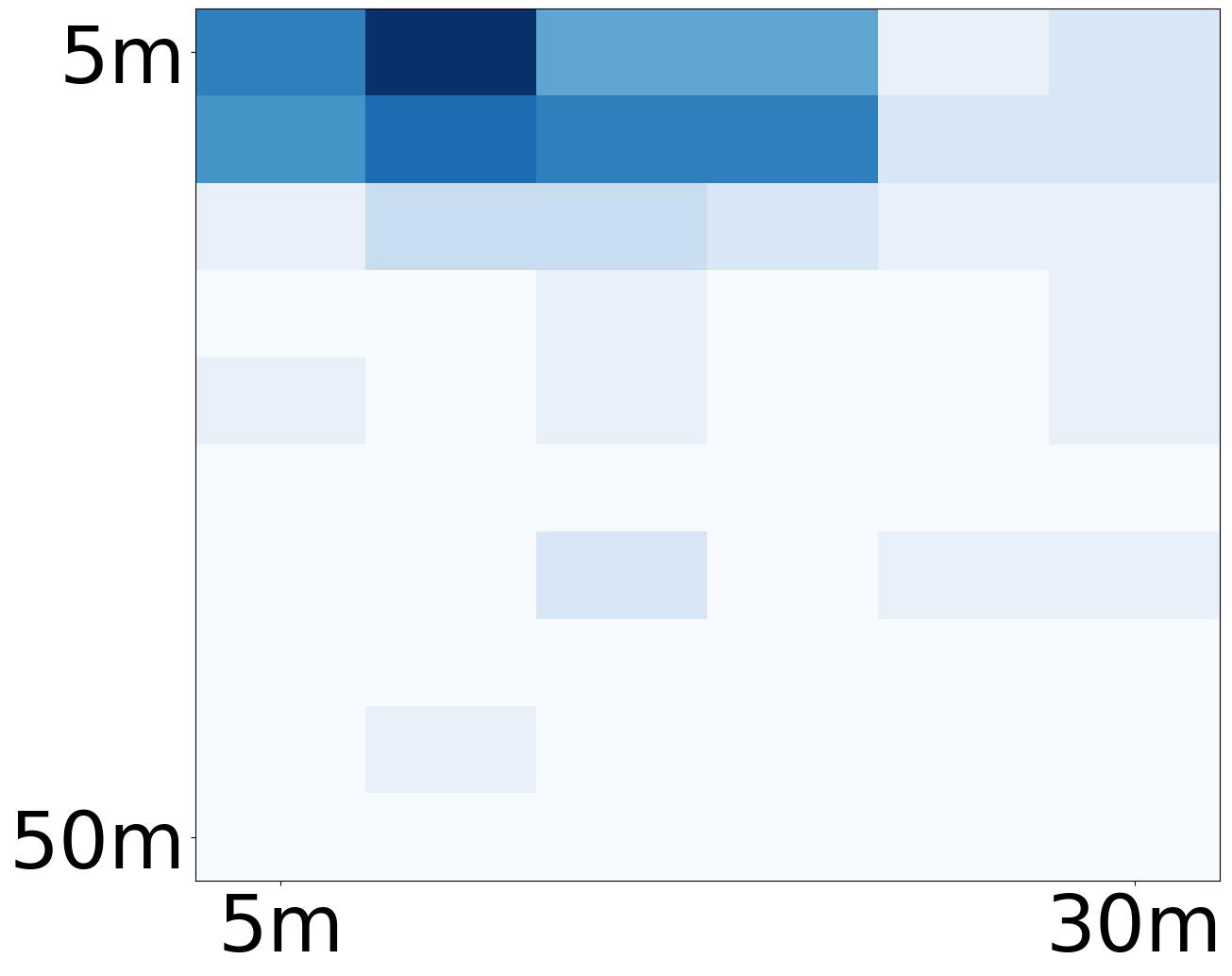} &
\includegraphics[trim=0mm 0mm 0mm 0mm,clip,width=0.185\linewidth]{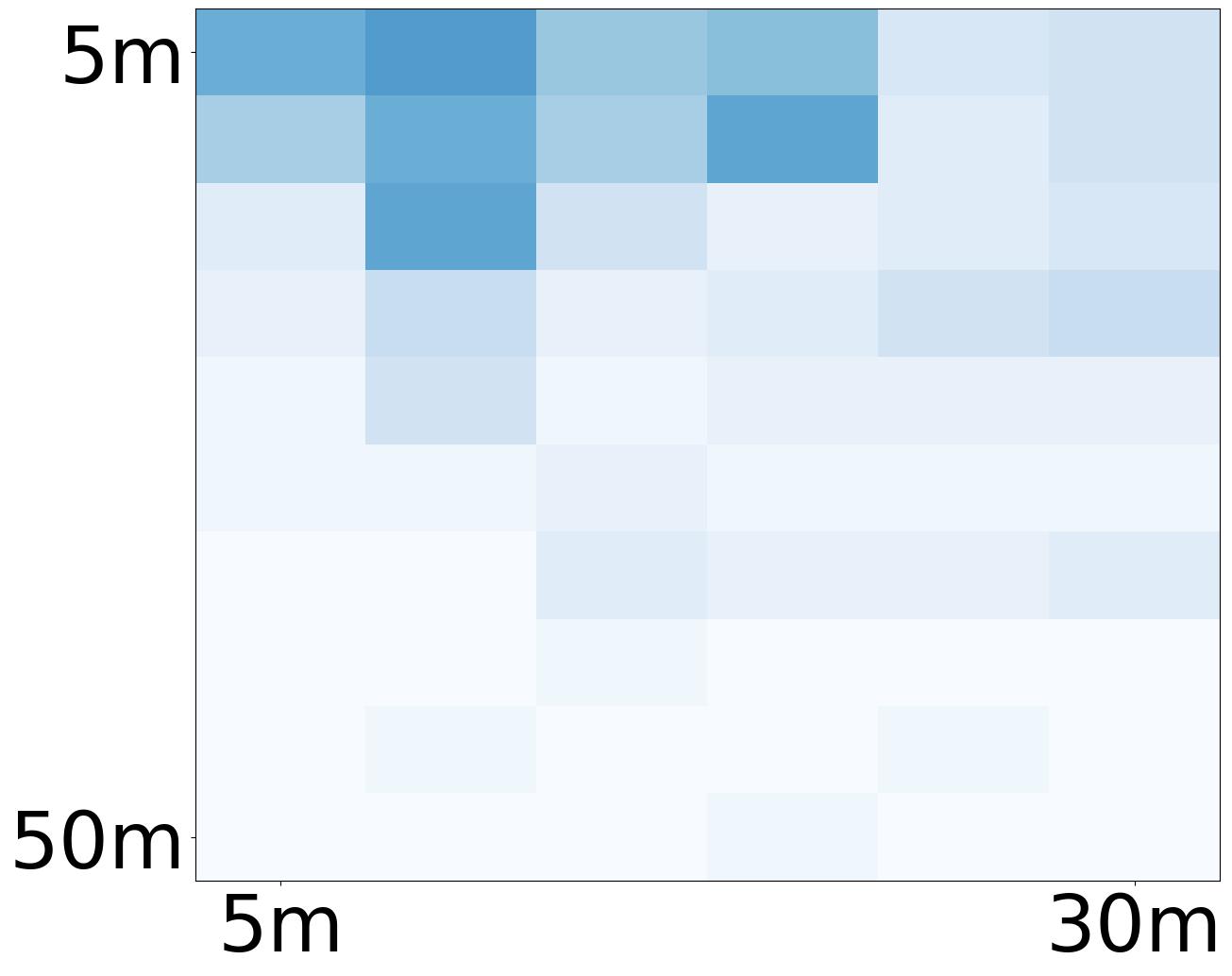} &
\includegraphics[trim=0mm 0mm 0mm 0mm,clip,width=0.185\linewidth]{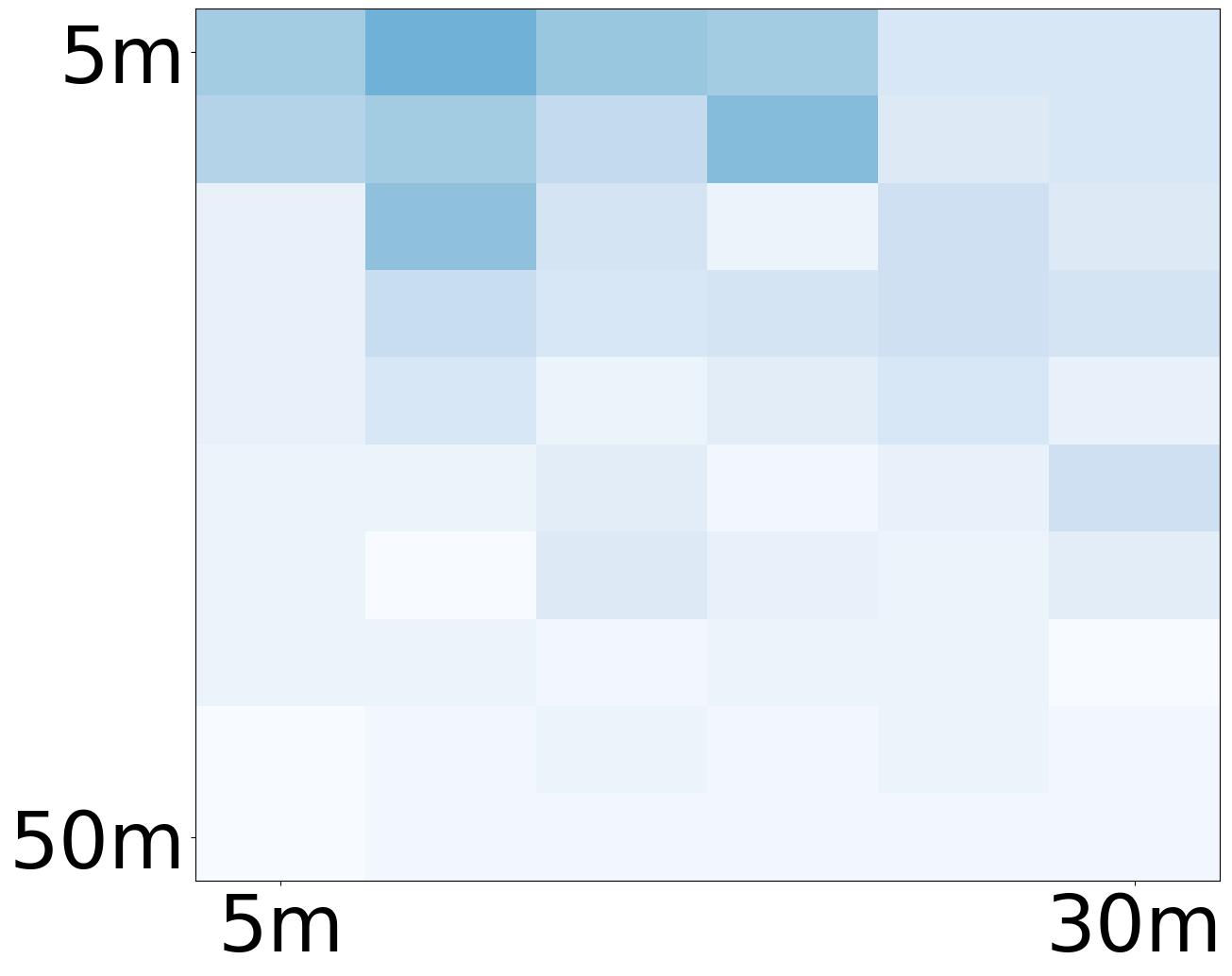} &
\includegraphics[trim=0mm 0mm 0mm 0mm,clip,width=0.185\linewidth]{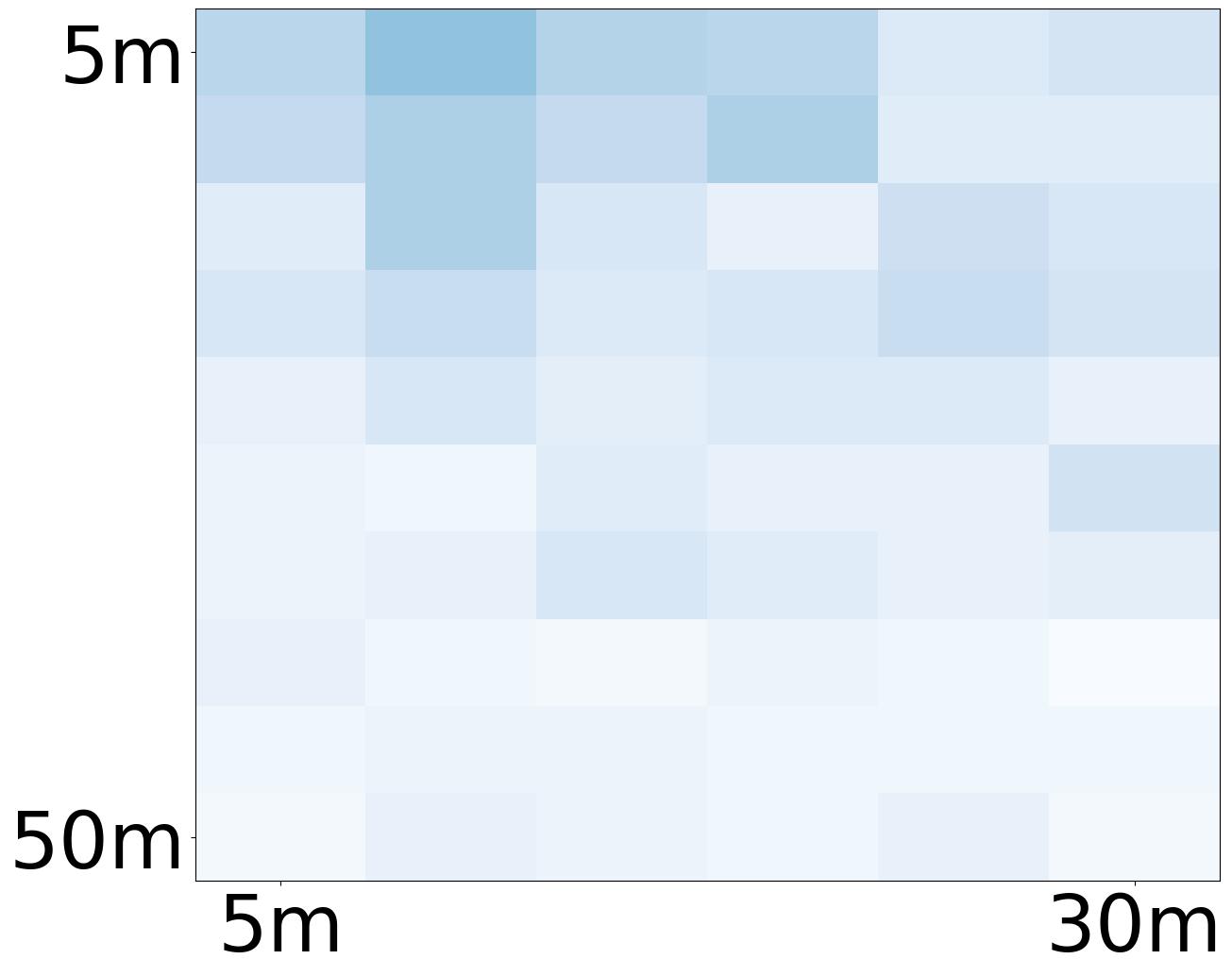} &
\includegraphics[trim=0mm 0mm 0mm 0mm,clip,width=0.185\linewidth]{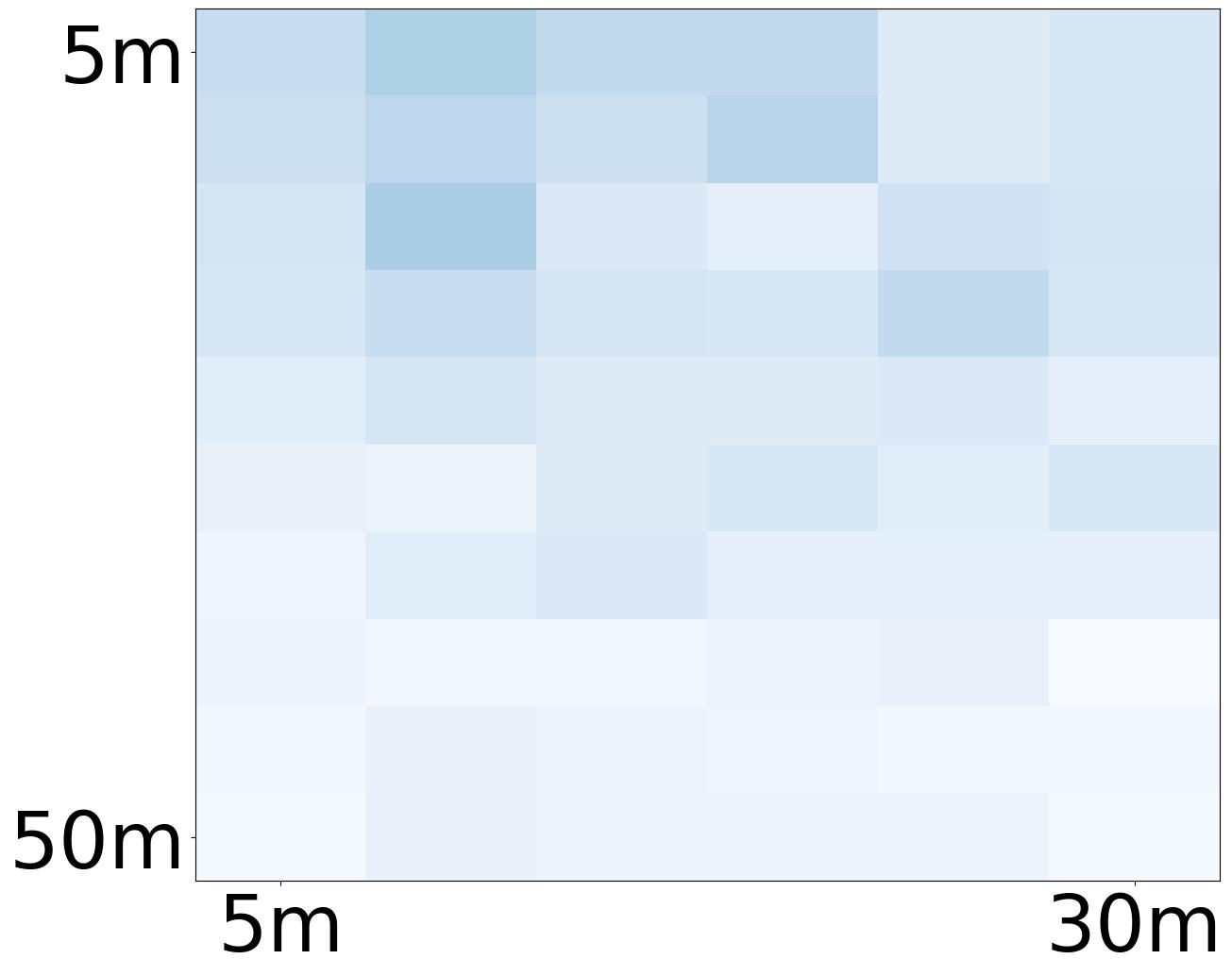}\\
\includegraphics[trim=0mm 0mm 0mm 0mm,clip,width=0.185\linewidth]{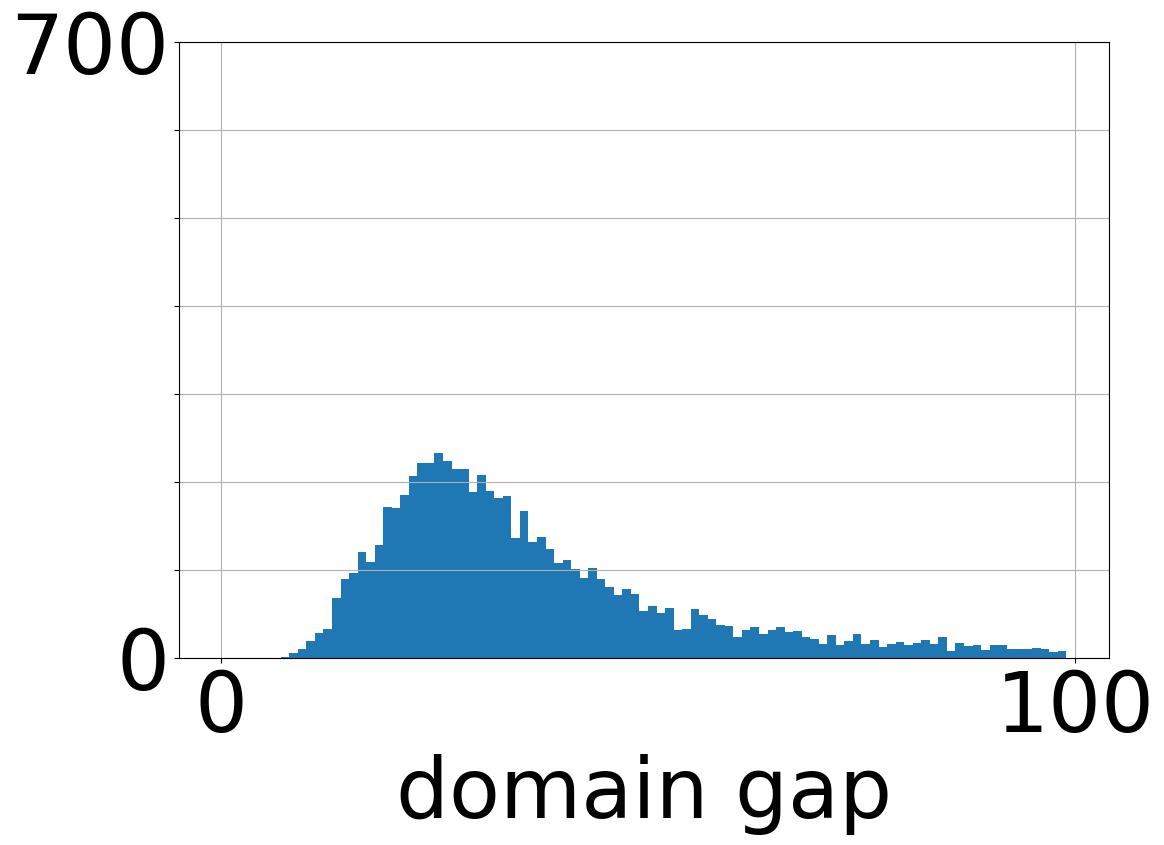} &
\includegraphics[trim=0mm 0mm 0mm 0mm,clip,width=0.185\linewidth]{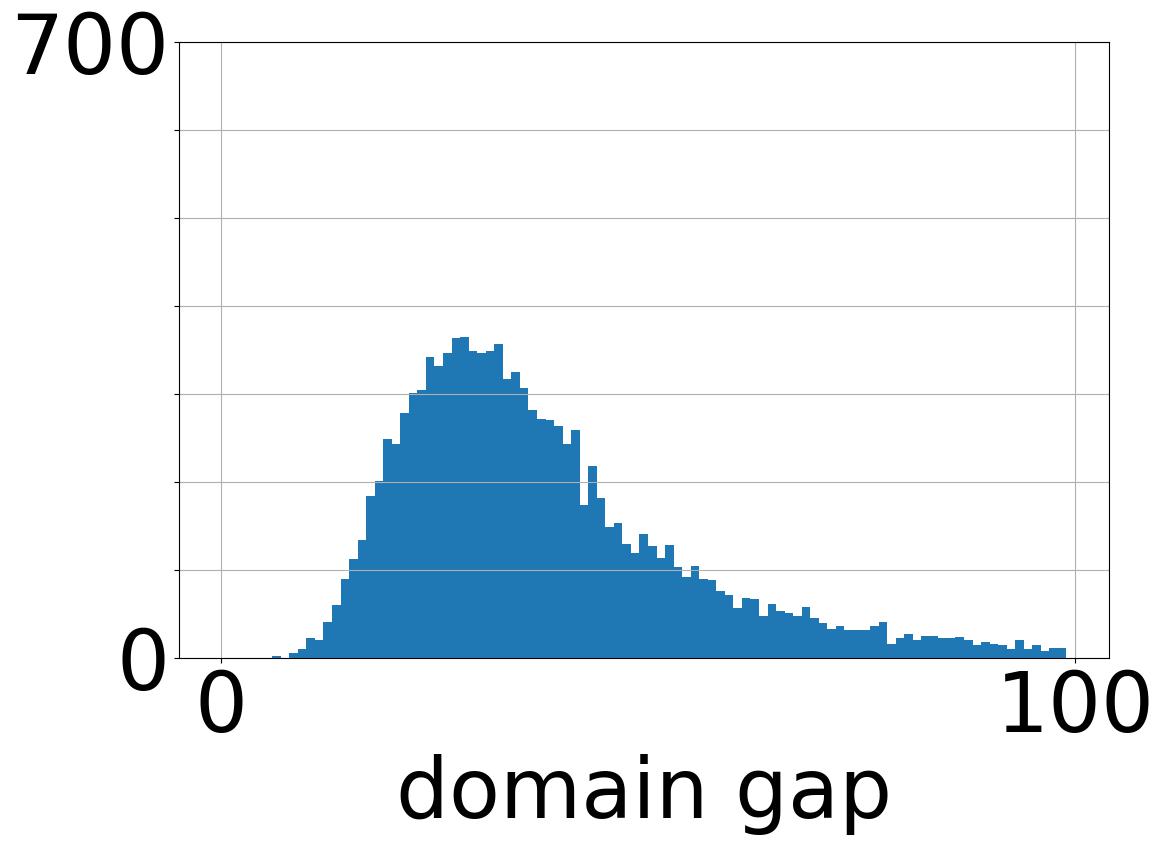} &
\includegraphics[trim=0mm 0mm 0mm 0mm,clip,width=0.185\linewidth]{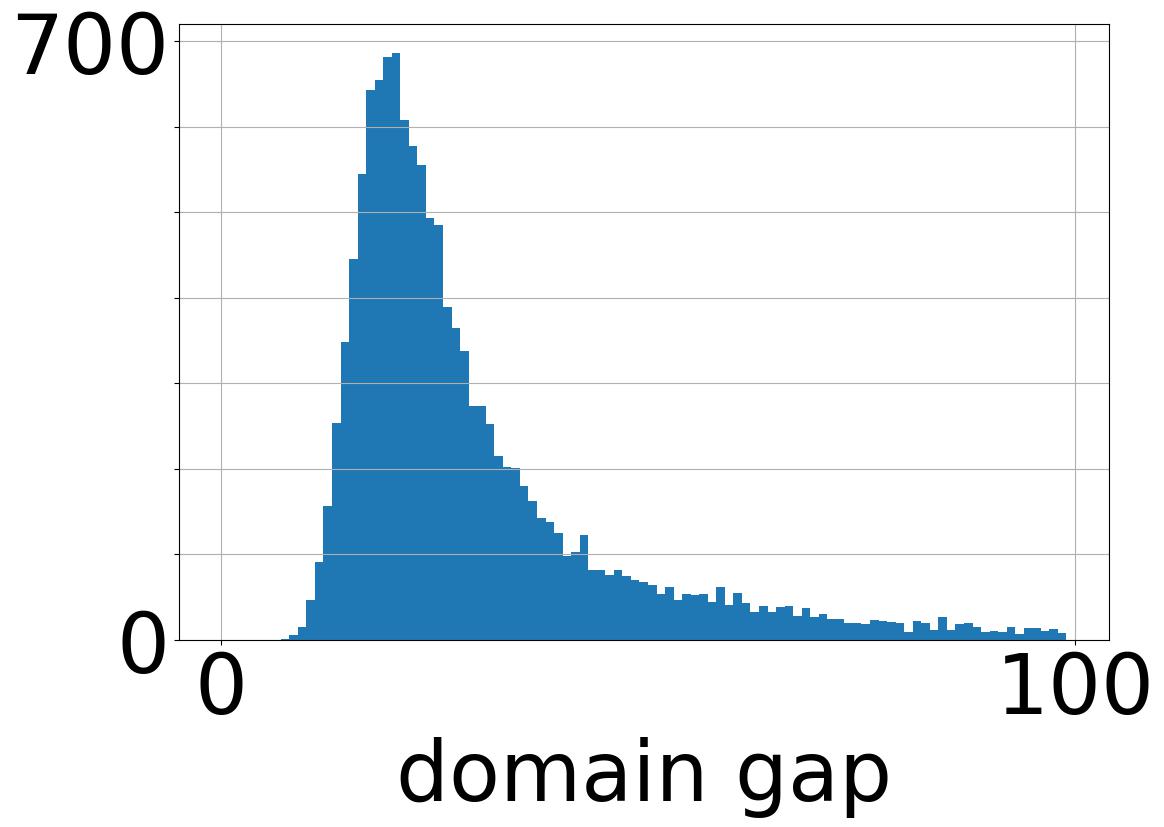} &
\includegraphics[trim=0mm 0mm 0mm 0mm,clip,width=0.185\linewidth]{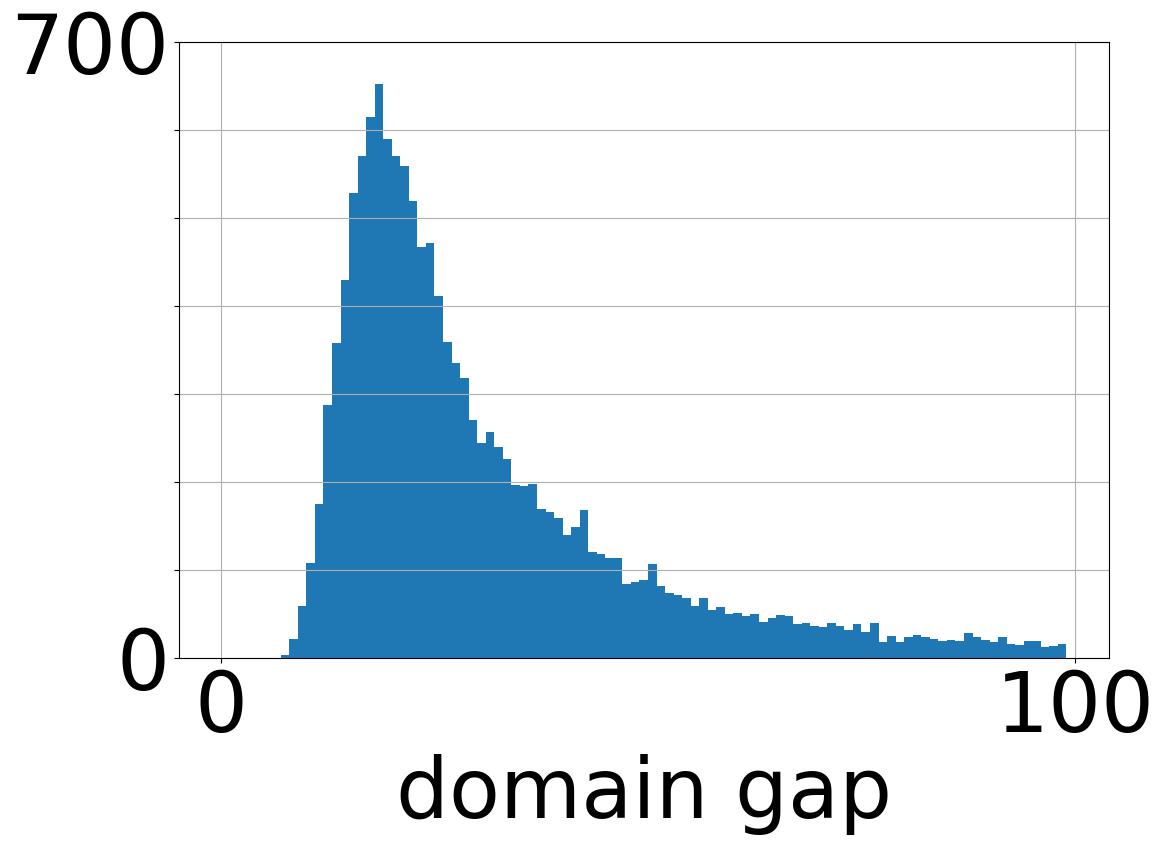} &
\includegraphics[trim=0mm 0mm 0mm 0mm,clip,width=0.185\linewidth]{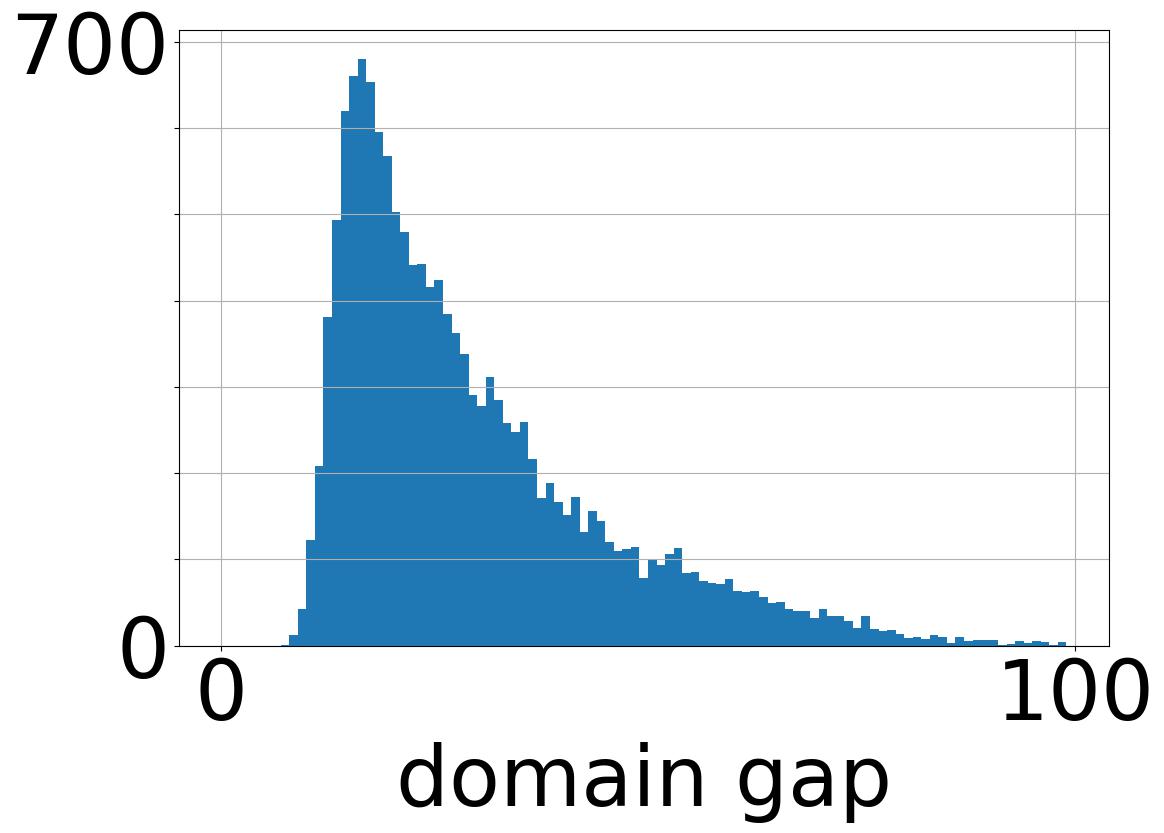}\\
\end{tabular}
}
\vspace{-0.4cm}
\caption{{\bf Analysis of the use of virtual images when PTL progresses.} The figures in the top row show the accumulated distribution of transformation candidates with respect to camera locations (i.e., altitude and rotation circle radius from the target human in $x$ and $y$ axes, respectively) for each PTL iteration. Darker bins indicate that more virtual images have been added to the training set. The figures in the bottom row ($x$ axis: domain gap, $y$ axis: the corresponding number of virtual images)  show the domain gap distribution of virtual images measured by eq.~\ref{eq:mahalanobis}. These figures are collected from the experimental setup of using 50 real images from the VisDrone dataset for training.}
\label{fig:analysis of virtualimg}
\end{figure*}

\begin{figure*}[t]
\centering
\resizebox{0.82\linewidth}{!}{%
\setlength{\tabcolsep}{5.0pt}
\begin{tabular}{cccccc}
{\bf AP@.5}: VisDrone & {\bf AP@[.5:.95]}: VisDrone & {\bf AP@.5}: Okutama-Action & {\bf AP@[.5:.95]}: Okutama-Action & {\bf AP@.5}: ICG & {\bf AP@[.5:.95]}: ICG \\
\includegraphics[trim=0mm 0mm 0mm 0mm,clip,width=0.32\linewidth]{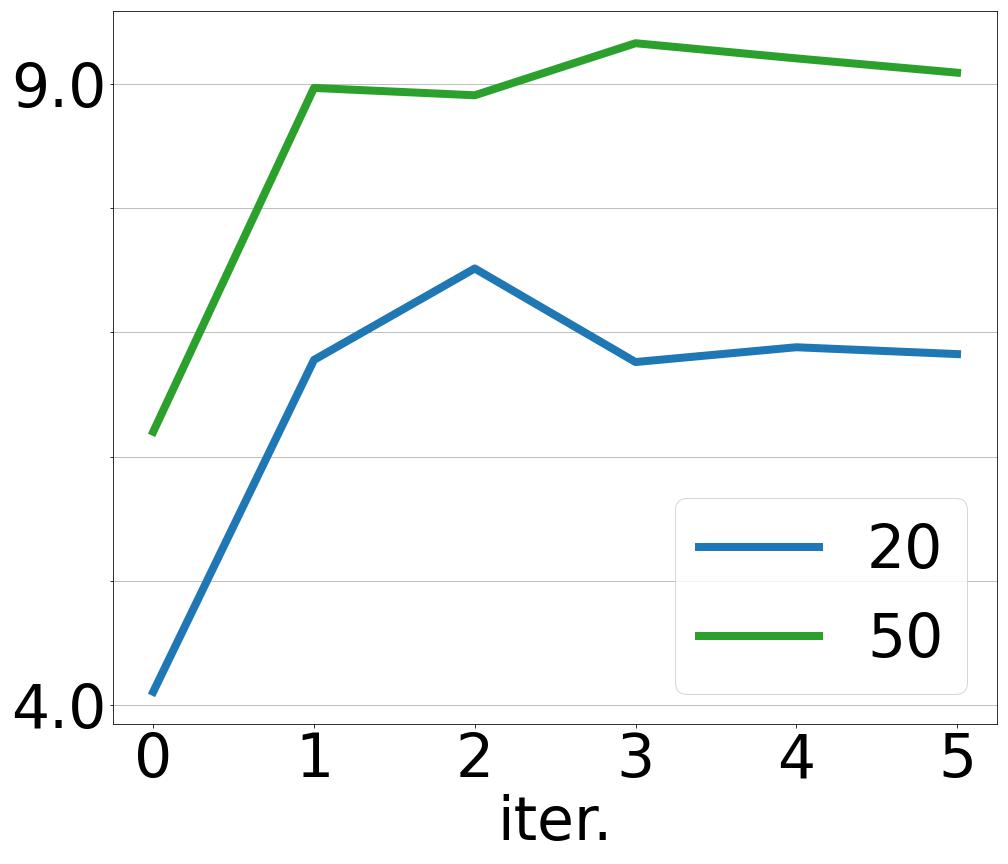} &
\includegraphics[trim=0mm 0mm 0mm 0mm,clip,width=0.32\linewidth]{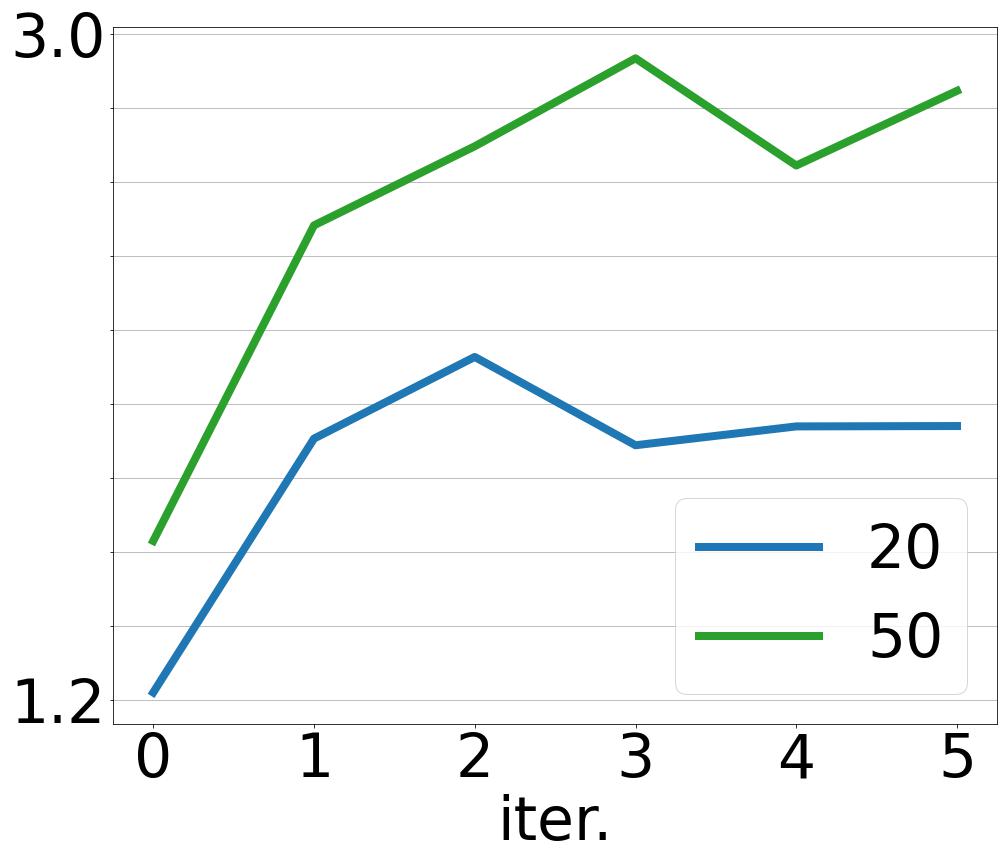} &
\includegraphics[trim=0mm 0mm 0mm 0mm,clip,width=0.32\linewidth]{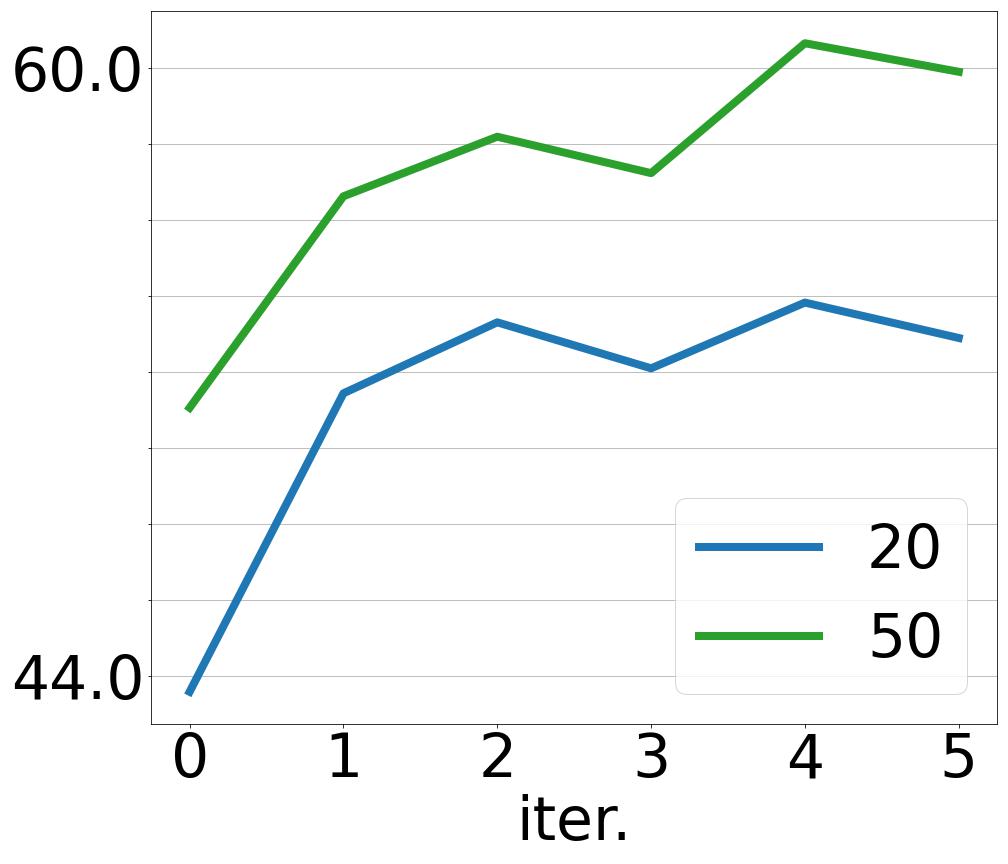} &
\includegraphics[trim=0mm 0mm 0mm 0mm,clip,width=0.32\linewidth]{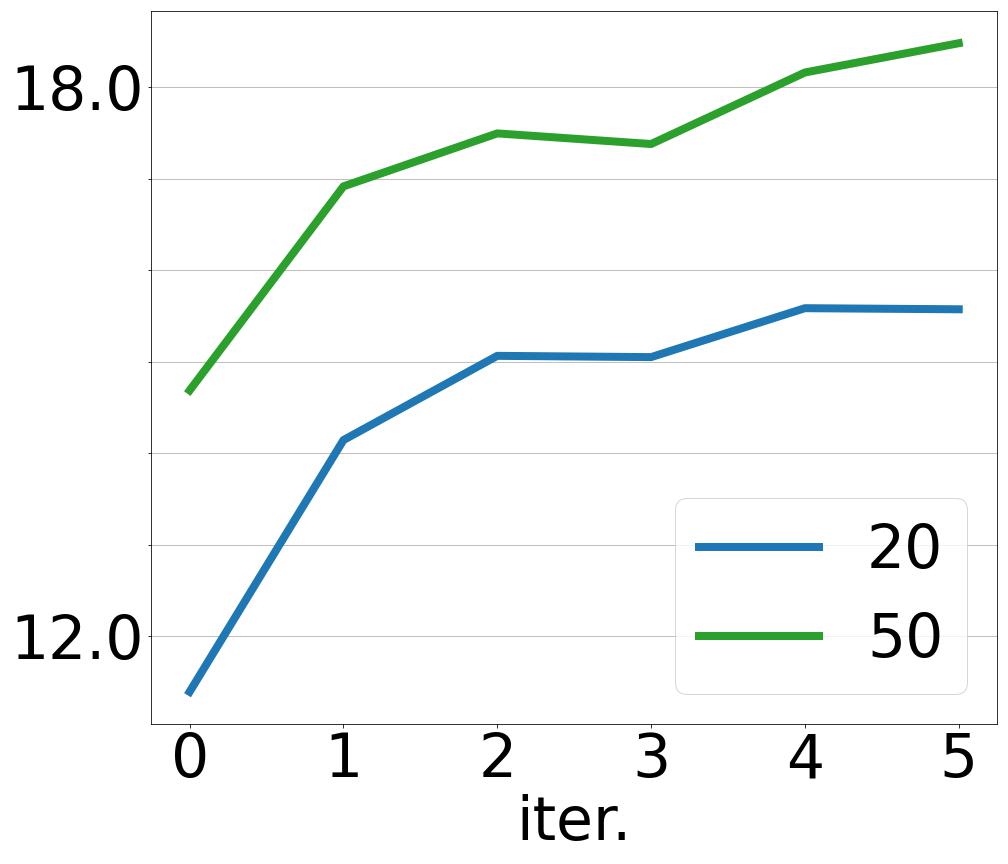} &
\includegraphics[trim=0mm 0mm 0mm 0mm,clip,width=0.32\linewidth]{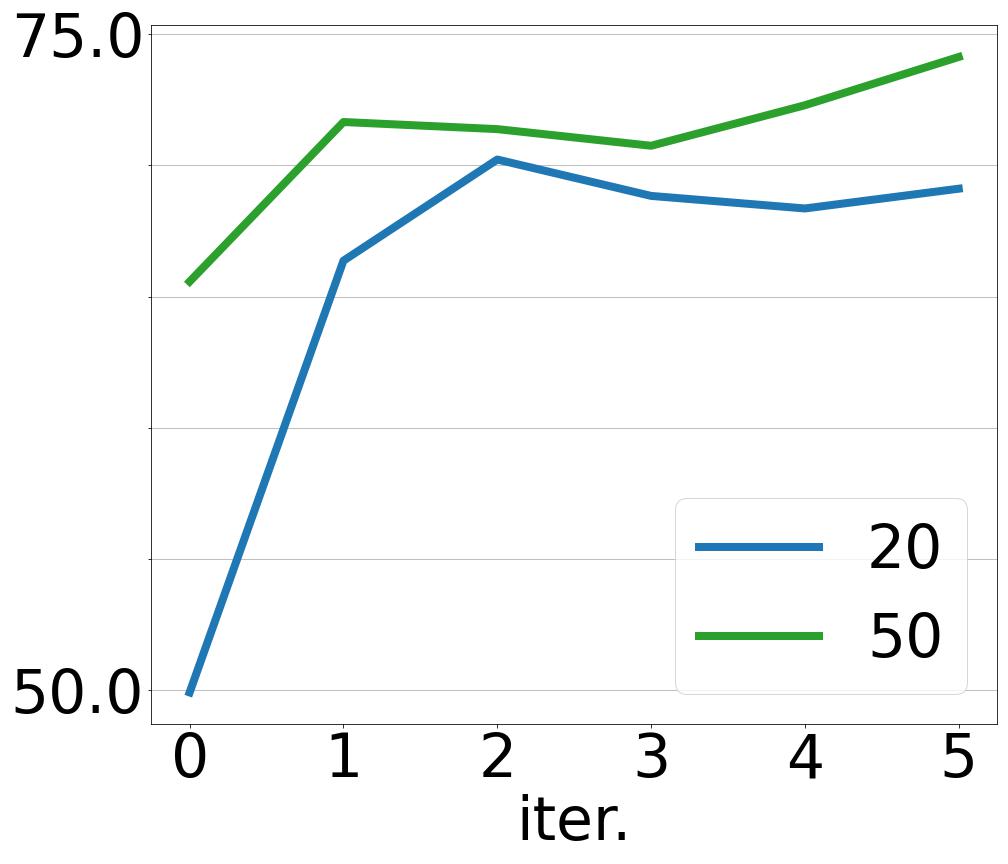} &
\includegraphics[trim=0mm 0mm 0mm 0mm,clip,width=0.32\linewidth]{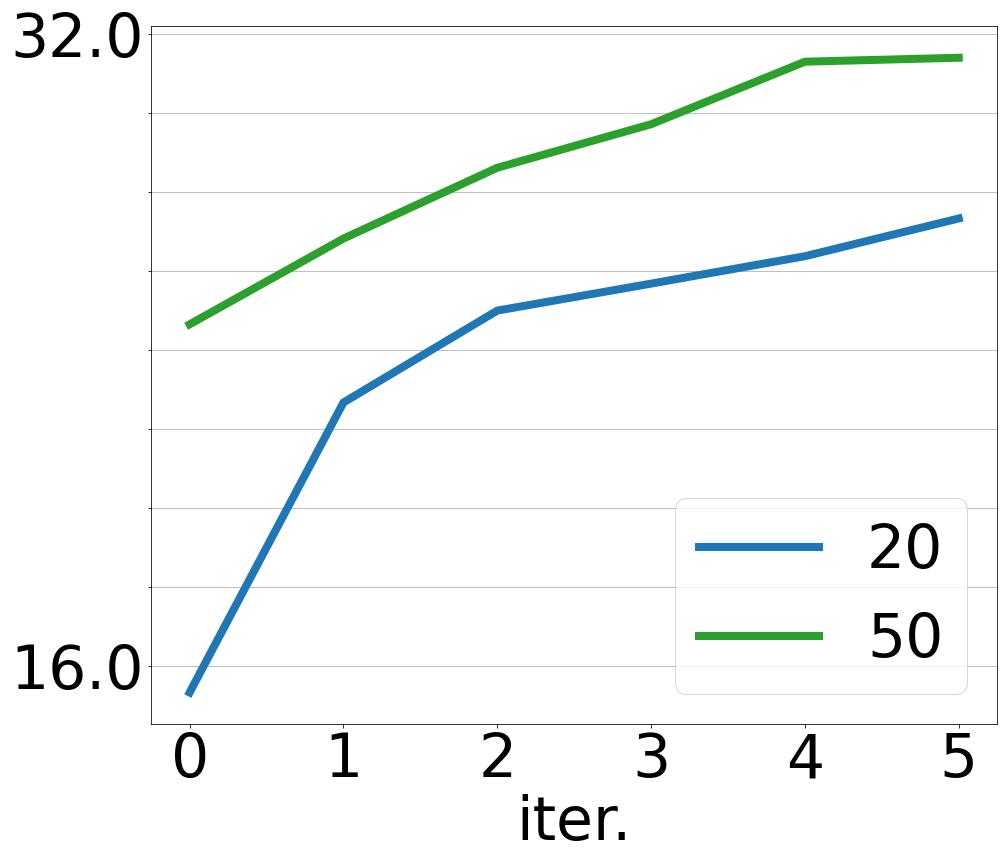}\\
\end{tabular}
}
\vspace{-0.4cm}
\caption{{\bf Learning curves} of the two metrics (AP@.5 and AP@[.5:.95]) on the three datasets}
\label{fig:accuracy_change}
\end{figure*}

\noindent{\bf Which virtual images are selected for each PTL iteration?} The top row of Figure~\ref{fig:analysis of virtualimg} shows the change in the accumulated distribution of virtual images added to the training set via each PTL iteration with respect to camera locations. By examining the distribution, we can identify which camera locations of the virtual images contribute more to the training set at each PTL iteration.

It is observed that after the 1st PTL iteration, most virtual images included in the training set are taken from the camera locations close to the human subjects. As PTL progresses, the camera locations of the virtual images included in the training set gradually spread across the UAV altitudes and rotation circle radii.  Consequently, after the 5th PTL iteration, transformed virtual images with diverse appearances from much broader camera locations are included in the final training set. This demonstrates that the proposed transformation candidate selection process is adequately designed to consider the two conflicting claims together.\smallskip

\noindent{\bf How close does the domain gap get as PTL progresses?} The bottom row of Figure~\ref{fig:analysis of virtualimg} shows the domain gap between the virtual images and the training set at each PTL iteration. We can observe the domain gap distribution of virtual images to the training set gradually becomes narrower and smaller.
Additionally, some virtual images which have not been included in the training set also appear in the long tail of the distribution.\smallskip

\begin{table*}[t]
\centering
\resizebox{\textwidth}{!}{%
\setlength{\tabcolsep}{10.0pt}
\renewcommand{\arraystretch}{1.1}
\begin{tabular}{l|c|cc|cc|cc}
\multirow{2}{*}{method} & \multirow{2}{*}{train set} & \multicolumn{2}{c|}{VisDrone} & \multicolumn{2}{c|}{Okutama-Action} & \multicolumn{2}{c}{ICG} \\
& & 20 & 50 & 20 & 50 & 20 & 50 \\ \hline
\textcolor{gray}{baseline} & \textcolor{gray}{R} & \textcolor{gray}{~~3.74/~~1.09} & \textcolor{gray}{~~6.42/~~1.86} & \textcolor{gray}{~~41.61/~~11.23} & \textcolor{gray}{~~49.84/~~13.76} & \textcolor{gray}{~~49.35/~~14.69} & \textcolor{gray}{~~66.75/~~23.91} \\\hline
pretrain-finetune & R+V & ~~4.99/~~1.46 & ~~6.25/~~1.99 & ~~44.57/~~12.78 & ~~49.06/~~15.08 & ~~66.92/~~26.67 & ~~68.41/~~29.73 \\
naive merge & R+V & ~~3.41/~~1.02 & ~~5.18/~~1.65 & ~~34.26/~~~~9.21 & ~~48.33/~~14.61 & ~~55.95/~~20.76 & ~~65.68/~~26.73 \\
~~~~~~w/ transform & R+V & ~~1.26/~~0.49 & ~~4.02/~~1.37 & ~~27.37/~~~~7.84 & ~~41.36/~~12.64 & ~~48.02/~~17.62 & ~~65.03/~~27.21 \\\hline
{\bf PTL} (5th itr.) & R+V & ~~6.83/~~1.94 & ~~9.09/~~2.85 & ~~52.89/~~15.57 & ~~59.90/~~{\bf 18.48} & ~~69.11/~~{\bf 27.33}  & ~~{\bf 74.14}/~~{\bf 31.41} \\
& & \textcolor{teal}{+3.09/+0.85} & \textcolor{teal}{+2.67/+0.99} & \textcolor{teal}{+11.28/~~+4.34} & \textcolor{teal}{+10.06/~~+4.72} & \textcolor{teal}{+19.76/+12.64} & \textcolor{teal}{~~+7.39/~~+7.50} \\
{\bf PTL} (best) & R+V & ~~{\bf 7.52}/~~{\bf 2.13} & ~~{\bf 9.33}/~~{\bf 2.94} & ~~{\bf 53.82}/~~{\bf 15.59} & ~~{\bf 60.65}/~~{\bf 18.48} & ~~{\bf 70.23}/~~{\bf 27.33}  & ~~{\bf 74.14}/~~{\bf 31.41} \\  
& & \textcolor{teal}{+3.78/+1.04} & \textcolor{teal}{+2.91/+1.08} & \textcolor{teal}{+12.21/~~+4.36} & \textcolor{teal}{+10.81/~~+4.72} & \textcolor{teal}{+20.88/+12.64} & \textcolor{teal}{~~+7.39/~~+7.50} \\
\end{tabular}
}
\vspace{-0.3cm}
\caption{{\bf Low-shot learning accuracy} with 20 and 50 real images.  AP@.5 and AP@[.5:.95] are reported in each bin. For PTL, the margin from the baseline accuracy is shown below the reported accuracy. The best accuracy for each setting is shown in bold. R and V denote the set of real images and the set of virtual images, respectively.}
\label{tab:small_shot}
\end{table*}

\noindent{\bf Accuracy variation as PTL evolves.} Figure~\ref{fig:accuracy_change} shows how the accuracy changes as PTL progresses on the three datasets. Overall, for AP@.5, accuracy increases rapidly until the 3rd iteration and does not change significantly thereafter. On the other hand, for AP@[.5:.95], accuracy continues to increase even after the 3rd iteration on the Okutama-Action and ICG datasets. This can be interpreted such that human bounding boxes are estimated more accurately as PTL progresses on these two datasets.

\subsection{Results on Low-shot Learning}
\label{ssec:low-shot_learning}

\noindent{\bf Baselines.} We compare PTL with the method utilizing only real images for training (i.e., `baseline') and other three methods also leveraging virtual images in conjunction with real images for training (i.e., `pretrain-finetune', `naive merge', and `naive merge w/ transform') in terms of human detection accuracy. `Pretrain-finetune' is the strategy to train a model by pre-training on virtual images and then fine-tuning on real images, which is the most widely used approach leveraging virtual images in previous works~\cite{AHandaCVPR2016,DKimCVPR2016,AGaidonCVPR2016,MFabbriECCV2018,GVarolCVPR2018,JMuCVPR2020,SMashraCVPR2022,XGuoCVPR2022,ZJinCVPR2022,KBaekCVPR2022}. `Naive merge' is the strategy that uses a training set naively merging from real and virtual images for model training, which has also been used in previous works~\cite{GRosCVPR2016,SRichterECCV2016}. `Naive merge w/ transform' naively adds transformed virtual images to the training set, where the transformation generator is trained with the CycleGAN framework by considering all virtual images and all real images as ``source'' and ``target''.\smallskip

\begin{table}[t]
\centering
\resizebox{\linewidth}{!}{%
\setlength{\tabcolsep}{10.0pt}
\renewcommand{\arraystretch}{1.1}
\begin{tabular}{c|c||c|c}
$\tau$ & VisDrone & Okutama-Action & ICG \\\hline
\multicolumn{1}{r|}{1} & ~~9.48/~~3.01 & ~~39.37/~~10.45 & ~~27.87/~~~~7.75 \\
\multicolumn{1}{r|}{5} & ~~9.09/~~2.85 & ~~{\bf 42.39}/~~{\bf 11.41} & ~~29.26/~~~~7.27 \\
\multicolumn{1}{r|}{10} & ~~{\bf 9.68}/~~{\bf 2.87} & ~~37.48/~~~~9.51 & ~~33.06/~~~~7.66 \\
\multicolumn{1}{r|}{100} & ~~8.97/~~2.54 & ~~38.25/~~10.18 & ~~33.78/~~~~9.29 \\
\multicolumn{1}{r|}{1000} & ~~8.90/~~2.63 & ~~39.15/~~10.00 & ~~{\bf 43.90}/~~{\bf 11.97} \\
\end{tabular}
}
\vspace{-0.3cm}
\caption{{\bf Varying $\tau$} in PTL (after 5th iteration). Models are trained on the VisDrone dataset with 50 shot learning setup.}
\label{tab:ablation_tau}
\end{table}

\noindent{\bf Main Results.} In Table~\ref{tab:small_shot}, we compare PTL to the baselines in terms of human detection accuracy in two low-shot detection regimes (i.e., 20 and 50 real images are used for training) on the three real-world UAV-based datasets. Low-shot detection is a suitable task to validate the proper use of virtual images as notable effects can be expected from adequately expanded datasets.

In all cases, the previous methods leveraging virtual images do not present significantly better, or even worse, accuracy than their counterpart (i.e., `baseline') using real images only for training. `Pretrain-finetune' is the only method, except for PTL, that presents better accuracy than the baseline in some cases. This two-step method is effective in avoiding adverse effects due to the large domain gap while indirectly taking advantage of the large-scale dataset. However, the increase in accuracy is marginal as the task-specific properties (i.e., human detection) of the large-scale virtual dataset used for pre-training are not fully exploited due to the catastrophic forgetting issue inherent in this indirect method.
In addition, `naive merge w/ transformation', which is the only previous method that uses transformed virtual images, presents significantly reduced accuracy, which can be regarded as the adverse effects (e.g., transformation quality degradation) when the large domain gap is not properly addressed. The final model obtained after the 5th PTL iteration or the best model achieved when PTL progresses consistently presents significantly better performance than any compared method. This demonstrates \emph{the effectiveness of PTL such that accuracy is substantially improved by expanding the training set using virtual images while the large domain gap is appropriately addressed}.\smallskip


\noindent{\bf Ablation: The effect of $\tau$.} In Table~\ref{tab:ablation_tau}, we compare five different $\tau$ values to investigate the effect of $\tau$ used to control the sampling weights when selecting transformation candidates (eq.~\ref{eq:weight}). It is found that as large $\tau$ values ($\geq$ 100) are used, the accuracy on the VisDrone dataset decreases while the accuracy on the ICG dataset increases. The finding indicates that adding more virtual images with a large domain gap to the training set using a large $\tau$ has no significant effect or even adverse effect in terms of accuracy when the training set and the test set are in the same domain. However, a large $\tau$ shows a remarkable effect in the cross-domain setting, especially when the training set and the test set have very different characteristics, such as the VisDrone train set and the ICG test set. We use $\tau$=5 throughout the experiment for the same-domain setup, but we can also consider using a large $\tau$ for the cross-domain setup.\smallskip

\noindent{\bf Ablation: The effect of $n$.} $n$, which is the number of virtual images added to the training set per PTL iteration, can affect not only the optimality of training but also the scalability of models trained by PTL toward other datasets different from the real dataset used for training. Accordingly, to find the optimal value of $n$ by taking these effects into account, we carry out ablation experiments under the cross-domain setup as shown in Table~\ref{tab:num_virtual_per_itr}.

We compare three cases when $n$ is 50, 100, and 200. When the training set and the test set comes from the same domain (i.e., training and testing on the VisDrone dataset), the human detection accuracy is not very sensitive to $n$. The best accuracy is obtained with the fewest virtual images (i.e., $n$=50). The insensitivity to the hyperparameter $n$ is also observed in the Okutama-Action dataset, having relatively similar properties to the VisDrone dataset under the cross-domain setup. However, for ICG, which is considered to have noticeably different properties from the VisDrone dataset under the cross-domain setup, the accuracy is significantly improved as more virtual images (i.e., $n$=200) are used per PTL iteration. This implies that using more virtual images through more PTL iterations tends to further reduce the cross-domain gap when the discrepancy between the training set and the test set is considerably large, such as VisDrone vs. ICG, resulting in improved PTL scalability. Considering the trade-off between the training optimality and the PTL scalability, we use $n$=100 throughout all other experiments in this paper.\smallskip

\begin{table}[t]
\centering
\resizebox{\linewidth}{!}{%
\setlength{\tabcolsep}{7.0pt}
\renewcommand{\arraystretch}{1.1}
\begin{tabular}{c|c||c|c}
\# img per itr & VisDrone & Okutama-Action & ICG \\\hline
50 & ~~{\bf 9.59}/~~2.90 & ~~41.62/~~11.19 & ~~25.94/~~~~6.04 \\
100 & ~~9.33/~~{\bf 2.94} & ~~{\bf 42.39}/~~{\bf 11.46} & ~~30.01/~~~~7.36 \\
200 & ~~9.04/~~2.91 & ~~41.29/~~11.18 & ~~{\bf 35.50}/{\bf ~~10.20} \\
\end{tabular}
}
\vspace{-0.3cm}
\caption{{\bf Varying \# of virtual images added to the training set per PTL iteration.} Models are trained on the VisDrone dataset with 50 shot learning setup. The reported accuracies are obtained by using the best PTL models.}
\label{tab:num_virtual_per_itr}
\end{table}

\begin{figure*}[t]
\centering
\resizebox{\linewidth}{!}{%
\setlength{\tabcolsep}{3.0pt}
\begin{tabular}{cccccc}
\makebox[.25\linewidth][c]{
\setlength{\tabcolsep}{0.5pt}
\begin{tabular}{ccc}
\includegraphics[trim=0mm 0mm 0mm 0mm,clip,width=.083\linewidth]{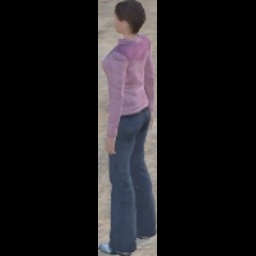}
& 
\includegraphics[trim=0mm 0mm 0mm 0mm,clip,width=.083\linewidth]{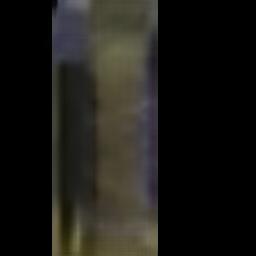}
& 
\includegraphics[trim=0mm 0mm 0mm 0mm,clip,width=.083\linewidth]{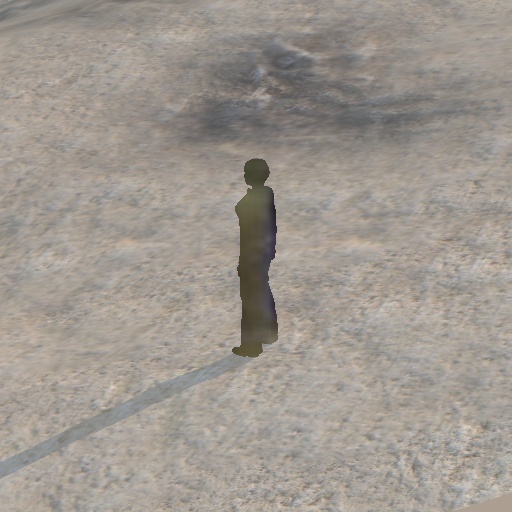}
\end{tabular}
}
&
\makebox[.25\linewidth][c]{
\setlength{\tabcolsep}{0.5pt}
\begin{tabular}{ccc}
\includegraphics[trim=0mm 0mm 0mm 0mm,clip,width=.083\linewidth]{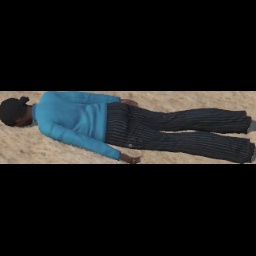}
&
\includegraphics[trim=0mm 0mm 0mm 0mm,clip,width=.083\linewidth]{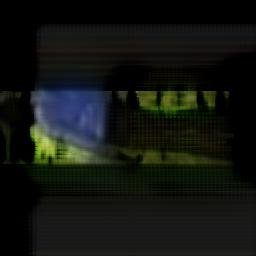}
& 
\includegraphics[trim=0mm 0mm 0mm 0mm,clip,width=.083\linewidth]{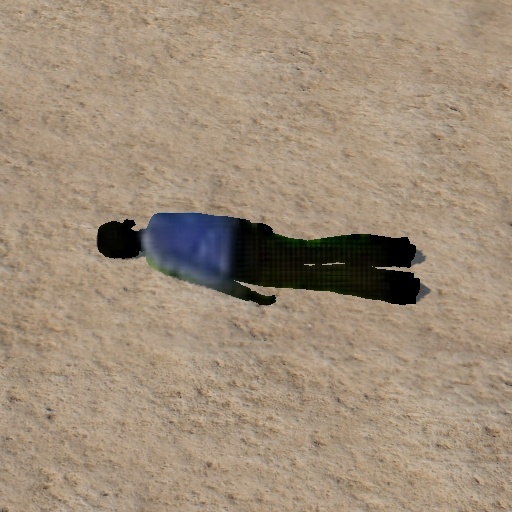}
\end{tabular}
}
&&&
\makebox[.25\linewidth][c]{
\setlength{\tabcolsep}{0.5pt}
\begin{tabular}{ccc}
\includegraphics[trim=0mm 0mm 0mm 0mm,clip,width=.083\linewidth]{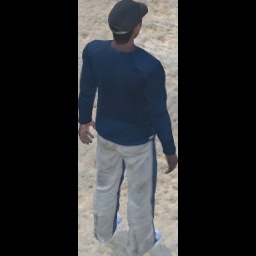}
& 
\includegraphics[trim=0mm 0mm 0mm 0mm,clip,width=.083\linewidth]{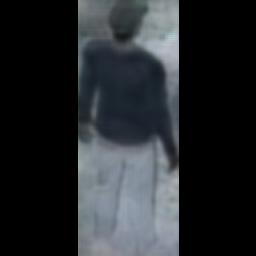}
& 
\includegraphics[trim=0mm 0mm 0mm 0mm,clip,width=.083\linewidth]{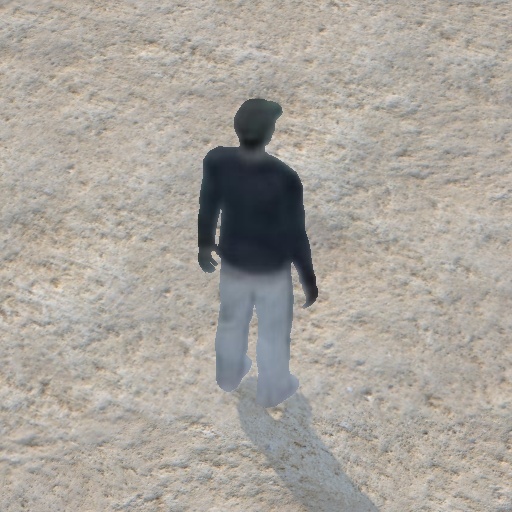}
\end{tabular}
}
&
\makebox[.25\linewidth][c]{
\setlength{\tabcolsep}{0.5pt}
\begin{tabular}{ccc}
\includegraphics[trim=0mm 0mm 0mm 0mm,clip,width=.083\linewidth]{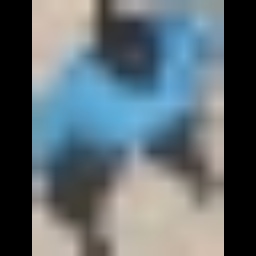}
& 
\includegraphics[trim=0mm 0mm 0mm 0mm,clip,width=.083\linewidth]{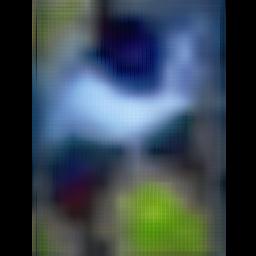}
& 
\includegraphics[trim=0mm 0mm 0mm 0mm,clip,width=.083\linewidth]{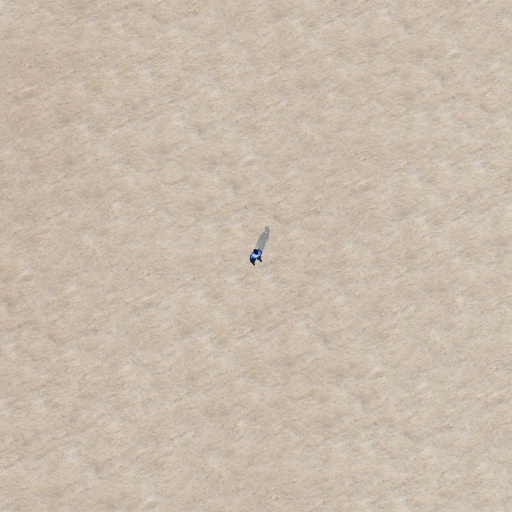}
\end{tabular}
}
\\
\makebox[.25\linewidth][c]{
\setlength{\tabcolsep}{0.5pt}
\begin{tabular}{ccc}
\includegraphics[trim=0mm 0mm 0mm 0mm,clip,width=.083\linewidth]{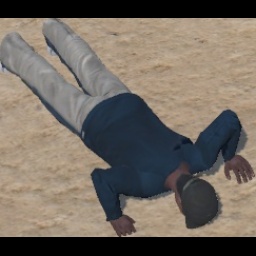}
& 
\includegraphics[trim=0mm 0mm 0mm 0mm,clip,width=.083\linewidth]{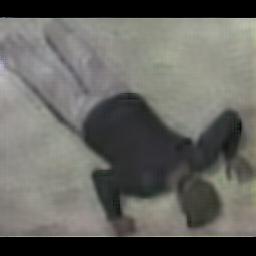}
& 
\includegraphics[trim=0mm 0mm 0mm 0mm,clip,width=.083\linewidth]{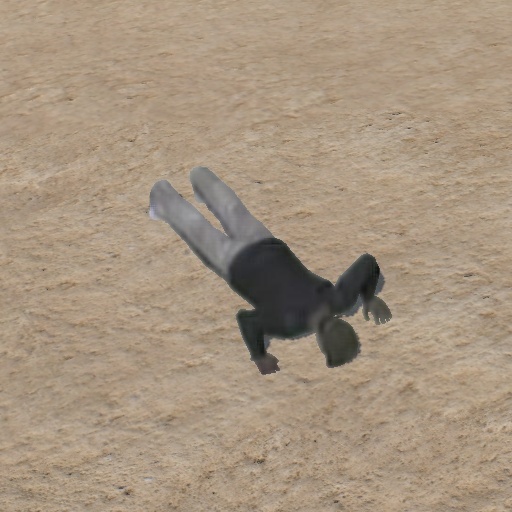}
\end{tabular}
}
&
\makebox[.25\linewidth][c]{
\setlength{\tabcolsep}{0.5pt}
\begin{tabular}{ccc}
\includegraphics[trim=0mm 0mm 0mm 0mm,clip,width=.083\linewidth]{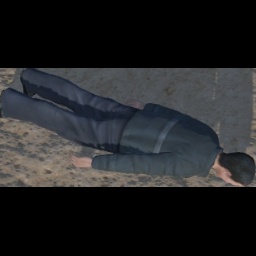}
&
\includegraphics[trim=0mm 0mm 0mm 0mm,clip,width=.083\linewidth]{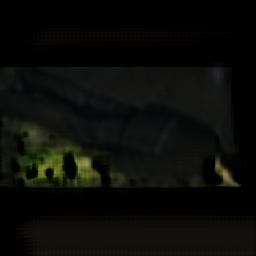}
& 
\includegraphics[trim=0mm 0mm 0mm 0mm,clip,width=.083\linewidth]{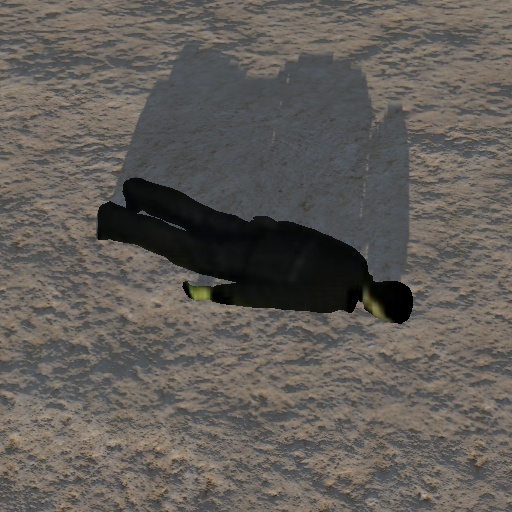}
\end{tabular}
}
&&&
\makebox[.25\linewidth][c]{
\setlength{\tabcolsep}{0.5pt}
\begin{tabular}{ccc}
\includegraphics[trim=0mm 0mm 0mm 0mm,clip,width=.083\linewidth]{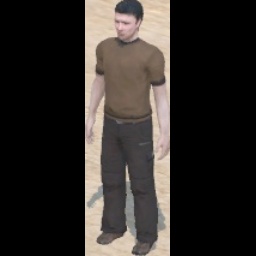}
& 
\includegraphics[trim=0mm 0mm 0mm 0mm,clip,width=.083\linewidth]{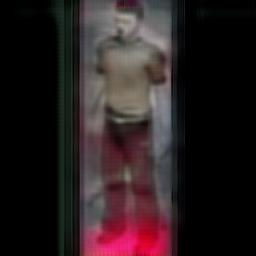}
& 
\includegraphics[trim=0mm 0mm 0mm 0mm,clip,width=.083\linewidth]{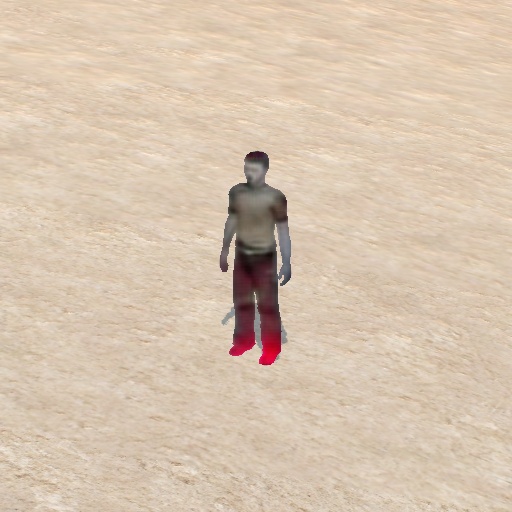}
\end{tabular}
}
&
\makebox[.25\linewidth][c]{
\setlength{\tabcolsep}{0.5pt}
\begin{tabular}{ccc}
\includegraphics[trim=0mm 0mm 0mm 0mm,clip,width=.083\linewidth]{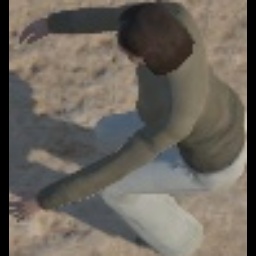}
& 
\includegraphics[trim=0mm 0mm 0mm 0mm,clip,width=.083\linewidth]{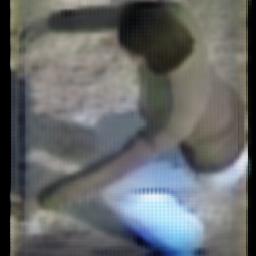}
& 
\includegraphics[trim=0mm 0mm 0mm 0mm,clip,width=.083\linewidth]{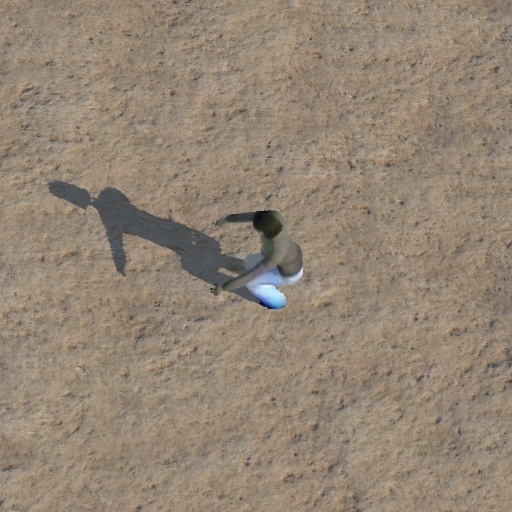}
\end{tabular}
}
\\
\multicolumn{2}{c}{Naive merge w/ transform} &&&
\multicolumn{2}{c}{PTL} \\
\end{tabular}
}
\vspace{-0.4cm}
\caption{{\bf Sample Virtual2Real transformation output (VisDrone, 50-shot).} Each set consists of three images: original virtual image (left), transformed image (middle), and transformed image with background (right).}
\label{fig:transformation_samples_Vis_50}
\end{figure*}

\begin{table*}[t]
\resizebox{\textwidth}{!}{%
\setlength{\tabcolsep}{13.0pt}
\renewcommand{\arraystretch}{1.1}
\centering
\begin{tabular}{l|cc|cc|cc}
\multirow{2}{*}{method} & \multicolumn{2}{c|}{VisDrone} & \multicolumn{2}{c|}{Okutama-Action} & \multicolumn{2}{c}{ICG} \\
& 20 & 50 & 20 & 50 & 20 & 50 \\\hline
& \multicolumn{2}{l|}{{\it Vis $\rightarrow$ Vis}} & \multicolumn{2}{l|}{{\it Oku $\rightarrow$ Oku}} & \multicolumn{2}{l}{{\it ICG $\rightarrow$ ICG}} \\
\textcolor{gray}{baseline} & \textcolor{gray}{~~~~3.74/~~~~1.09} & \textcolor{gray}{~~~~6.42/~~~~1.86} & \textcolor{gray}{~~41.61/~~11.23} & \textcolor{gray}{~~49.84/~~13.76} & \textcolor{gray}{~~49.35/~~14.69} & \textcolor{gray}{~~66.75/~~23.91} \\
PTL (5th itr.) & ~~~~6.83/~~~~1.94 & ~~~~9.09/~~~~2.85 & ~~52.89/~~15.57 & ~~59.90/~~18.48 & ~~69.11/~~27.33 & ~~74.14/~~31.41\\
PTL (best) & ~~~~7.52/~~~~2.13 & ~~~~9.33/~~~~2.94 & ~~53.82/~~15.59 & ~~60.65/~~18.48 & ~~70.23/~~27.33 & ~~74.14/~~31.41\\\hline\hline
& \multicolumn{2}{l|}{{\it Oku $\rightarrow$ Vis}} & \multicolumn{2}{l|}{{\it Vis $\rightarrow$ Oku}} & \multicolumn{2}{l}{{\it Vis $\rightarrow$ ICG}} \\
\textcolor{gray}{baseline} & \textcolor{gray}{~~~~1.62/~~~~0.47} & \textcolor{gray}{~~~~2.04/~~~~0.57} & \textcolor{gray}{~~17.13/~~~~4.53} & \textcolor{gray}{~~36.82/~~~~9.87} & \textcolor{gray}{~~~~2.92/~~~~0.56} & \textcolor{gray}{~~~~7.46/~~~~1.83} \\
PTL (5th itr.) & ~~~~2.72/~~~~0.94 & ~~~~3.05/~~~~1.07 & ~~30.72/~~~~7.45 & ~~42.39/~~11.41 & ~~26.86/~~~~7.22 & ~~29.26/~~~~7.27\\
PTL (best) & ~~~~3.00/~~~~1.22 & ~~~~3.56/~~~~1.17 & ~~33.25/~~~~8.59 & ~~42.39/~~11.46 & ~~29.60/~~~~7.69 & ~~30.01/~~~~7.36\\\hline
& \multicolumn{2}{l|}{{\it ICG $\rightarrow$ Vis}} & \multicolumn{2}{l|}{{\it ICG $\rightarrow$ Oku}} & \multicolumn{2}{l}{{\it Oku $\rightarrow$ ICG}} \\
\textcolor{gray}{baseline} & \textcolor{gray}{~~~~0.54/~~~~0.13} & \textcolor{gray}{~~~~0.99/~~~~0.26} & \textcolor{gray}{~~~~3.56/~~~~0.75} & \textcolor{gray}{~~10.27/~~~~2.49} & \textcolor{gray}{~~~~5.37/~~~~1.25} & \textcolor{gray}{~~~~5.23/~~~~1.20} \\
PTL (5th itr.) & ~~~~1.09/~~~~0.33 & ~~~~1.61/~~~~0.50 & ~~11.19/~~~~2.58 & ~~14.20/~~~~3.56 & ~~28.98/~~~~8.14 & ~~25.39/~~~~6.53\\
PTL (best) & ~~~~1.58/~~~~1.02 & ~~~~1.70/~~~~0.63 & ~~12.82/~~~~2.96 & ~~14.20/~~~~3.71 & ~~28.98/~~~~8.14 & ~~26.62/~~~~6.53\\
\end{tabular}
}
\vspace{-0.3cm}
\caption{{\bf Cross-domain detection accuracy.} The table shows experiments with 3$\times$3 cross-domain setups. For each setup, datasets shown to the left and right of the arrow are the training and test sets, respectively. The accuracies of PTL and the baseline without using virtual images for training are shown. Setups on the top use training and test images from the same domain, which provides a baseline accuracy in the cross-domain setups. All setups in each column are tested on the same dataset and the same low-shot regime.}
\label{tab:cross_domain}
\end{table*}

\noindent{\bf Qualitative analysis of transformation.} Figures~\ref{fig:transformation_samples_Vis_50} show several samples of transformed virtual images included in the training set using methods with virtual2real transformation (i.e., PTL and `naive merge w/ transformation') in the `VisDrone and 50-shot' setup. First, the quality of the transformed image by 'naive merge w/transformation' generally deteriorates a lot, and there are samples, in some severe cases, in which it is difficult to recognize the human appearance anymore. On the other hand, in the case of the virtual image transformed by PTL considering the domain gap when training the transformation generator, it can be seen that only the pattern is changed to be similar to the image of the real dataset while the human pose is maintained well. This qualitative analysis supports the validity of our claim that \emph{the domain gap between virtual images and real images should be considered when training the virtual2real transformation generator}.

\subsection{Results on Cross-domain Detection}
\label{ssec:cross-domain_detect}

In Table~\ref{tab:cross_domain}, we show the accuracy for the cross-domain setups on the three datasets. The ``cross-domain detection'' setups are used to validate the impact of using virtual images for training when training and test images have distinct characteristics, such as when the distributions of human appearances with respect to human poses and viewing angles are different.

First, it is observed that using PTL yields much better accuracies than the baseline when using the same real training dataset. Despite the inherent difficulty of cross-domain learning, it is also observed that leveraging virtual images through PTL produces results not far behind the accuracy of the baseline (shown in the first row of Table~\ref{tab:cross_domain}) which uses the training images from the same domain as the test images. However, the improvement in the cross-domain detection accuracy when using the ICG dataset for training is relatively low compared to other cases. This is because the ICG dataset has very different characteristics from other datasets, and virtual images added to the training set do not reduce this difference. Note that if the real images in the initial training set have very distinct appearances, a very different set of virtual images may be selected in the first PTL iteration. Thus, the disparity may not be overcome even if the PTL progresses further.
Nevertheless, in general, we can confirm that \emph{human detectors trained using virtual images through PTL can improve substantially detection accuracy, regardless of which real dataset is used during training}.

\section{Discussion}
\label{sec:discussion}

Our method has been proved effective in leveraging virtual images during training as it presents much better accuracy than any other previous methods for low-shot learning task, where scaling up the training dataset can significantly impact. In addition, compared to the baseline which does not use virtual images, our method also presents remarkable accuracy on cross-domain detection, where the real training and test datasets are from two distinct domains with very different characteristics. 

Despite the merit of the proposed method, there is still room to improve PTL further. Specifically, the current version of PTL can only leverage a subset of the entire virtual images within a limited number of iterations, beyond which the accuracy may decrease. We hope that more advanced methods can be developed to address this issue so that PTL can progress for more iterations with a continuous accuracy increase until all virtual images are leveraged for training. 


\medskip

\noindent{\bf Acknowledgements.} This research was sponsored by the Defense Threat Reduction Agency (DTRA) Contract Number: A2205097021023559.

\appendix

\section{Gaussian Discriminant Analysis for Modeling Representation Space of Detector}
\label{ssec:GDA}

In this section, we describe modeling the representation space of a general object detector by fitting a multivariate Gaussian distribution. We denote the random variable of the input and its label of a linear classifier as ${\bf x} \in \mathcal{X}$ and $y = \{y_c\}_{c=1,\cdots,C} \in \mathcal{Y}, y_c=\{0, 1\}$, respectively. Then, the posterior distribution defined by the linear classifier whose output formed by the sigmoid function can be expressed as follows: 
\begin{eqnarray}
    P(y_c=1|{\bf x}) &=& \frac{1}{1+\exp{\left(\minus w_c{\bf x}\minus b_c\right)}} \nonumber\\
    &=& \frac{\exp{\left(w_c{\bf x}+b_c\right)}}{\exp{\left(w_c{\bf x}+b_c\right)}+1},\label{eq:post_sigmoid}
\end{eqnarray}
where $w_c$ and $b_c$ are weights and bias of the linear classifier for a category $c$, respectively.

Gaussian Discriminant Analysis (GDA) models the posterior distribution of the classifier by assuming that the class conditional distribution ($P({\bf x}|y)$) and the class prior distribution ($P(y)$) follow the multivariate Gaussian and the Bernoulli distributions, respectively, as follows:
\begin{eqnarray}
P({\bf x}|y_c=0) &=& \mathcal{N}(\mu_0, \Sigma_0), \nonumber\\
P({\bf x}|y_c=1) &=& \mathcal{N}(\mu_1, \Sigma_1),\nonumber\\
P(y_c=0) &=& \beta_0/\left(\beta_0+\beta_1\right),\nonumber\\
P(y_c=1) &=& \beta_1/\left(\beta_0+\beta_1\right),
\end{eqnarray}
where $\mu_{\{0, 1\}}$ and $\Sigma_{\{0, 1\}}$ are the mean and covariance of the multivariate Gaussian distribution, and $\beta_{\{0, 1\}}$ is the unnormalized prior for the category $c$ and the background.

For the special case of GDA where all categories share the same covariance matrix (i.e., $\Sigma_0 = \Sigma_1 = \Sigma_c$), known as Linear Discriminant Analysis (LDA), the posterior distribution ($P(y_c|{\bf x})$) can be expressed with $P({\bf x}|y_c)$ and $P(y_c)$ as follows:
\begin{eqnarray}
&&\!\!\!\!\!P(y_c=1|{\bf x}) \nonumber\\
&& = \frac{P(y_c=1)P({\bf x}|y_c=1)}{P(y_c=0)P({\bf x}|y_c=0)+P(y_c=1)P({\bf x}|y_c=1)}\nonumber\\
&& = \frac{
\begin{aligned}\exp\Bigl(\left(\mu_1\minus\mu_0\right)^\top\Sigma_c^{\minus 1}{\bf x}\ldots~~~~~~~~~~~~~~~~~~~~~~~~~~~~~~~~~~~\\
-\frac{1}{2}\mu_1^\top\Sigma_c^{\minus 1}\mu_1+\frac{1}{2}\mu_0^\top\Sigma_c^{\minus 1}\mu_0+\log{\beta_1/\beta_0} \Bigr)\end{aligned}
}{
\begin{aligned}\exp\Bigl(\left(\mu_1\minus\mu_0\right)^\top\Sigma_c^{\minus 1}{\bf x}\ldots~~~~~~~~~~~~~~~~~~~~~~~~~~~~~~~~~~~\\
-\frac{1}{2}\mu_1^\top\Sigma_c^{\minus 1}\mu_1+\frac{1}{2}\mu_0^\top\Sigma_c^{\minus 1}\mu_0+\log{\beta_1/\beta_0} \Bigr)+1\end{aligned}
}. \nonumber\\
\label{eq:post_gda}
\end{eqnarray}
Note that the quadratic term is canceled out since the shared covariance matrix is used. The posterior distribution derived by GDA in eq.~\ref{eq:post_gda} then becomes equivalent to the posterior distribution of the linear classifier with the sigmoid function in eq.~\ref{eq:post_sigmoid} when $w_c = \left(\mu_1\minus\mu_0\right)^\top\Sigma_c^{\minus 1}$ and $b_c = -\frac{1}{2}\mu_1^\top\Sigma_c^{\minus 1}\mu_1+\frac{1}{2}\mu_0^\top\Sigma_c^{\minus 1}\mu_0+\log{\beta_1/\beta_0}$. This implies that the representation space formed by {\bf x} can be modeled by a multivariate Gaussian distribution. 

Based on the above derivation, if ${\bf x}$ is the output of the penultimate layer of an object detector for a region proposal, and a linear classifier defined by $w_c$ and $b_c$ is the last layer of the object detector, it can be said that the representation space of the object detector for a category $c$ can be modeled with a multivariate Gaussian distribution. In other words, the representation space for a category $c$ can be represented by two parameters $\mu_1$ (i.e., $\mu_c$) and $\Sigma_c$ of the multivariate Gaussian distribution.\smallskip

\begin{table*}[t]
\centering
\begin{tabular}{ccc}
\begin{subtable}{.50\linewidth}
\caption{Detector training}
\label{tab:detector_training}
\centering
\resizebox{\textwidth}{!}{%
\setlength{\tabcolsep}{6.0pt}
\renewcommand{\arraystretch}{1.1}
\begin{tabular}{l|ccccc}
\multirow{2}{*}{config} & \multirow{2}{*}{baseline} & \multicolumn{2}{c}{pretrain-finetune} & \multirow{2}{*}{naive merge} & \multirow{2}{*}{PTL} \\\cline{3-4}
 & & pretrain & finetune & & \\ \hline
optimizer & \multicolumn{5}{c}{SGD} \\
momentum & \multicolumn{5}{c}{0.9} \\
weight decay & \multicolumn{5}{c}{0.0001} \\
base $lr$ & \multicolumn{5}{c}{0.001} \\
$lr$ schedule & \multicolumn{5}{c}{multi-step $lr$} \\
gamma & \multicolumn{5}{c}{0.1} \\
warmup iter. & \multicolumn{5}{c}{1000} \\
total iter. & 6000 & 6000 & 600 & 6000 & 6000 \\
steps & 5000 & 5000 & 500 & 5000 & 5000 \\
batch size & \multicolumn{5}{c}{16} \\
filter empty annot. & \multicolumn{5}{c}{False} \\
box threshold & \multicolumn{5}{c}{10} \\
aspect ratio grouping & \multicolumn{5}{c}{False} \\
\end{tabular}
}
\end{subtable}
&
&
\begin{subtable}{.255\linewidth}
\caption{Generator training}
\label{tab:generator_training}
\centering
\resizebox{\textwidth}{!}{%
\setlength{\tabcolsep}{6.0pt}
\renewcommand{\arraystretch}{1.1}
\begin{tabular}{l|cc}
\multirow{2}{*}{config} & \multirow{2}{2cm}{naive merge w/ transform} & \multirow{2}{*}{PTL} \\
&&\\\hline
optimizer & \multicolumn{2}{c}{Adam} \\
momentum & \multicolumn{2}{c}{$\beta_1$, $\beta_2$ = 0.5, 0.999} \\
$lr$ & \multicolumn{2}{c}{0.0002} \\
total epochs & 80 & 100 \\
batch size & \multicolumn{2}{c}{8} \\
load size & \multicolumn{2}{c}{256} \\
preprocess & \multicolumn{2}{c}{None} \\
\multicolumn{3}{c}{}\\
\multicolumn{3}{c}{}\\
\multicolumn{3}{c}{}\\
\multicolumn{3}{c}{}\\
\multicolumn{3}{c}{}\\
\multicolumn{3}{c}{}\\
\end{tabular}
}
\end{subtable}
\end{tabular}
\caption{{\bf Training settings.}}
\label{tab:train_setup}
\end{table*}

\noindent{\bf Discussion.} The sigmoid function can be viewed as a special case of the softmax function defined for a single category as both functions take the form of an exponential term for the category-of-interest normalized by the sum of exponential terms for all considered categories. Therefore, it is straightforward to derive the modeling for the sigmoid-based detector from the previous work  by \cite{KLeeNeurIPS2018}, who shows that the softmax-based classifier can be modeled with a multivariate Gaussian distribution in the representation space. However, our derivation is still meaningful in that it extends the applicability of an existing modeling limited to a certain type of classifier (i.e., based on softmax) to general object detectors (i.e., based on sigmoid). Most object detectors, especially one-stage detectors, generally use the sigmoid function, which does not consider other categories when calculating the model output for a certain category, since more than one category can be active on a single output.

\section{Implementation Details}

\noindent{\bf Multi-scale training.} We apply the multi-scale training strategy when preparing input images, in addition to the 5-level multi-scaling property provided by the detector's FPN module, to train the detector. The goal is to make the detector more robust to the variations of human size in the images. Since the real and virtual datasets have widely varying human sizes in the images, we apply different scaling factors for each dataset to share similar human sizes after image rescaling. Specifically, for the real dataset, the input image is resized by one of the scaling factors randomly selected from $\{768, 800, 832, 864\}$ for the short side, in which the long side is constrained not to be larger than 1440, for every training iteration. For the virtual dataset, the scaling factors are $\{128, 256, 384, 512\}$ for the short side and 512 for the long side. \smallskip

\begin{figure*}[t]
\centering
\resizebox{.7\linewidth}{!}{%
\setlength{\tabcolsep}{4.0pt}
\begin{tabular}{c|c|c|c|c|c|}
\multicolumn{2}{c}{\includegraphics[trim=0mm 0mm 0mm 0mm,clip,width=0.4\linewidth]{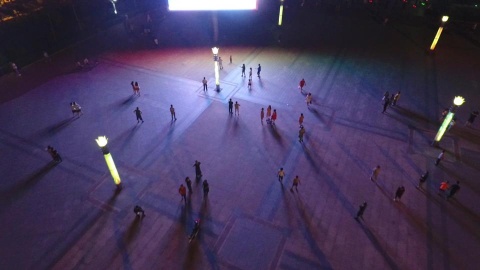}} &
\multicolumn{2}{c}{\includegraphics[trim=0mm 0mm 0mm 0mm,clip,width=0.4\linewidth]{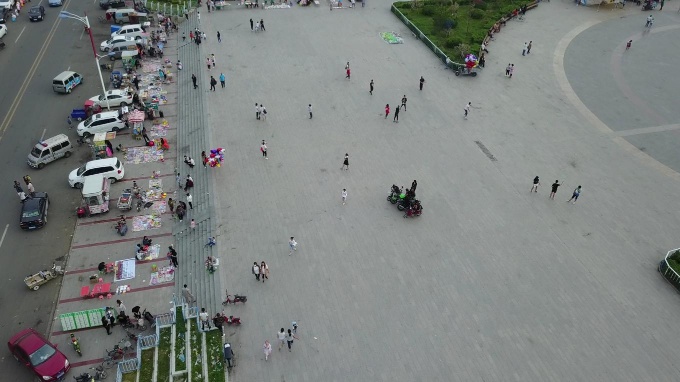}} &
\multicolumn{2}{c}{\includegraphics[trim=0mm 0mm 0mm 0mm,clip,width=0.4\linewidth]{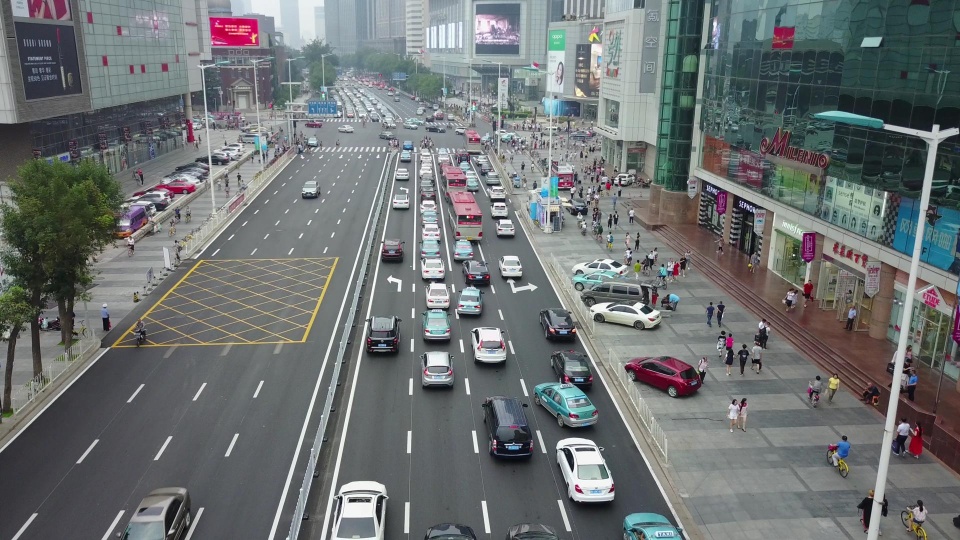}} \\
\multicolumn{6}{c}{\Large{VisDrone}} \\
\multicolumn{6}{c}{}\\
\multicolumn{2}{c}{\includegraphics[trim=0mm 0mm 0mm 0mm,clip,width=0.4\linewidth]{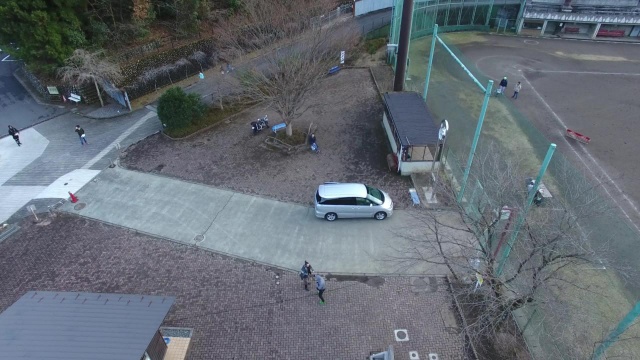}} &
\multicolumn{2}{c}{\includegraphics[trim=0mm 0mm 0mm 0mm,clip,width=0.4\linewidth]{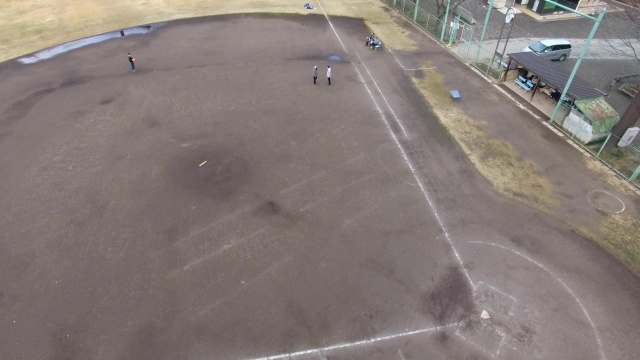}} &
\multicolumn{2}{c}{\includegraphics[trim=0mm 0mm 0mm 0mm,clip,width=0.4\linewidth]{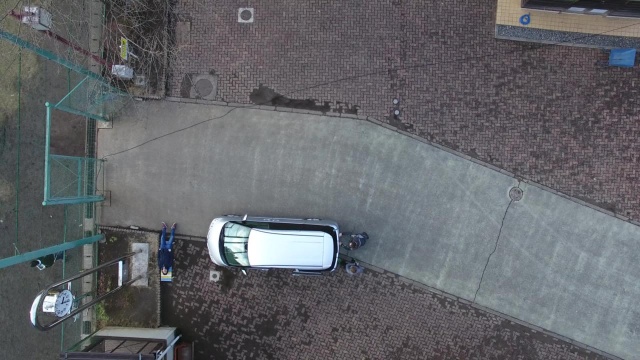}} \\
\multicolumn{6}{c}{\Large{Okutama-Action}} \\
\multicolumn{6}{c}{}\\
\multicolumn{2}{c}{\includegraphics[trim=0mm 0mm 0mm 0mm,clip,width=0.4\linewidth]{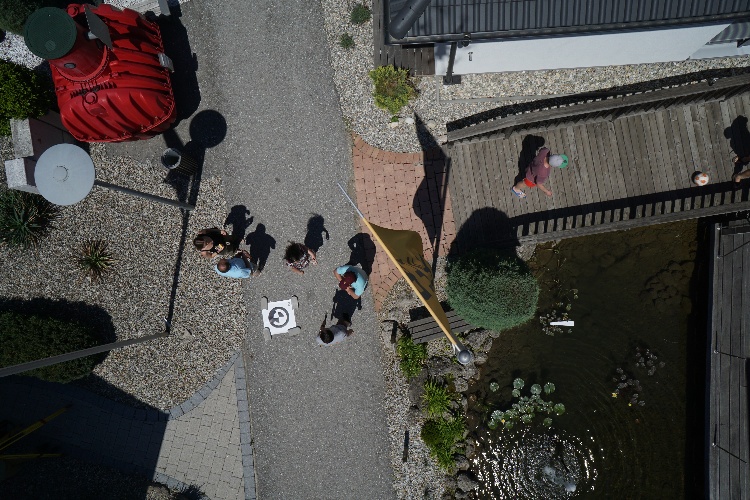}} &
\multicolumn{2}{c}{\includegraphics[trim=0mm 0mm 0mm 0mm,clip,width=0.4\linewidth]{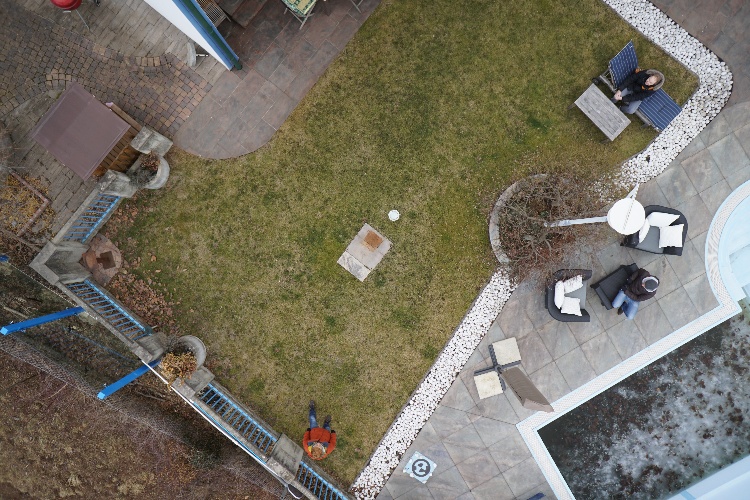}} &
\multicolumn{2}{c}{\includegraphics[trim=0mm 0mm 0mm 0mm,clip,width=0.4\linewidth]{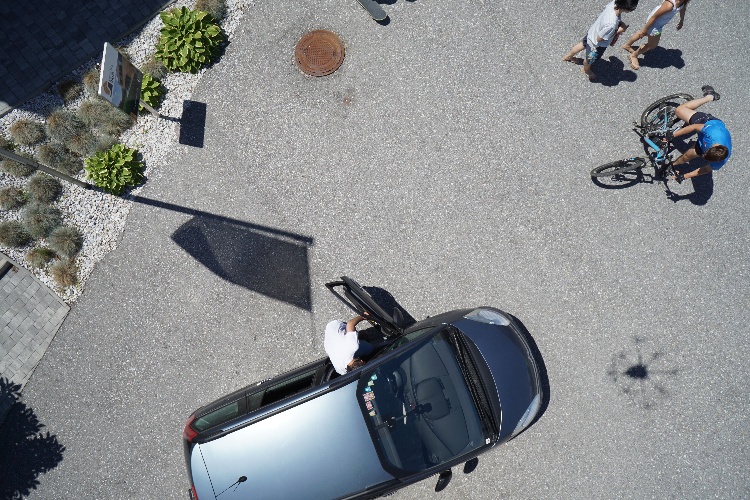}} \\
\multicolumn{6}{c}{\Large{ICG}} \\
\multicolumn{6}{c}{}\\\cline{2-2}\cline{4-4}\cline{6-6}
\raisebox{-.5\height}{\includegraphics[trim=0mm 0mm 0mm 0mm,clip,width=0.2\linewidth]{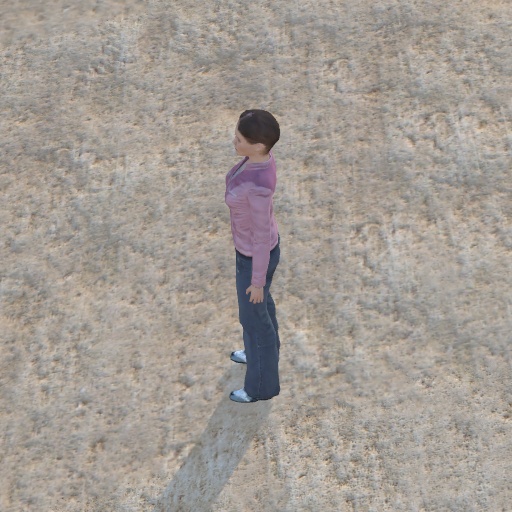}} & 
\makebox[2.7cm][l]{
\normalsize{
\begin{tabular}{l}
\\
\hspace{-0.5cm}$\cdot$ character: juliet \\
\hspace{-0.5cm}$\cdot$ pose: standing \\
\hspace{-0.5cm}$\cdot$ altitude: 5$m$ \\
\hspace{-0.5cm}$\cdot$ radius: 5$m$ \\
\hspace{-0.5cm}$\cdot$ camera angle: 90$^\circ$ \\
\hspace{-0.5cm}$\cdot$ sun angle: 1 \\\\
\end{tabular}
}}
& \raisebox{-.5\height}{\includegraphics[trim=0mm 0mm 0mm 0mm,clip,width=0.2\linewidth]{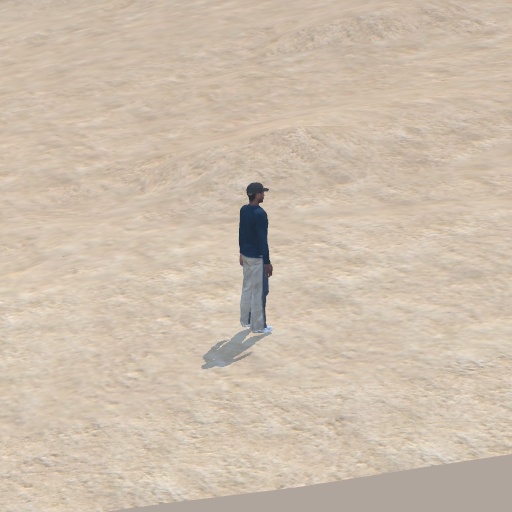}} & 
\makebox[2.7cm][l]{
\normalsize{
\begin{tabular}{l}
\\
\hspace{-0.5cm}$\cdot$ character: scott \\
\hspace{-0.5cm}$\cdot$ pose: standing \\
\hspace{-0.5cm}$\cdot$ altitude: 5$m$ \\
\hspace{-0.5cm}$\cdot$ radius: 15$m$ \\
\hspace{-0.5cm}$\cdot$ camera angle: 240$^\circ$ \\
\hspace{-0.5cm}$\cdot$ sun angle: 2 \\\\
\end{tabular}
}}
& \raisebox{-.5\height}{\includegraphics[trim=0mm 0mm 0mm 0mm,clip,width=0.2\linewidth]{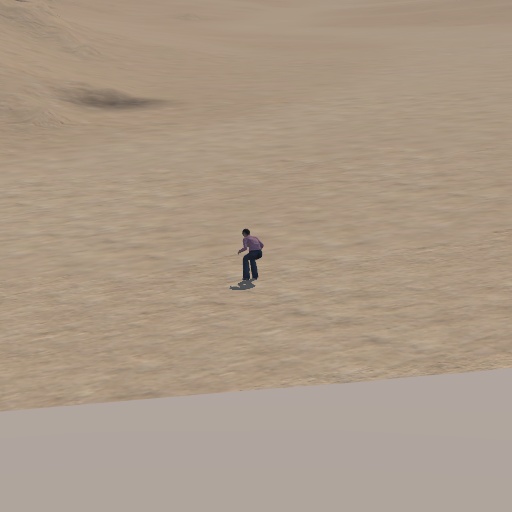}} & 
\makebox[2.7cm][l]{
\normalsize{
\begin{tabular}{l}
\\
\hspace{-0.5cm}$\cdot$ character: juliet \\
\hspace{-0.5cm}$\cdot$ pose: squatting \\
\hspace{-0.5cm}$\cdot$ altitude: 5$m$ \\
\hspace{-0.5cm}$\cdot$ radius: 30$m$ \\
\hspace{-0.5cm}$\cdot$ camera angle: 150$^\circ$ \\
\hspace{-0.5cm}$\cdot$ sun angle: 3 \\\\
\end{tabular}
}}\\\cline{2-2}\cline{4-4}\cline{6-6}
\multicolumn{6}{c}{}\\[-0.2cm]
\cline{2-2}\cline{4-4}\cline{6-6}
\raisebox{-.5\height}{\includegraphics[trim=0mm 0mm 0mm 0mm,clip,width=0.2\linewidth]{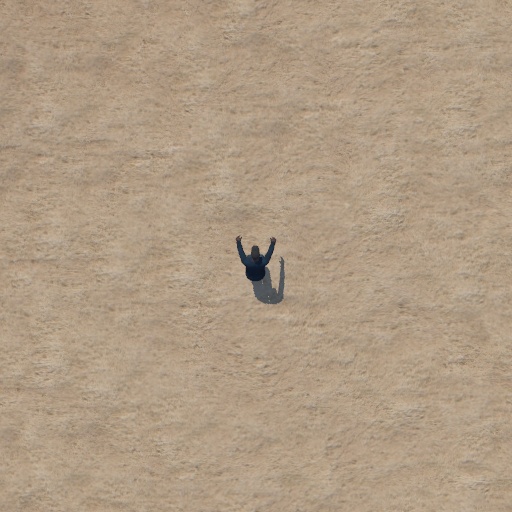}} & 
\makebox[2.7cm][l]{
\normalsize{
\begin{tabular}{l}
\\
\hspace{-0.5cm}$\cdot$ character: scott \\
\hspace{-0.5cm}$\cdot$ pose: squatting \\
\hspace{-0.5cm}$\cdot$ altitude: 25$m$ \\
\hspace{-0.5cm}$\cdot$ radius: 5$m$ \\
\hspace{-0.5cm}$\cdot$ camera angle: 180$^\circ$ \\
\hspace{-0.5cm}$\cdot$ sun angle: 2 \\\\
\end{tabular}
}}
& \raisebox{-.5\height}{\includegraphics[trim=0mm 0mm 0mm 0mm,clip,width=0.2\linewidth]{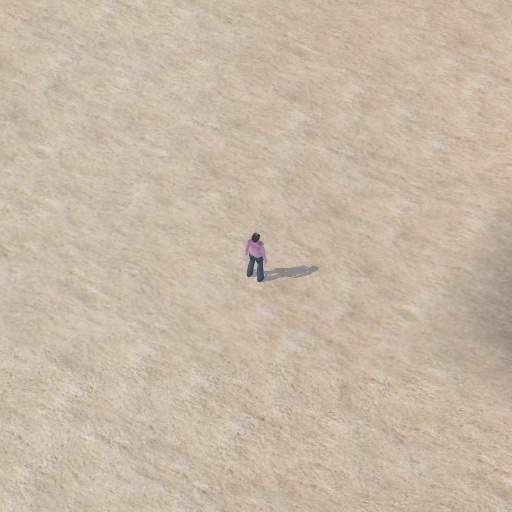}} &
\makebox[2.7cm][l]{
\normalsize{
\begin{tabular}{l}
\\
\hspace{-0.5cm}$\cdot$ character: juliet \\
\hspace{-0.5cm}$\cdot$ pose: standing \\
\hspace{-0.5cm}$\cdot$ altitude: 25$m$ \\
\hspace{-0.5cm}$\cdot$ radius: 15$m$ \\
\hspace{-0.5cm}$\cdot$ camera angle: 210$^\circ$ \\
\hspace{-0.5cm}$\cdot$ sun angle: 3 \\\\
\end{tabular}
}}
& \raisebox{-.5\height}{\includegraphics[trim=0mm 0mm 0mm 0mm,clip,width=0.2\linewidth]{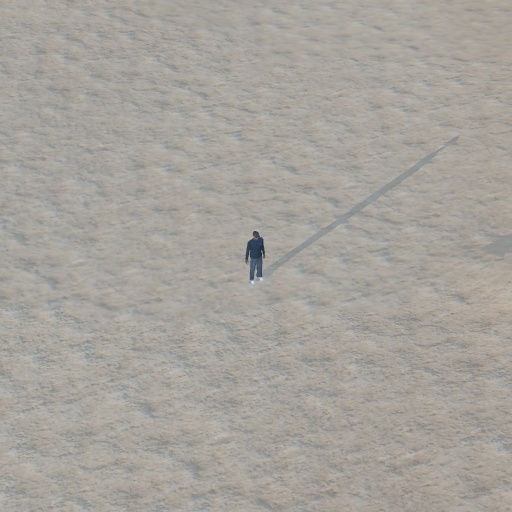}} &
\makebox[2.7cm][l]{
\normalsize{
\begin{tabular}{l}
\\
\hspace{-0.5cm}$\cdot$ character: scott \\
\hspace{-0.5cm}$\cdot$ pose: standing \\
\hspace{-0.5cm}$\cdot$ altitude: 25$m$ \\
\hspace{-0.5cm}$\cdot$ radius: 30$m$ \\
\hspace{-0.5cm}$\cdot$ camera angle: 330$^\circ$ \\
\hspace{-0.5cm}$\cdot$ sun angle: 4 \\\\
\end{tabular}
}}\\\cline{2-2}\cline{4-4}\cline{6-6}
\multicolumn{6}{c}{}\\[-0.2cm]
\cline{2-2}\cline{4-4}\cline{6-6}
\raisebox{-.5\height}{\includegraphics[trim=0mm 0mm 0mm 0mm,clip,width=0.2\linewidth]{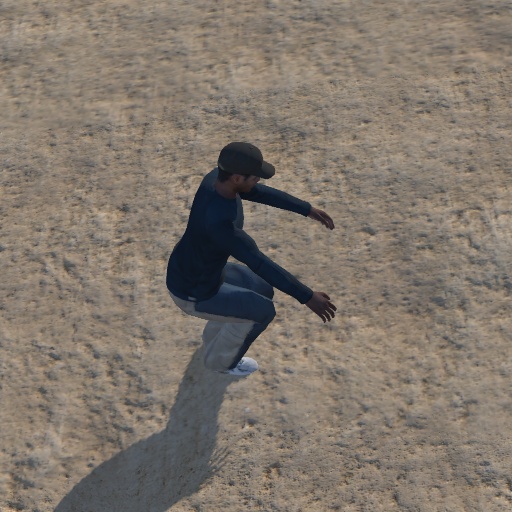}} & 
\makebox[2.7cm][l]{
\normalsize{
\begin{tabular}{l}
\\
\hspace{-0.5cm}$\cdot$ character: scott \\
\hspace{-0.5cm}$\cdot$ pose: squatting \\
\hspace{-0.5cm}$\cdot$ altitude: 5$m$ \\
\hspace{-0.5cm}$\cdot$ radius: 5$m$ \\
\hspace{-0.5cm}$\cdot$ camera angle: 270$^\circ$ \\
\hspace{-0.5cm}$\cdot$ sun angle: 1 \\\\
\end{tabular}
}}
& \raisebox{-.5\height}{\includegraphics[trim=0mm 0mm 0mm 0mm,clip,width=0.2\linewidth]{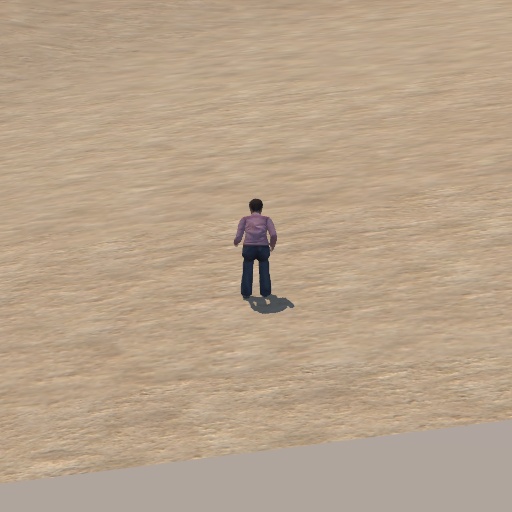}} & 
\makebox[2.7cm][l]{
\normalsize{
\begin{tabular}{l}
\\
\hspace{-0.5cm}$\cdot$ character: juliet \\
\hspace{-0.5cm}$\cdot$ pose: squatting \\
\hspace{-0.5cm}$\cdot$ altitude: 5$m$ \\
\hspace{-0.5cm}$\cdot$ radius: 15$m$ \\
\hspace{-0.5cm}$\cdot$ camera angle: 180$^\circ$ \\
\hspace{-0.5cm}$\cdot$ sun angle: 2 \\\\
\end{tabular}
}}
& \raisebox{-.5\height}{\includegraphics[trim=0mm 0mm 0mm 0mm,clip,width=0.2\linewidth]{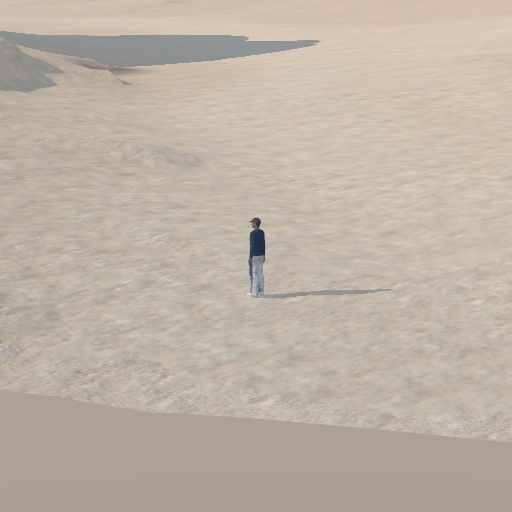}} & 
\makebox[2.7cm][l]{
\normalsize{
\begin{tabular}{l}
\\
\hspace{-0.5cm}$\cdot$ character: scott \\
\hspace{-0.5cm}$\cdot$ pose: standing \\
\hspace{-0.5cm}$\cdot$ altitude: 5$m$ \\
\hspace{-0.5cm}$\cdot$ radius: 30$m$ \\
\hspace{-0.5cm}$\cdot$ camera angle: 120$^\circ$ \\
\hspace{-0.5cm}$\cdot$ sun angle: 1 \\\\
\end{tabular}
}}\\\cline{2-2}\cline{4-4}\cline{6-6}
\multicolumn{6}{c}{}\\[-0.2cm]
\cline{2-2}\cline{4-4}\cline{6-6}
\raisebox{-.5\height}{\includegraphics[trim=0mm 0mm 0mm 0mm,clip,width=0.2\linewidth]{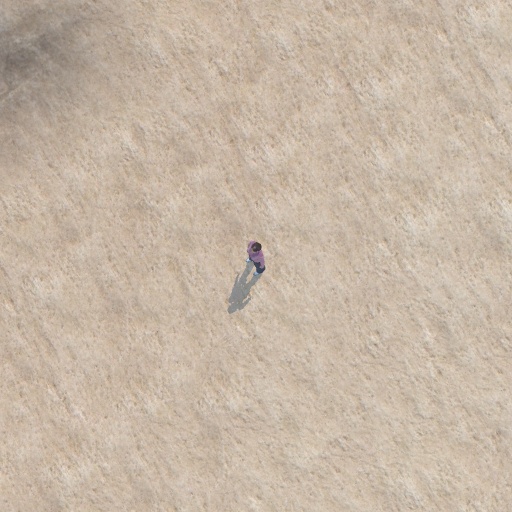}} & 
\makebox[2.7cm][l]{
\normalsize{
\begin{tabular}{l}
\\
\hspace{-0.5cm}$\cdot$ character: juliet \\
\hspace{-0.5cm}$\cdot$ pose: standing \\
\hspace{-0.5cm}$\cdot$ altitude: 25$m$ \\
\hspace{-0.5cm}$\cdot$ radius: 5$m$ \\
\hspace{-0.5cm}$\cdot$ camera angle: 60$^\circ$ \\
\hspace{-0.5cm}$\cdot$ sun angle: 2 \\\\
\end{tabular}
}}
& \raisebox{-.5\height}{\includegraphics[trim=0mm 0mm 0mm 0mm,clip,width=0.2\linewidth]{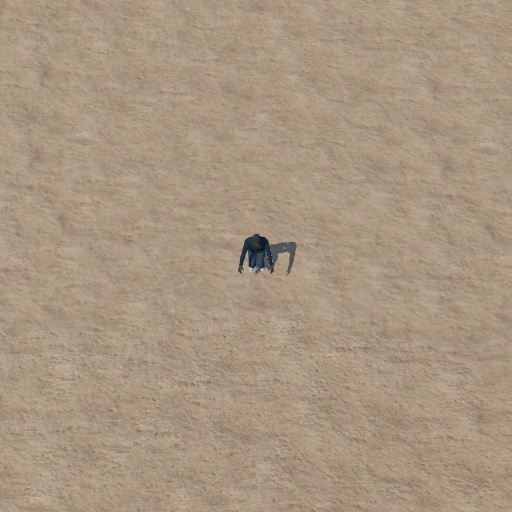}} &
\makebox[2.7cm][l]{
\normalsize{
\begin{tabular}{l}
\\
\hspace{-0.5cm}$\cdot$ character: scott \\
\hspace{-0.5cm}$\cdot$ pose: squatting \\
\hspace{-0.5cm}$\cdot$ altitude: 25$m$ \\
\hspace{-0.5cm}$\cdot$ radius: 15$m$ \\
\hspace{-0.5cm}$\cdot$ camera angle: 0$^\circ$ \\
\hspace{-0.5cm}$\cdot$ sun angle: 3 \\\\
\end{tabular}
}}
& \raisebox{-.5\height}{\includegraphics[trim=0mm 0mm 0mm 0mm,clip,width=0.2\linewidth]{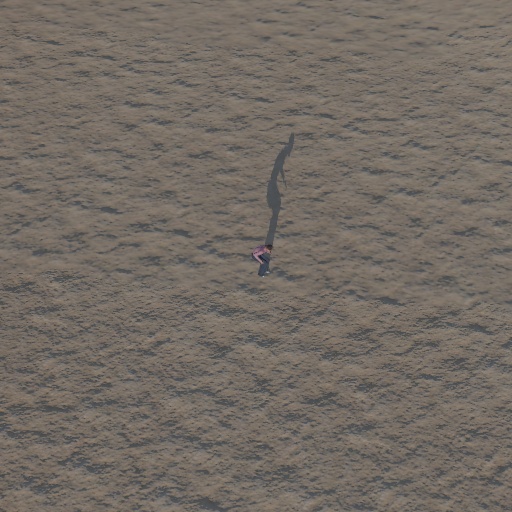}} &
\makebox[2.7cm][l]{
\normalsize{
\begin{tabular}{l}
\\
\hspace{-0.5cm}$\cdot$ character: juliet \\
\hspace{-0.5cm}$\cdot$ pose: squatting \\
\hspace{-0.5cm}$\cdot$ altitude: 25$m$ \\
\hspace{-0.5cm}$\cdot$ radius: 30$m$ \\
\hspace{-0.5cm}$\cdot$ camera angle: 300$^\circ$ \\
\hspace{-0.5cm}$\cdot$ sun angle: 4 \\\\
\end{tabular}
}}\\\cline{2-2}\cline{4-4}\cline{6-6}
\multicolumn{6}{c}{}\\[-0.2cm]
\multicolumn{6}{c}{\Large{Archangel-Synthetic}} \\
\end{tabular}
}
\caption{{\bf Sample images from each dataset.} For the Archangel-Synthetic dataset, metadata for each image is shown to the right of the image.}
\label{fig:samples_dataset}
\end{figure*}

\noindent{\bf Network architecture.} To obtain a generator, we adopt CycleGAN, where two generators and two discriminators are involved during model training. For each generator, we use a 24-layer UNet-like architecture (i.e., the \texttt{resnet\_9blocks} generator in the official repository of CycleGAN~\cite{CycleGANlink}), which contains nine 2-layer residual modules in the middle of the architecture where the encoder and decoder are connected. For each discriminator, a 5-layer fully convolutional network (i.e., the \texttt{basic} discriminator in the official repository of CycleGAN~\cite{CycleGANlink}) is used.

For the detector, we adopt the official RetinaNet architecture implemented in Detectron2~\cite{YWUDetectron2} with few modifications. First of all, the feature dimension of RetinaNet's classification subnet and box regression subnet are decreased from 256 to 32 since we are dealing with single category (i.e., human) instead of multiple categories (e.g., 80 categories for detectors trained on the MS COCO dataset). In addition, we switch the activation function used in these two subnets from ReLU to LeakyReLU to avoid the singular covariance matrix problem, which may occur when calculating the Mahalanobis distance, occasionally triggered by the dying ReLU problem. Finally, the kernel size of the last convolutional layer in the classification subnet, the layer just in front of the sigmoid layer, is reduced from 3$\times$3 to 1$\times$1 so that each output prediction is associated with only one feature vector in the feature representation space. Here, ResNet50 is used as the backbone.\smallskip

\noindent{\bf Training details.} The settings used for training the human detector through our method and the other baseline methods are listed in Table~\ref{tab:detector_training}. Without further specification, we follow all the settings and initialization strategies defined by the original RetinaNet training~\cite{TLinICCV2017}. Unlike other model training, using fewer iterations for fine-tuning when adopting the pretrain-finetune method is common because training converges faster. Although fine-tuning usually uses a lower learning rate than pre-training, we use the same learning rate (i.e., 0.001) for both stages because we also fine-tune the ImageNet-pretrained backbone on the virtual images in the pre-training stage.

The settings used for training the generator through the baseline method using transformation (i.e., `naive merge w/ transformation') and our method are listed in Table~\ref{tab:generator_training}. These settings and initialization strategies are taken from the original CycleGAN training~\cite{JZhuICCV2017}. `Naive merge w/ transform' used fewer training epochs than PTL (80 vs 100) because all images in the virtual set are used for training. In general, as the size of the dataset increases, the number of epochs required for training decreases.\smallskip

\noindent{\bf Datasets.} In this section, we provide the details of the three real UAV-based datasets, VisDrone~\cite{PZhuTPAMI2021}, Okutama-Action~\cite{MBerkatainCVPRW2017}, and ICG~\cite{ICGlink}), and the virtual UAV-based dataset, \emph{Archangel-Synthetic}, used in this paper. Sample images for each dataset are shown in Figure~\ref{fig:samples_dataset}.

{\bf VisDrone} has four tracks (i.e., object detection (DET), video object detection (VID), single object tracking (SOT), and multi-object tracking (MOT)) with a separate dataset for each track. We use the dataset from the DET track and focus on detecting the person and the pedestrian categories among the ten object categories covered by the track. The VisDrone dataset in the DET track consists of 10,209 images, where 6,471 images are used in the \texttt{training} set, 548 images are used in the \texttt{validation} set, 1,580 images are used in the \texttt{test-challenge} set, and 1,610 images are used in the \texttt{test-dev} set. We use the \texttt{training} set for model training and the \texttt{test-dev} set for model testing. The maximal resolution of images in the VisDrone dataset is 2000$\times$1500.

\begin{figure*}[t]
\centering
\resizebox{.75\linewidth}{!}{%
\setlength{\tabcolsep}{2.0pt}
\begin{tabular}{ccccc}
\scriptsize{{\bf iter. 1}} & \scriptsize{{\bf iter. 2}} & \scriptsize{{\bf iter. 3}} & \scriptsize{{\bf iter. 4}} & \scriptsize{{\bf iter. 5}}\\
\includegraphics[trim=0mm 0mm 0mm 0mm,clip,width=0.185\linewidth]{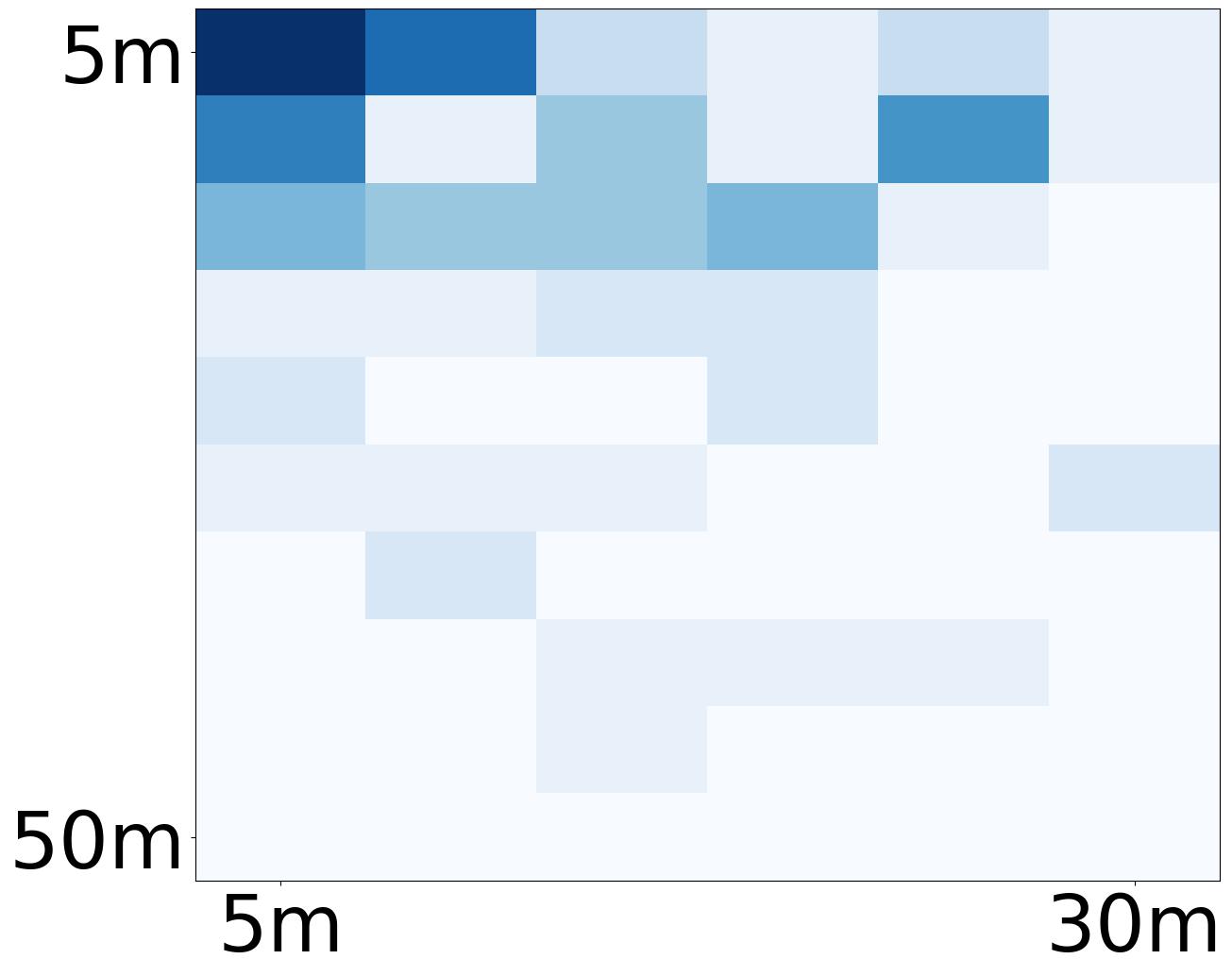} &
\includegraphics[trim=0mm 0mm 0mm 0mm,clip,width=0.185\linewidth]{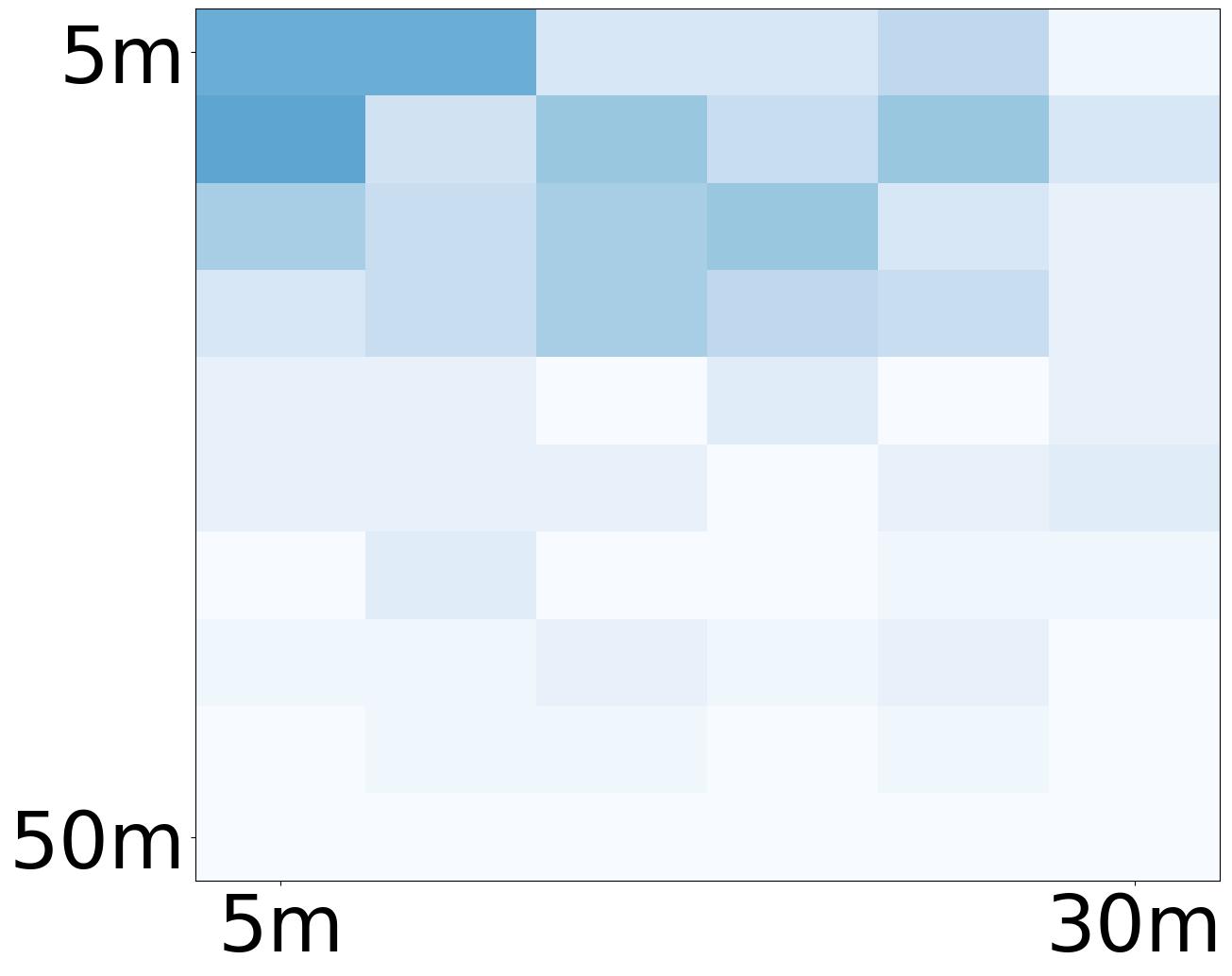} &
\includegraphics[trim=0mm 0mm 0mm 0mm,clip,width=0.185\linewidth]{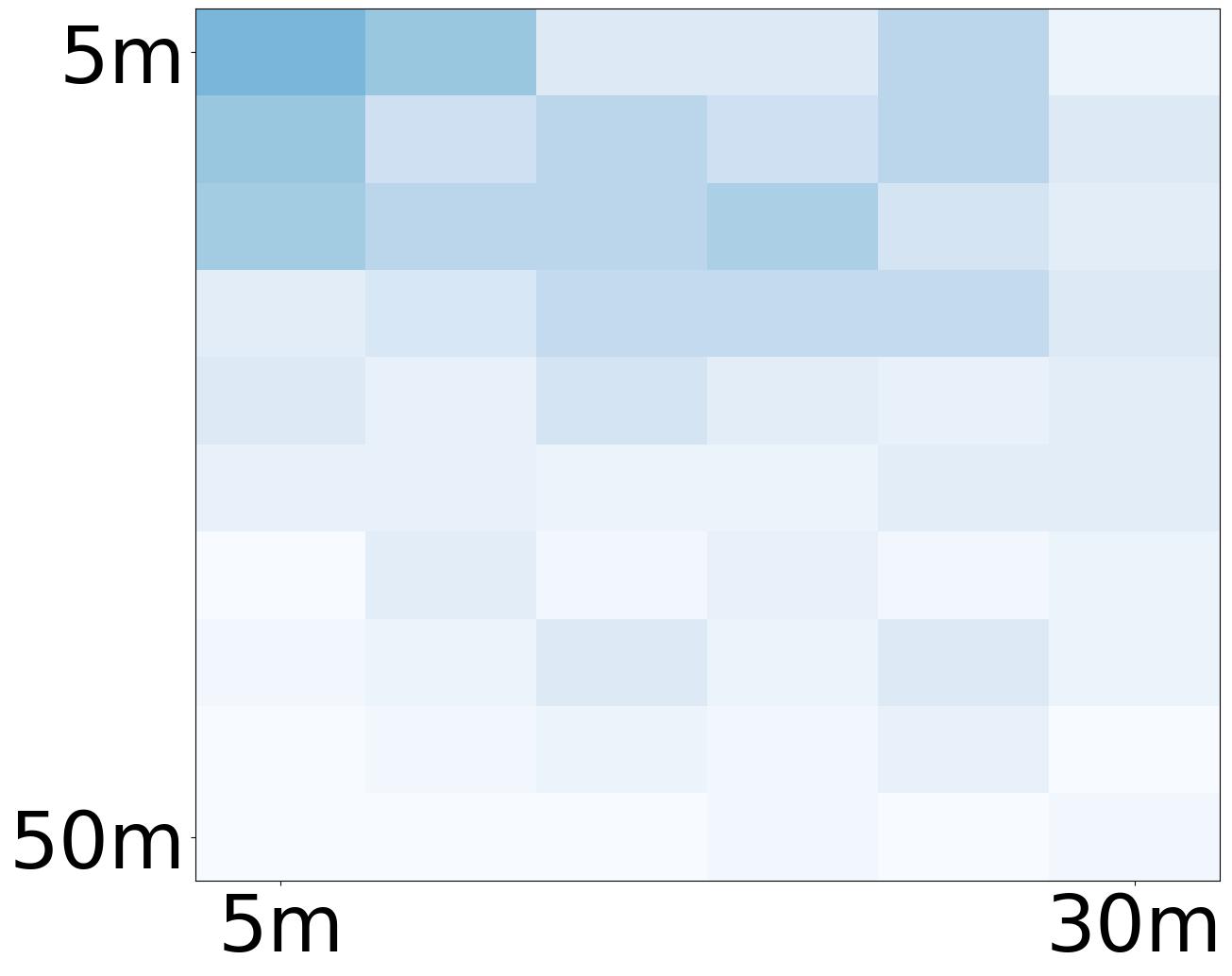} &
\includegraphics[trim=0mm 0mm 0mm 0mm,clip,width=0.185\linewidth]{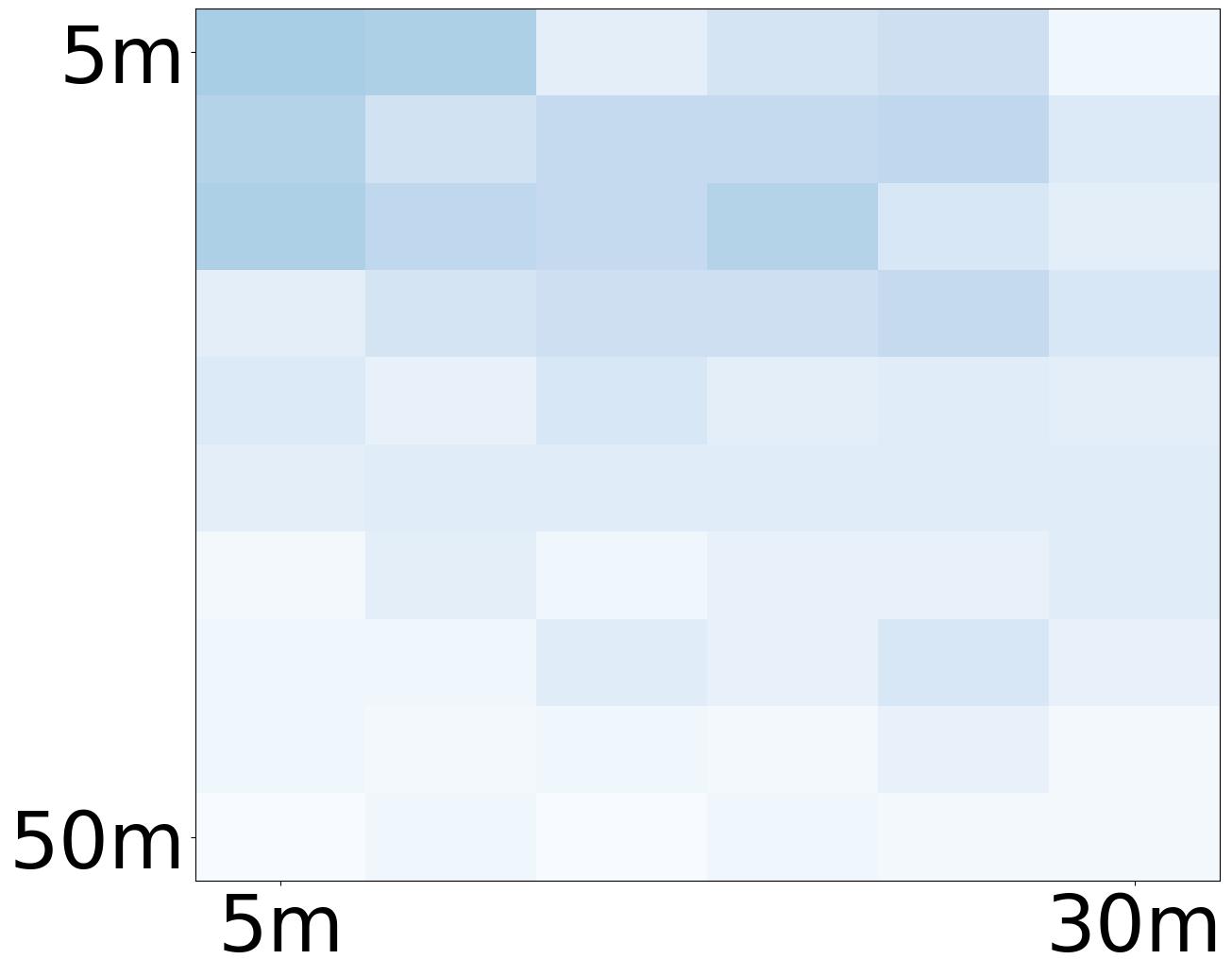} &
\includegraphics[trim=0mm 0mm 0mm 0mm,clip,width=0.185\linewidth]{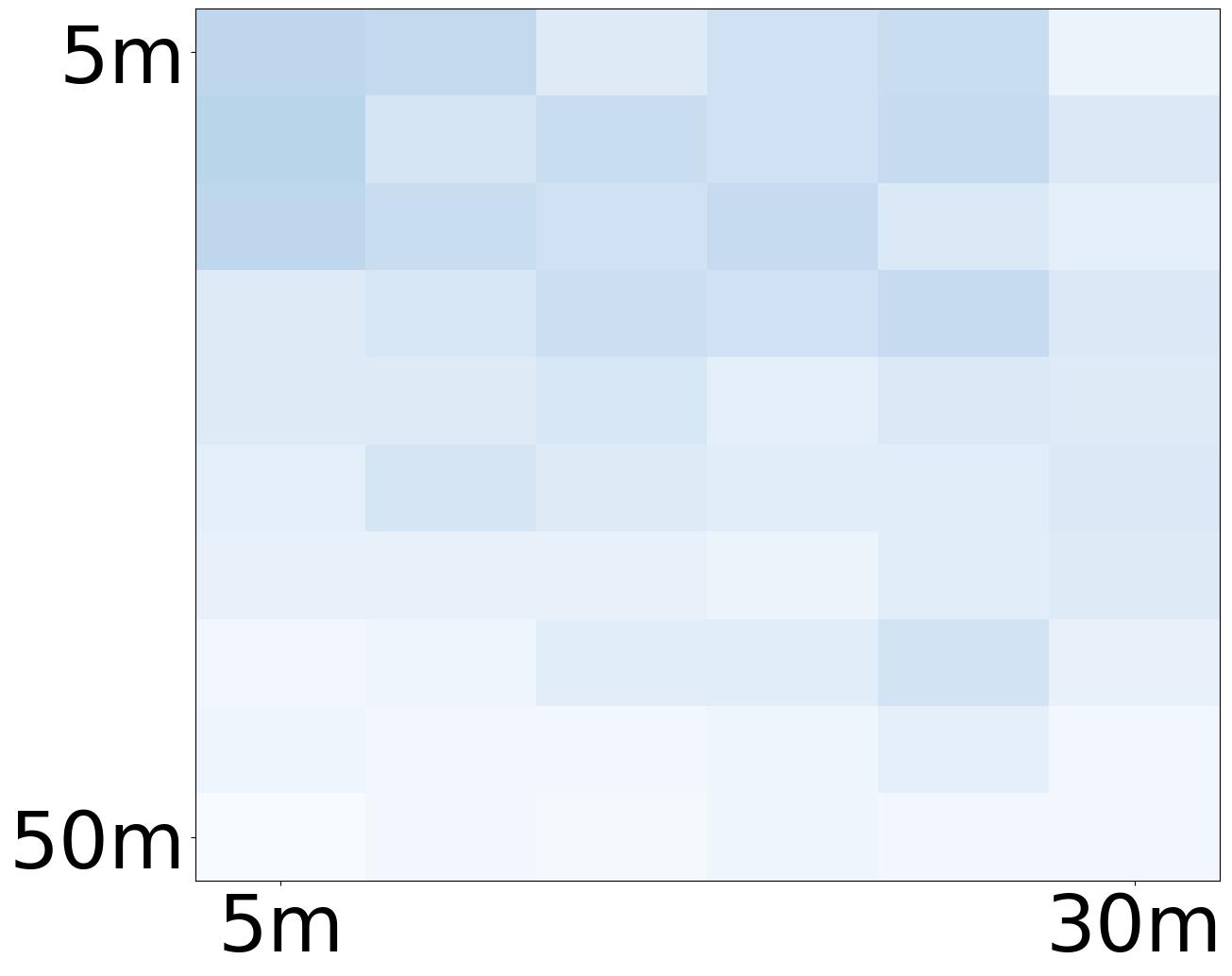} \\
\multicolumn{5}{c}{VisDrone, 20-img}\\
\scriptsize{{\bf iter. 1}} & \scriptsize{{\bf iter. 2}} & \scriptsize{{\bf iter. 3}} & \scriptsize{{\bf iter. 4}} & \scriptsize{{\bf iter. 5}}\\
\includegraphics[trim=0mm 0mm 0mm 0mm,clip,width=0.185\linewidth]{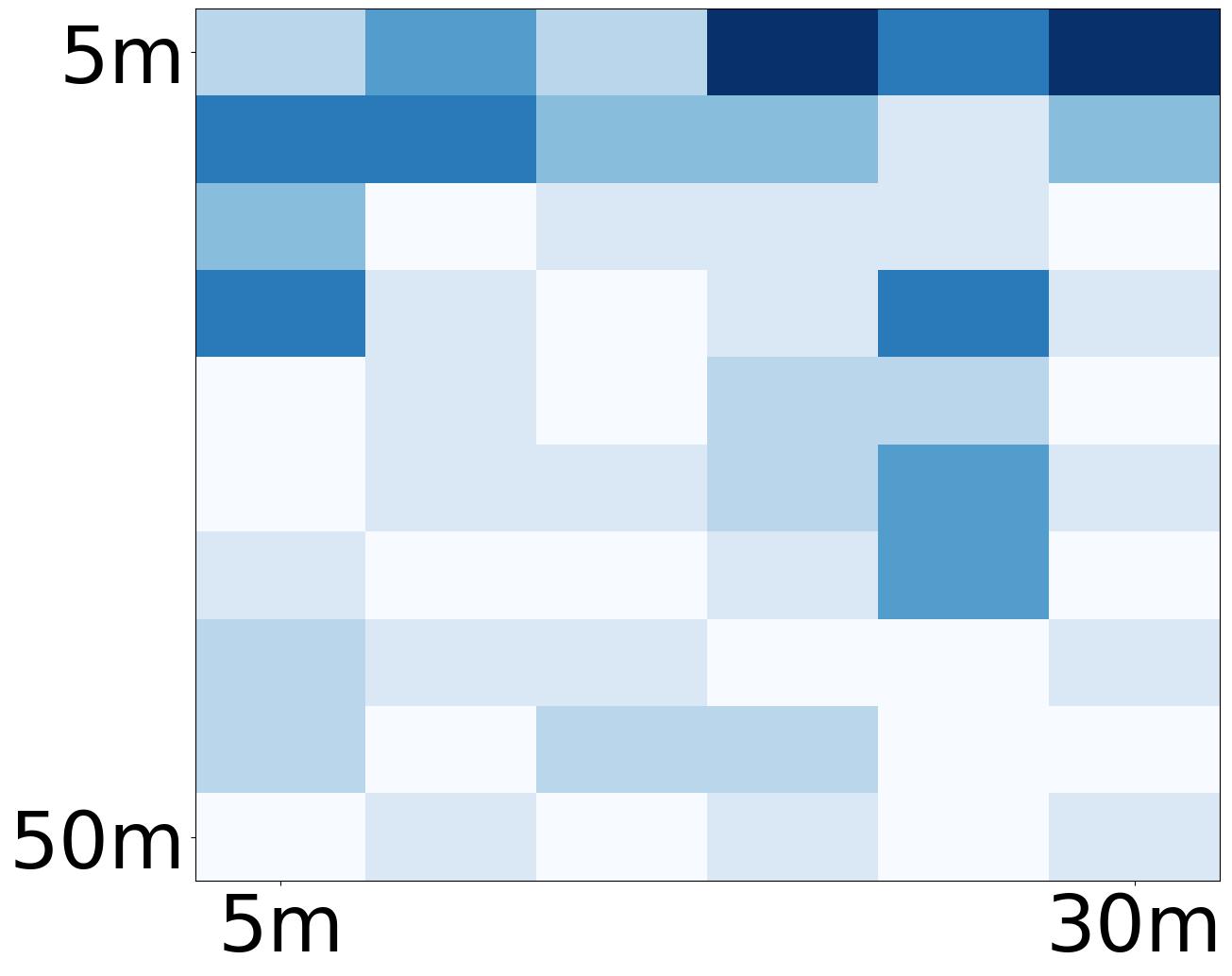} &
\includegraphics[trim=0mm 0mm 0mm 0mm,clip,width=0.185\linewidth]{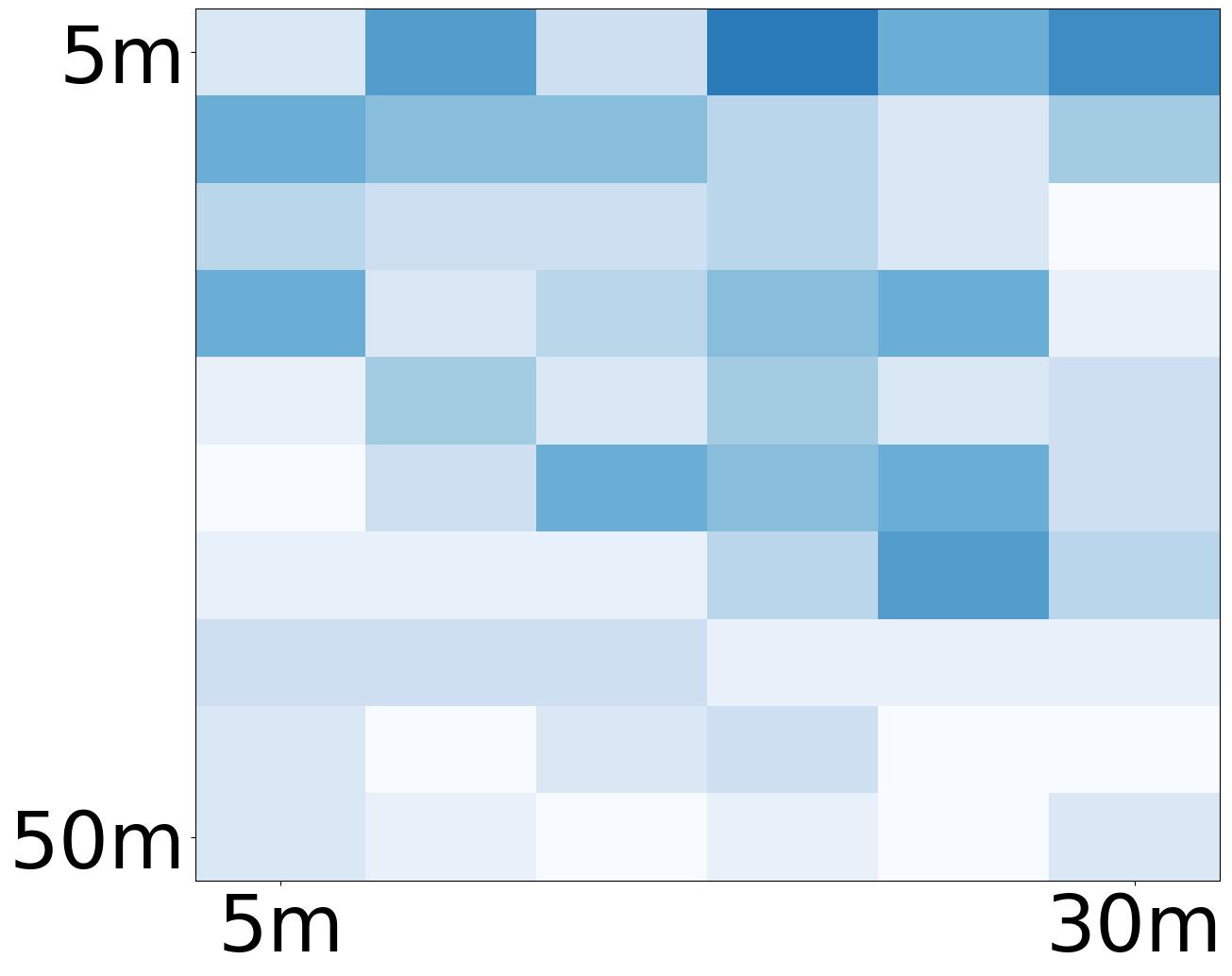} &
\includegraphics[trim=0mm 0mm 0mm 0mm,clip,width=0.185\linewidth]{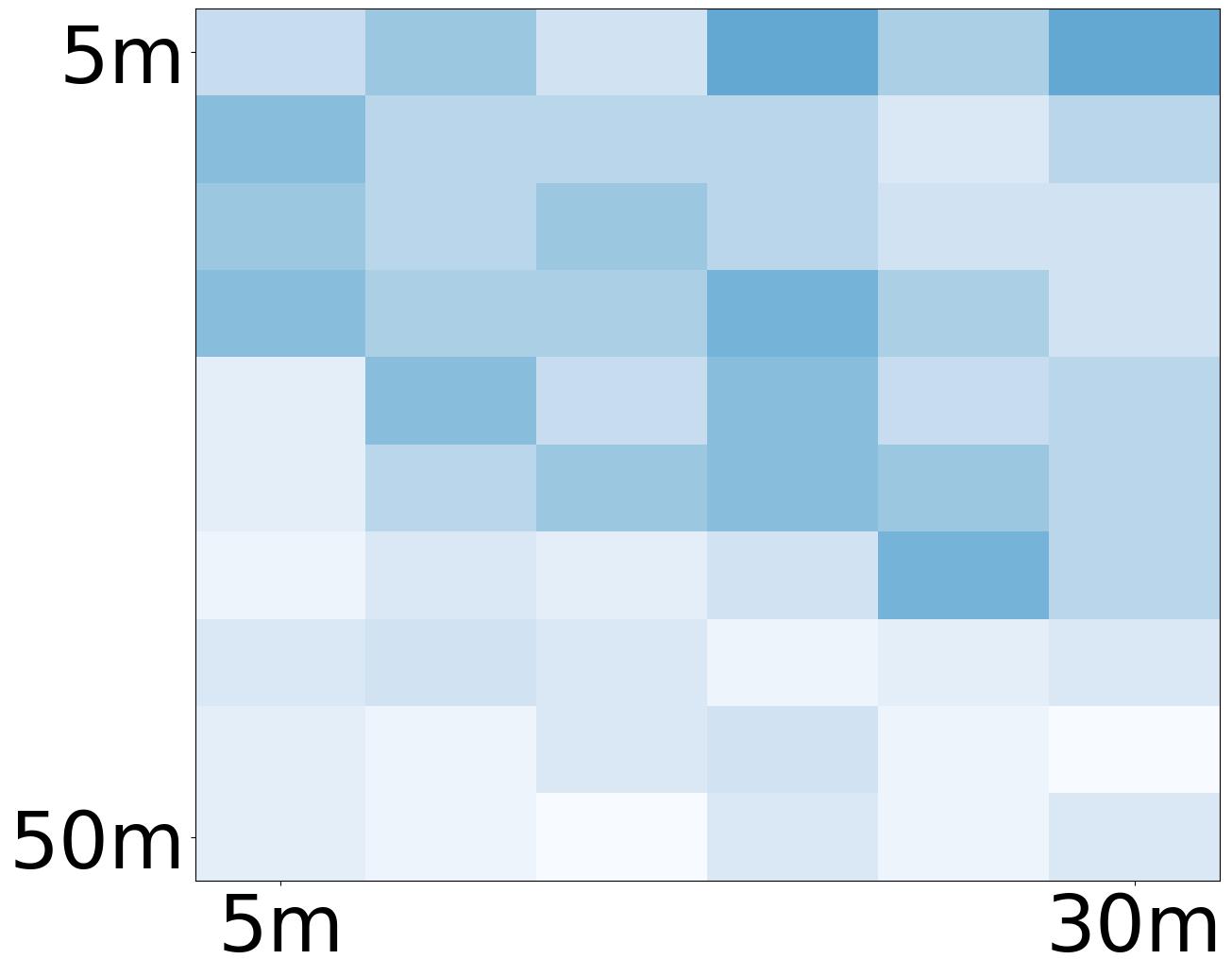} &
\includegraphics[trim=0mm 0mm 0mm 0mm,clip,width=0.185\linewidth]{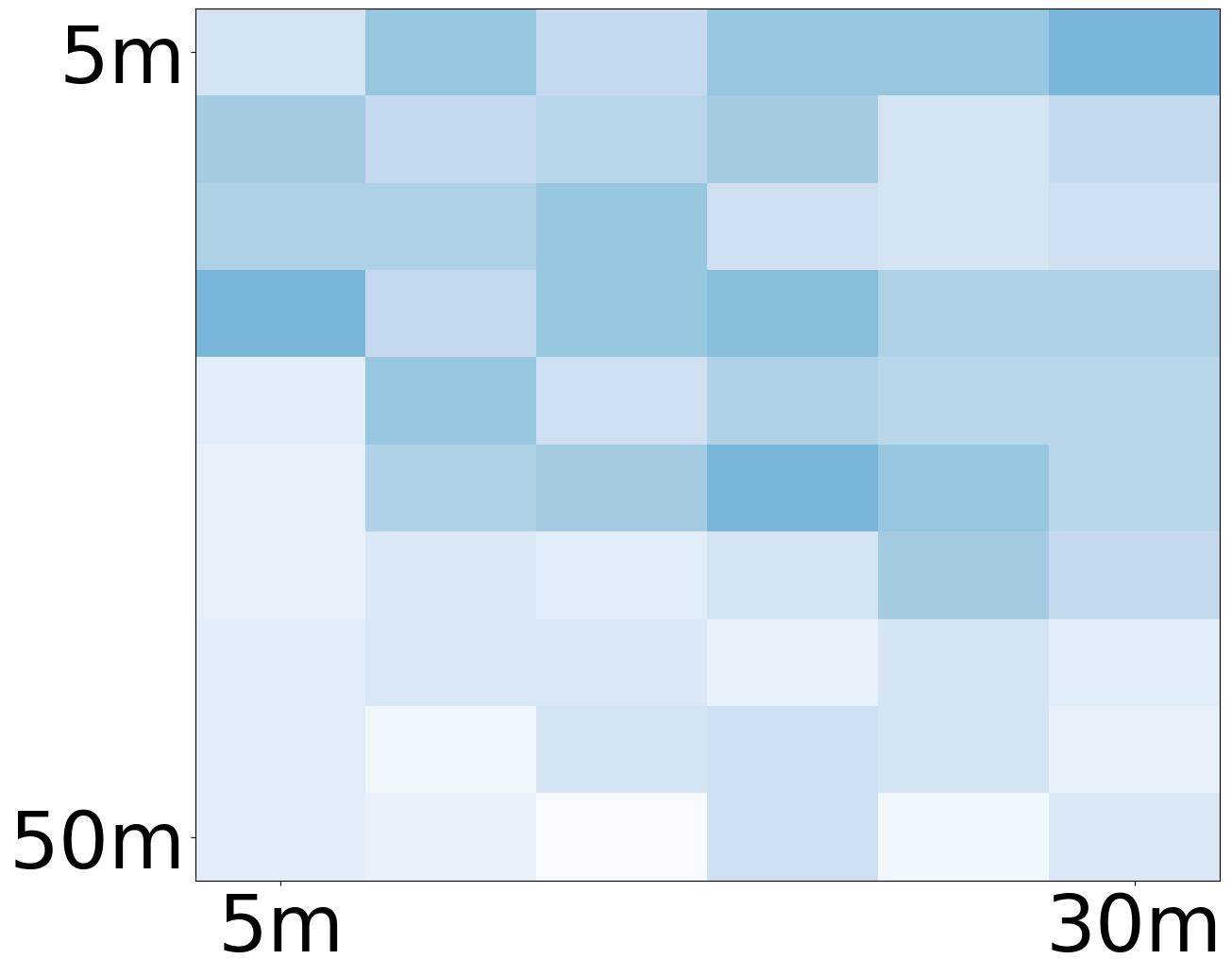} &
\includegraphics[trim=0mm 0mm 0mm 0mm,clip,width=0.185\linewidth]{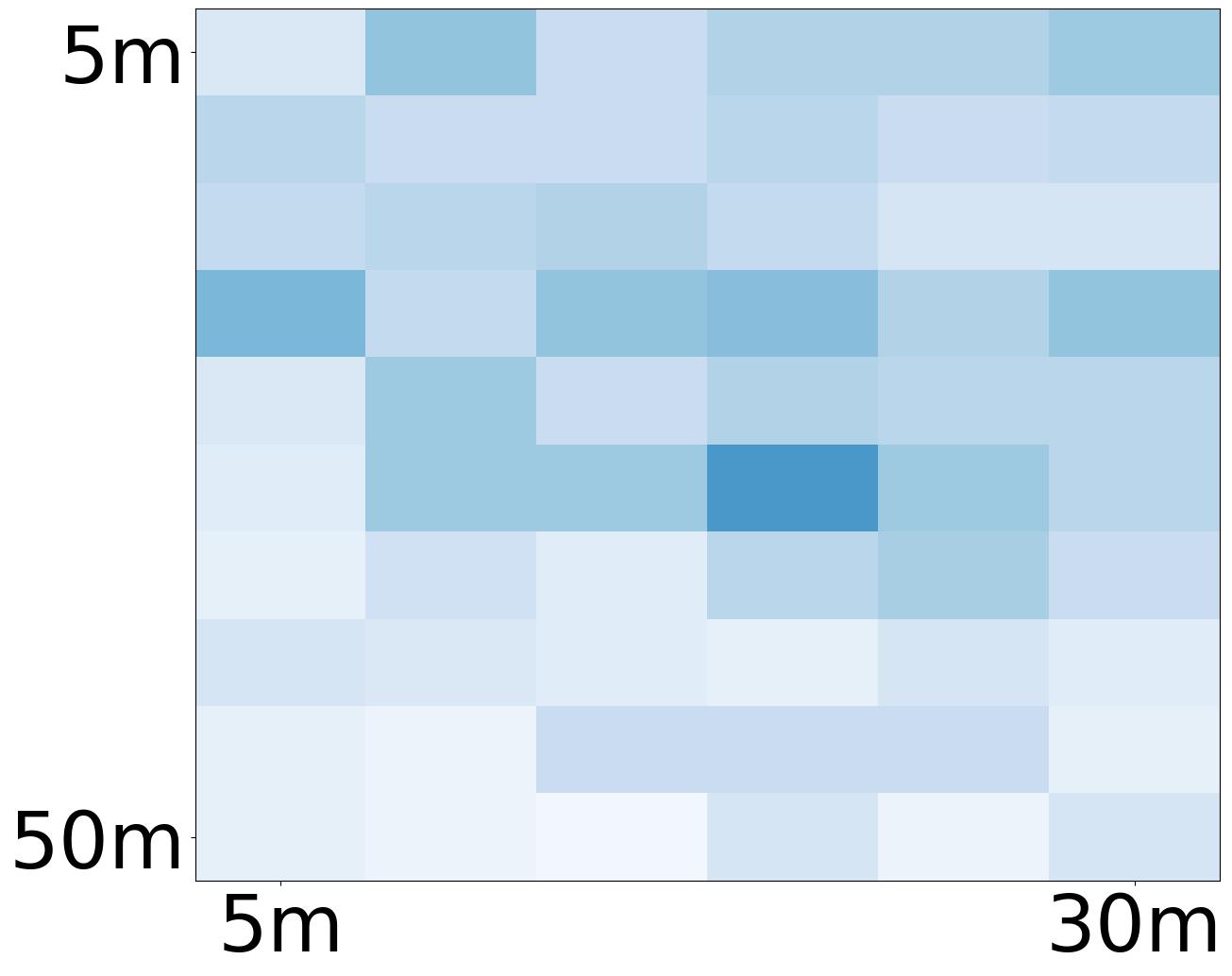} \\
\multicolumn{5}{c}{Okutama-Action, 20-img}\\
\scriptsize{{\bf iter. 1}} & \scriptsize{{\bf iter. 2}} & \scriptsize{{\bf iter. 3}} & \scriptsize{{\bf iter. 4}} & \scriptsize{{\bf iter. 5}}\\
\includegraphics[trim=0mm 0mm 0mm 0mm,clip,width=0.185\linewidth]{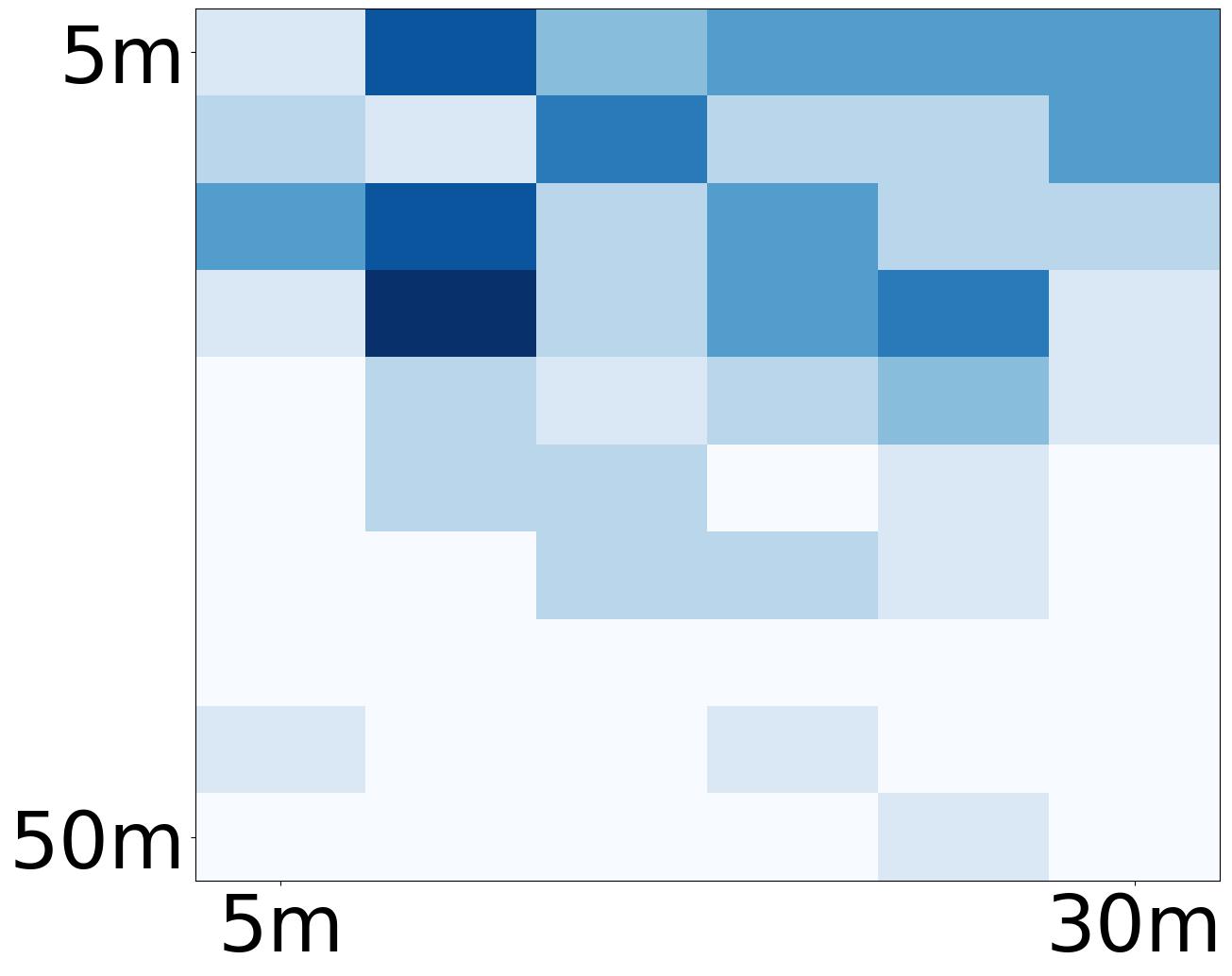} &
\includegraphics[trim=0mm 0mm 0mm 0mm,clip,width=0.185\linewidth]{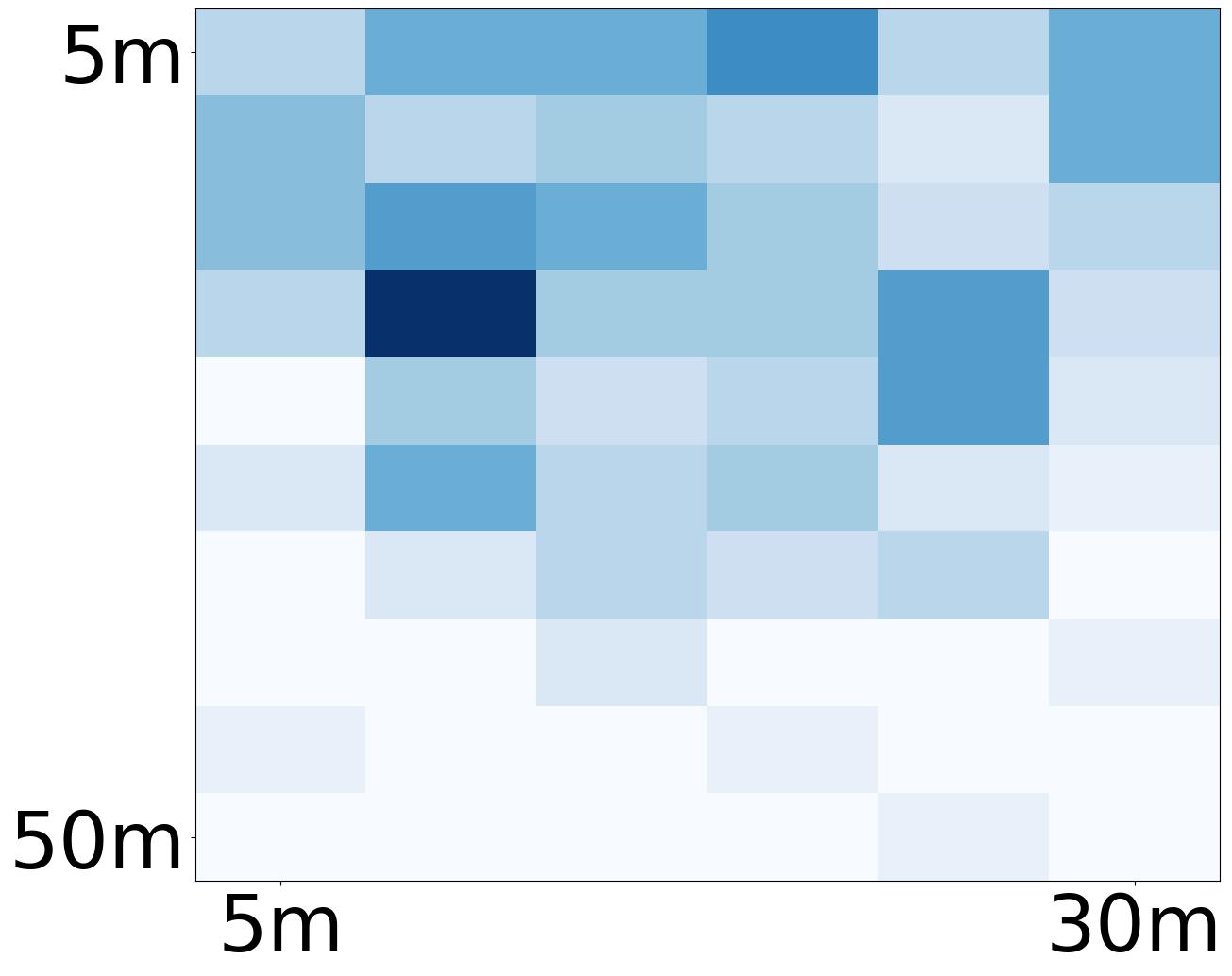} &
\includegraphics[trim=0mm 0mm 0mm 0mm,clip,width=0.185\linewidth]{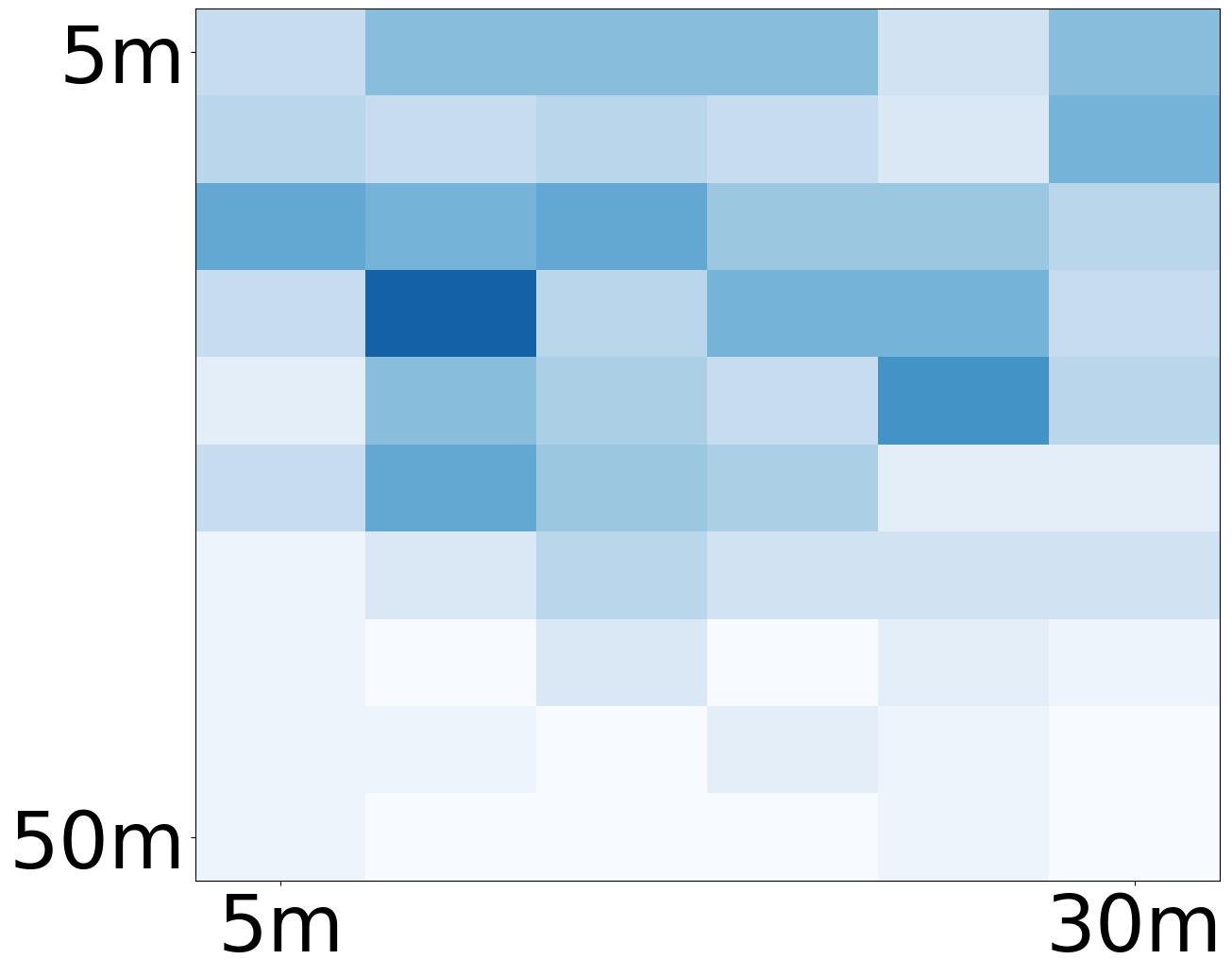} &
\includegraphics[trim=0mm 0mm 0mm 0mm,clip,width=0.185\linewidth]{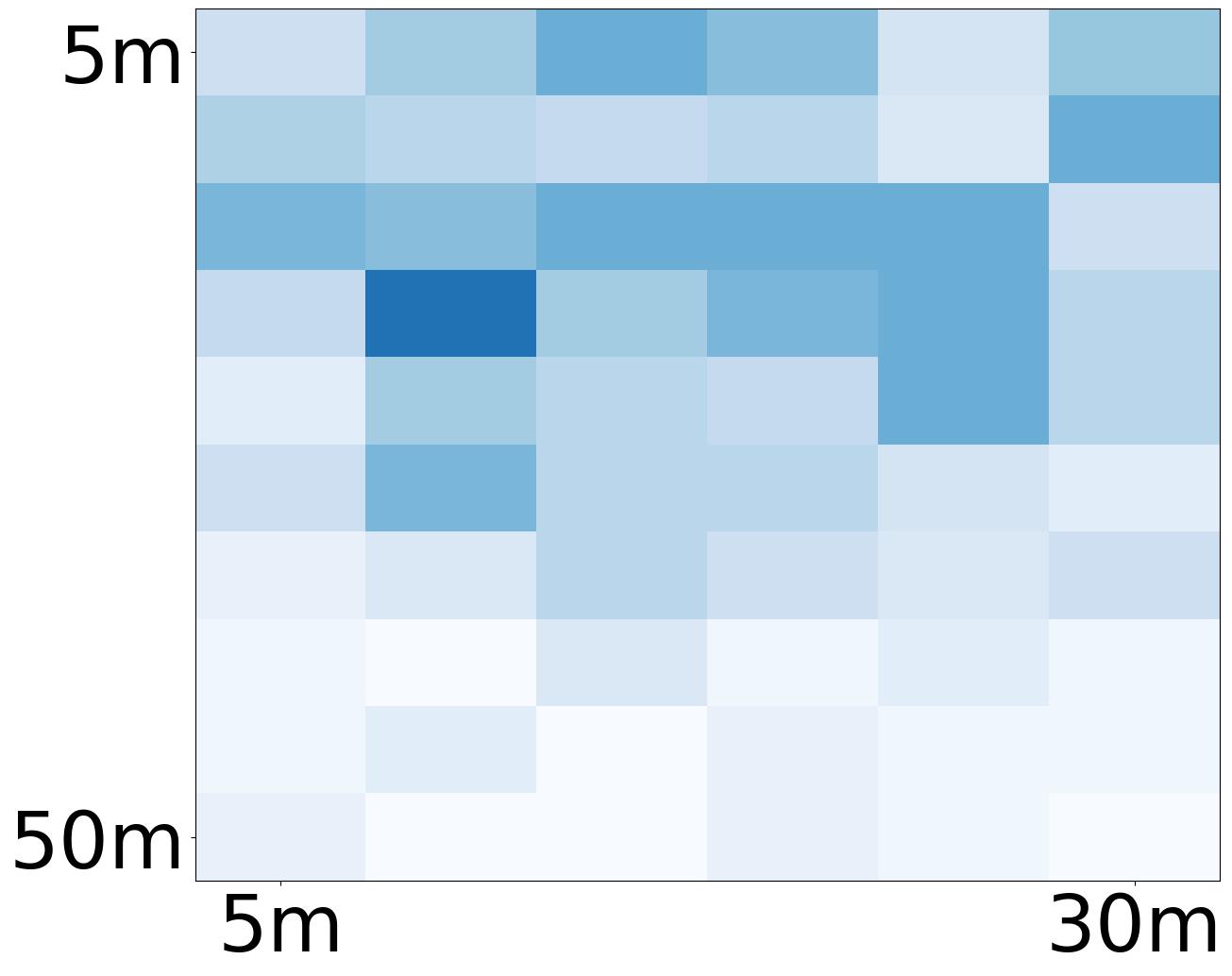} &
\includegraphics[trim=0mm 0mm 0mm 0mm,clip,width=0.185\linewidth]{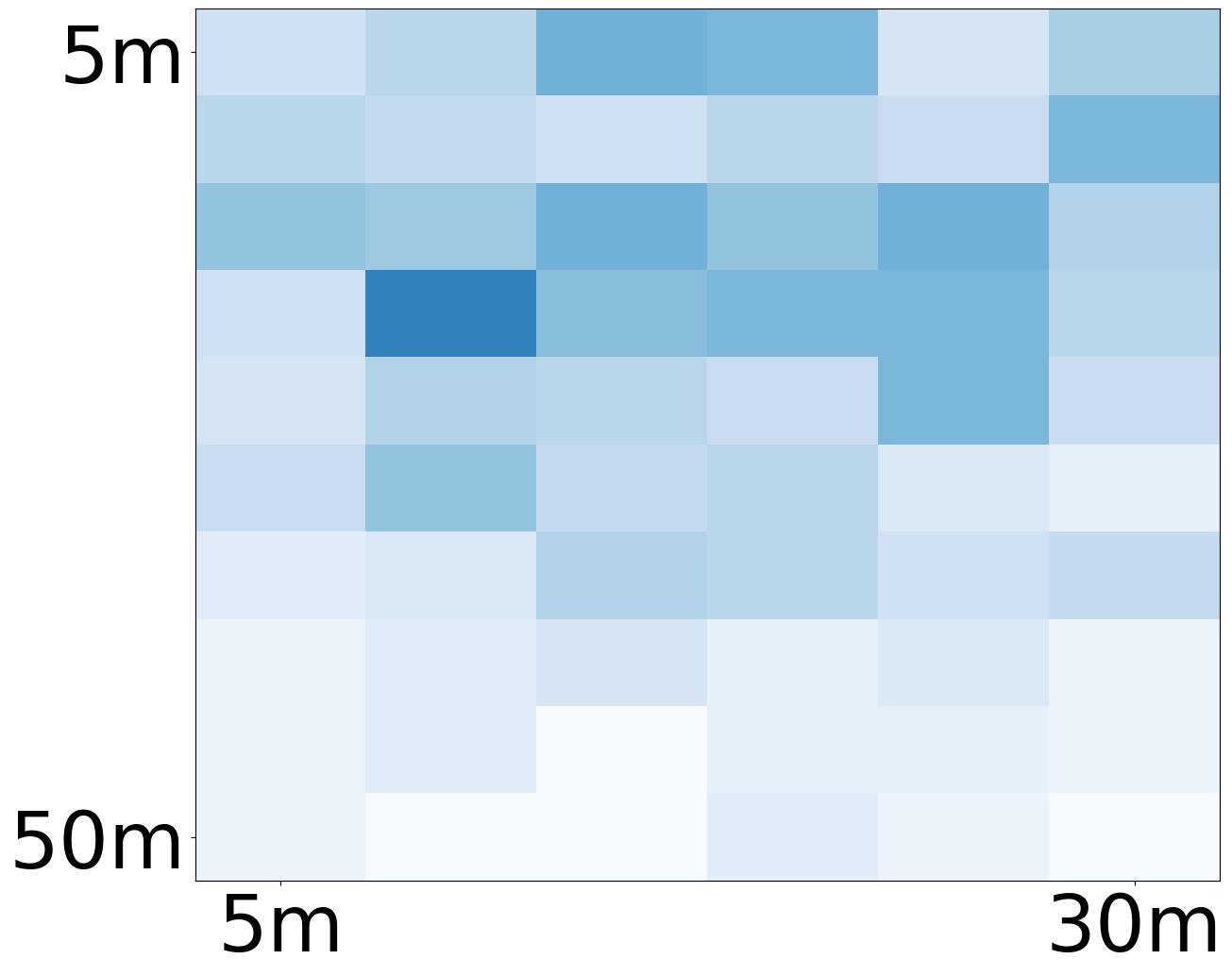} \\
\multicolumn{5}{c}{Okutama-Action, 50-img}\\
\scriptsize{{\bf iter. 1}} & \scriptsize{{\bf iter. 2}} & \scriptsize{{\bf iter. 3}} & \scriptsize{{\bf iter. 4}} & \scriptsize{{\bf iter. 5}}\\
\includegraphics[trim=0mm 0mm 0mm 0mm,clip,width=0.185\linewidth]{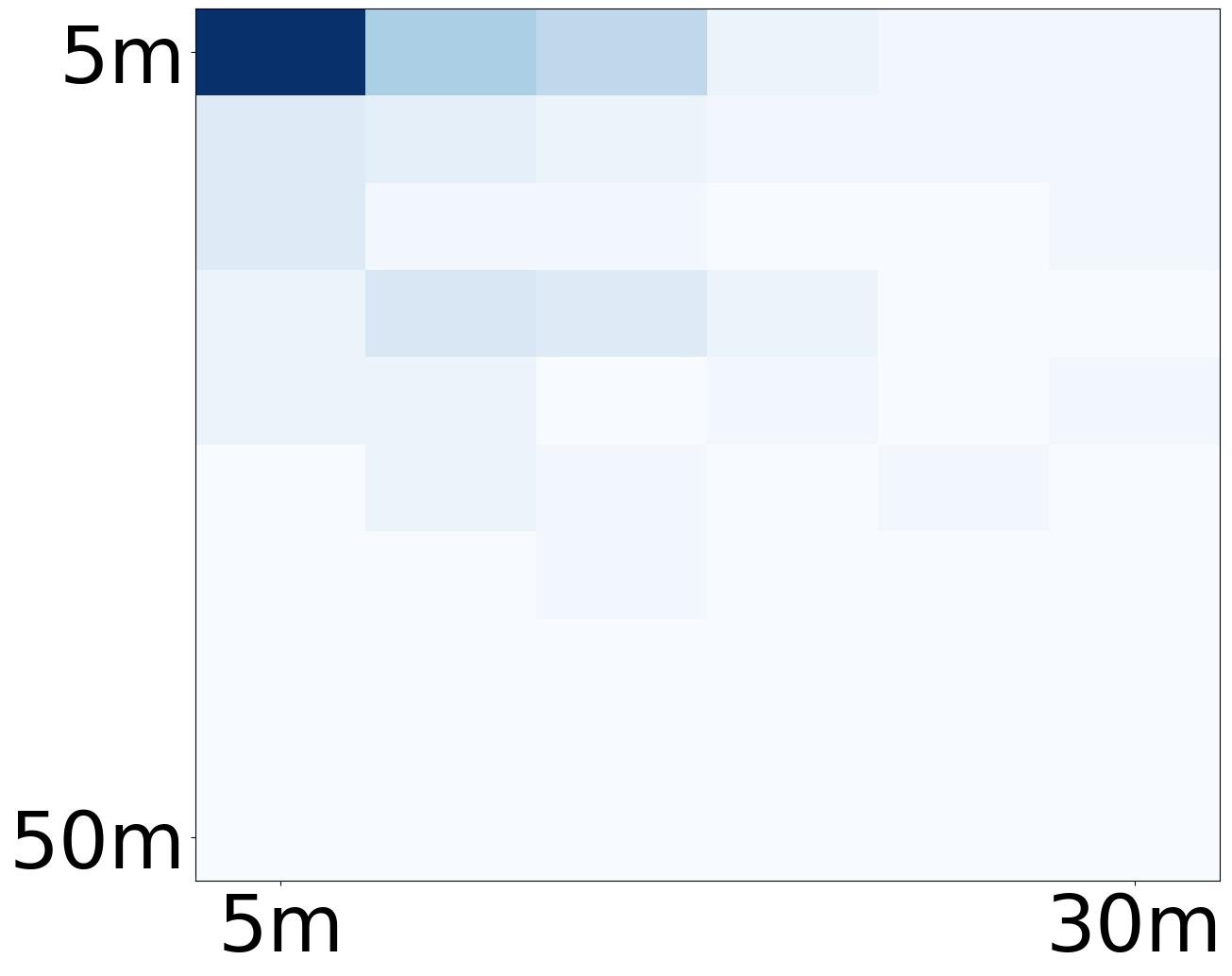} &
\includegraphics[trim=0mm 0mm 0mm 0mm,clip,width=0.185\linewidth]{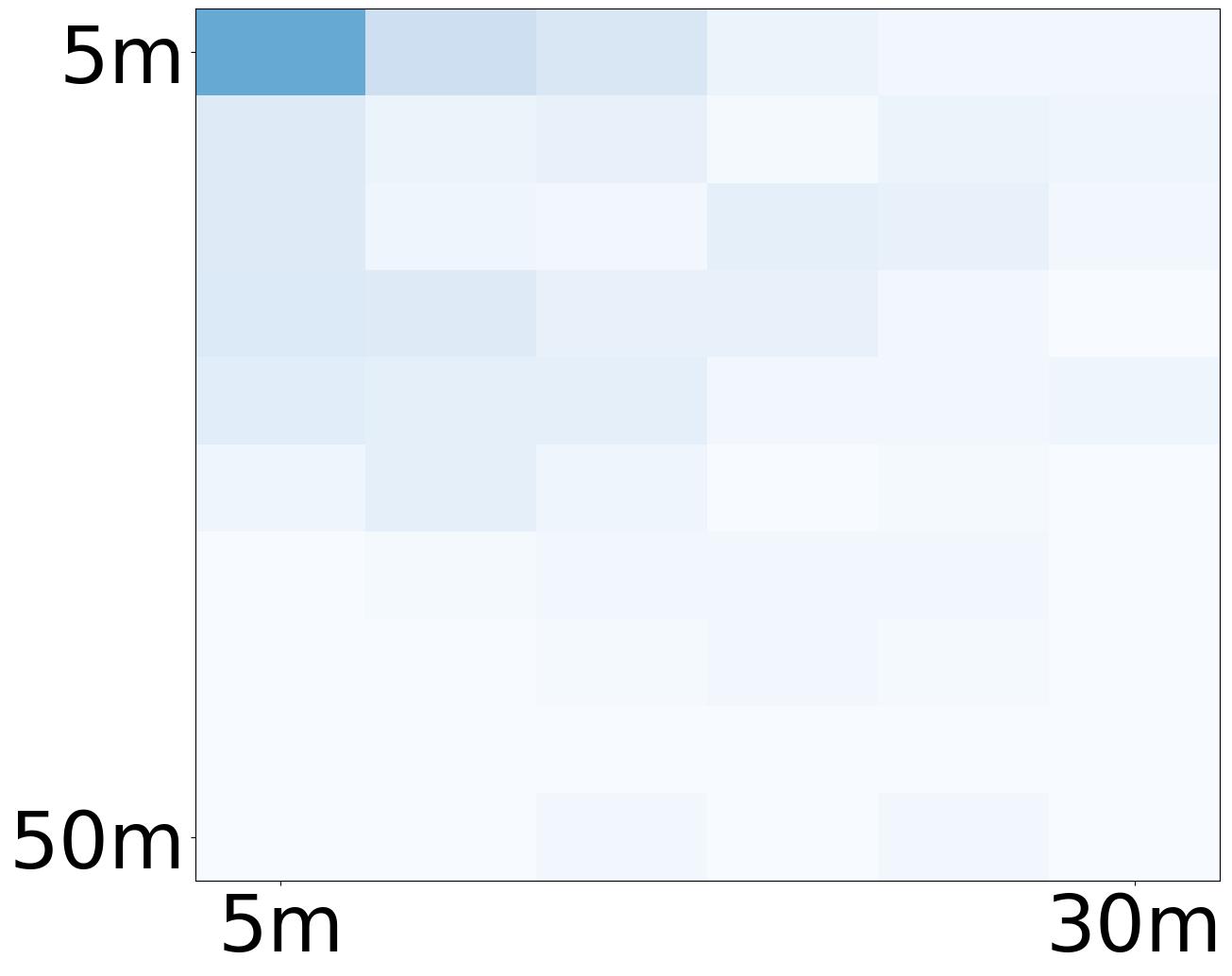} &
\includegraphics[trim=0mm 0mm 0mm 0mm,clip,width=0.185\linewidth]{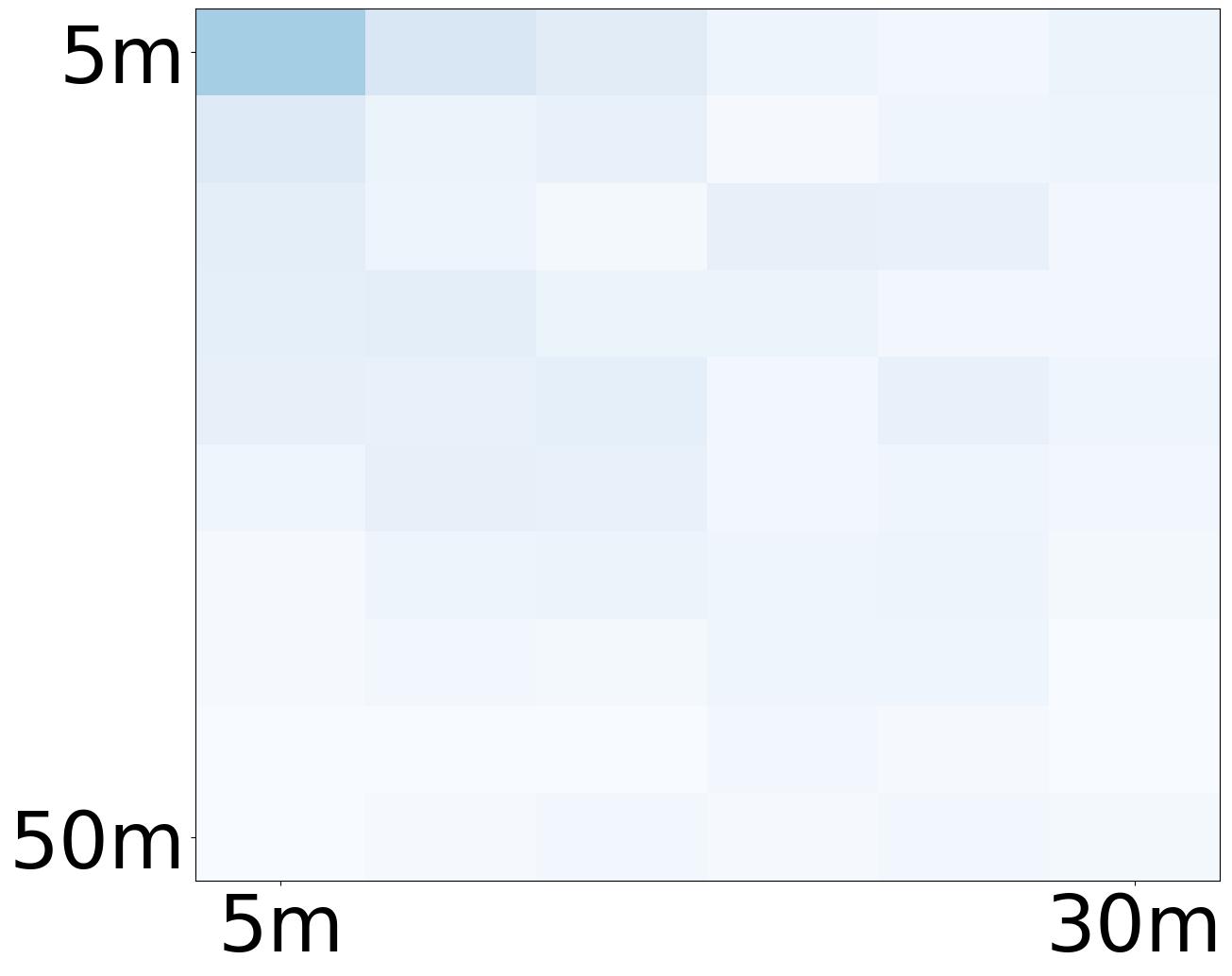} &
\includegraphics[trim=0mm 0mm 0mm 0mm,clip,width=0.185\linewidth]{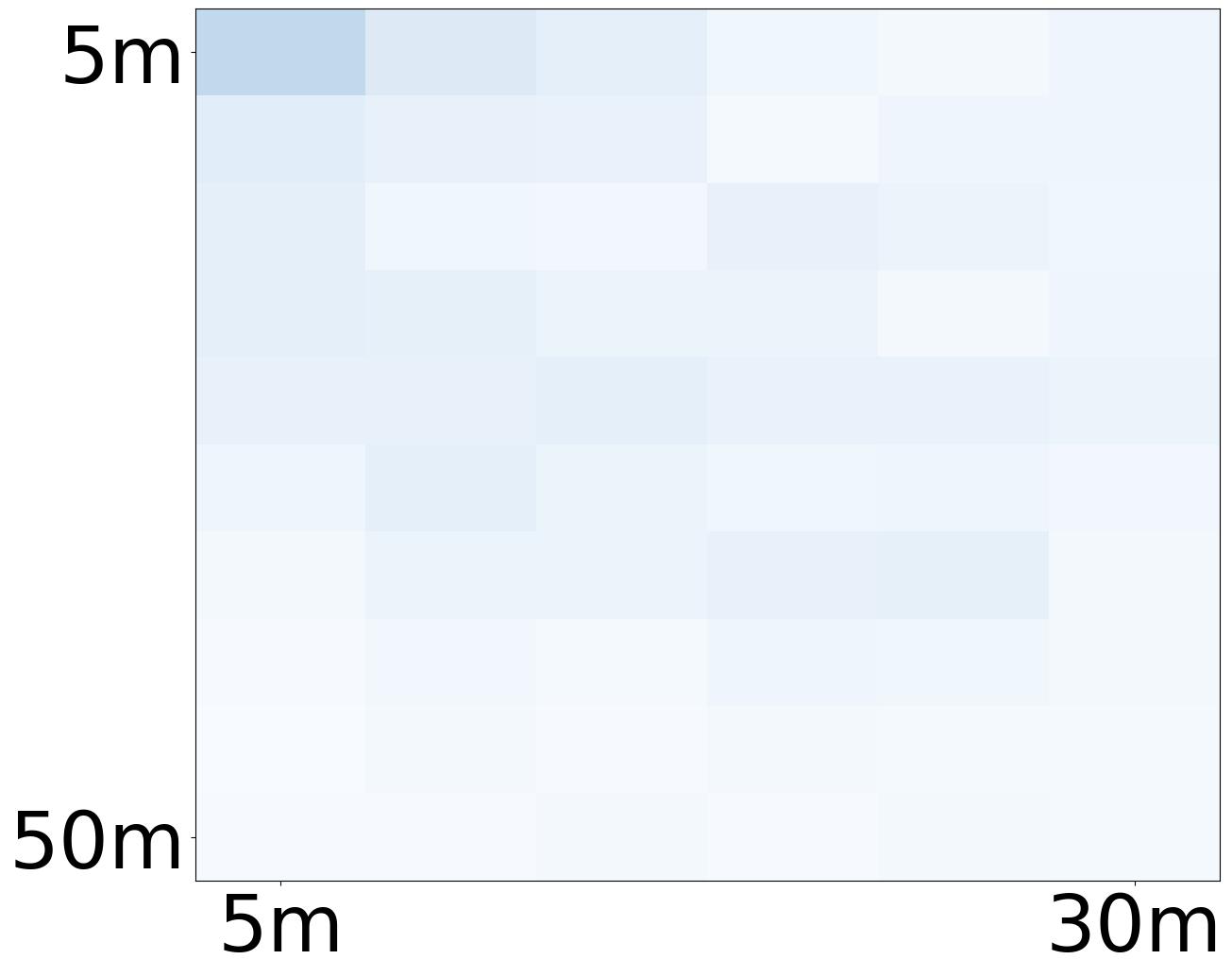} &
\includegraphics[trim=0mm 0mm 0mm 0mm,clip,width=0.185\linewidth]{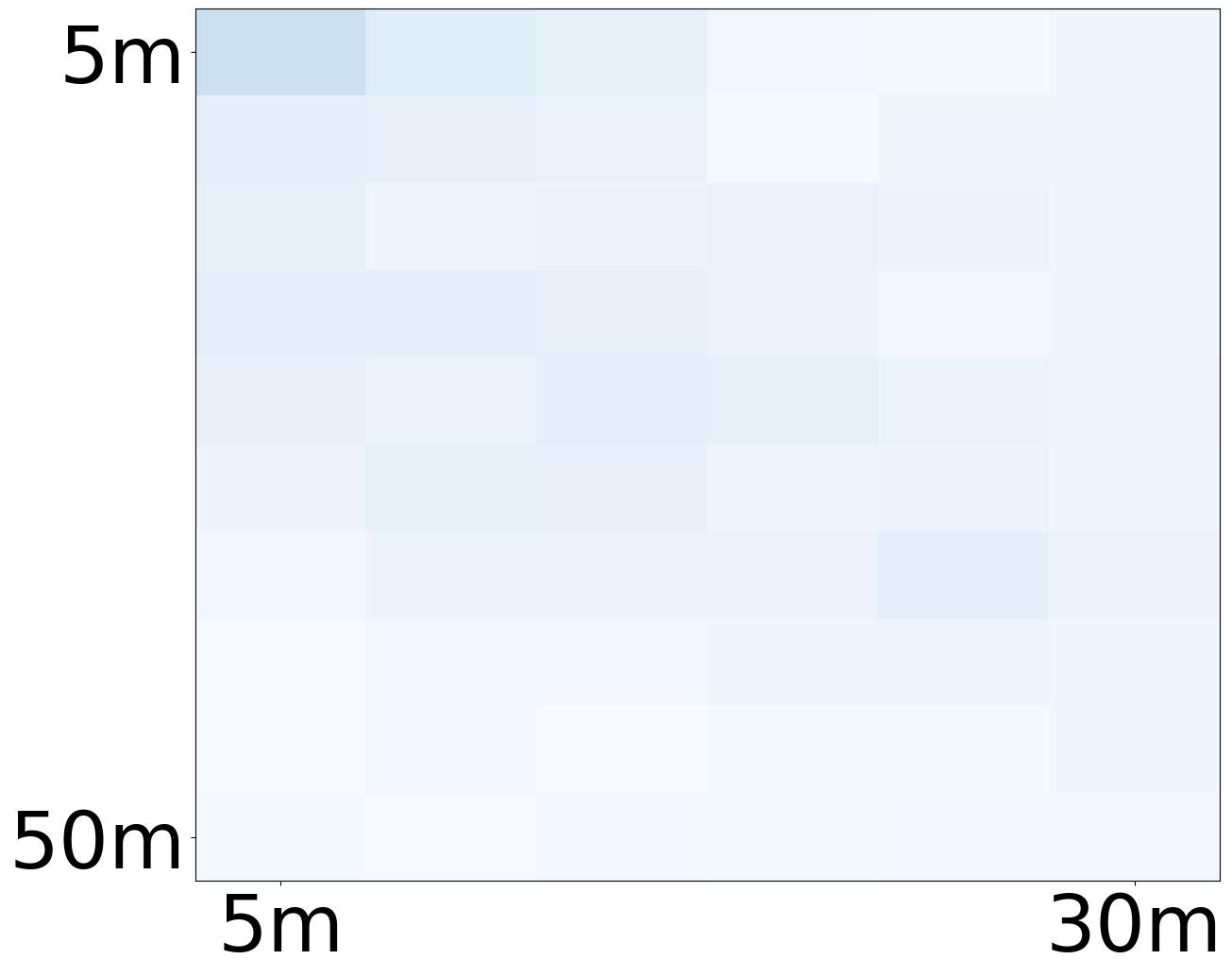} \\
\multicolumn{5}{c}{ICG, 20-img}\\
\scriptsize{{\bf iter. 1}} & \scriptsize{{\bf iter. 2}} & \scriptsize{{\bf iter. 3}} & \scriptsize{{\bf iter. 4}} & \scriptsize{{\bf iter. 5}}\\
\includegraphics[trim=0mm 0mm 0mm 0mm,clip,width=0.185\linewidth]{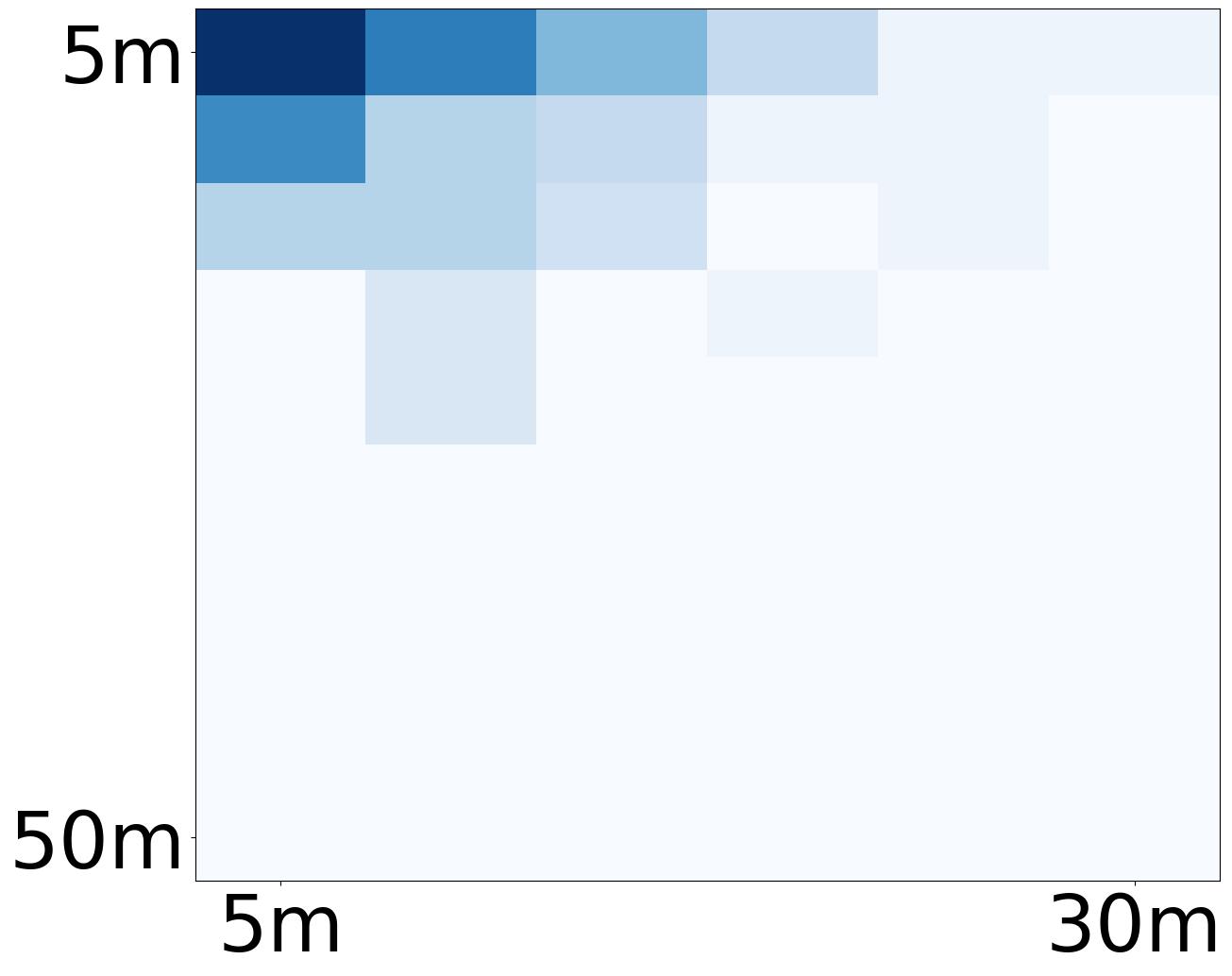} &
\includegraphics[trim=0mm 0mm 0mm 0mm,clip,width=0.185\linewidth]{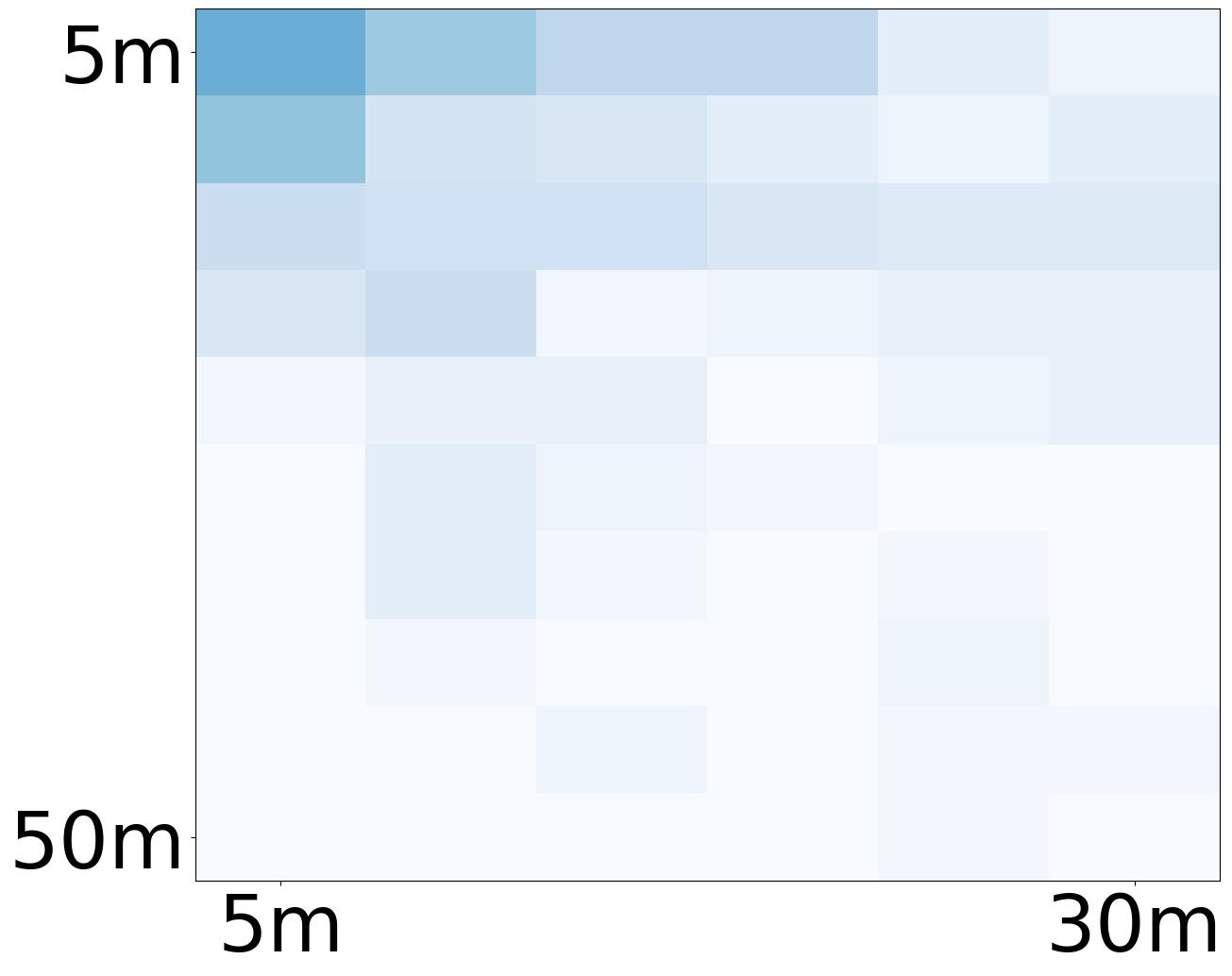} &
\includegraphics[trim=0mm 0mm 0mm 0mm,clip,width=0.185\linewidth]{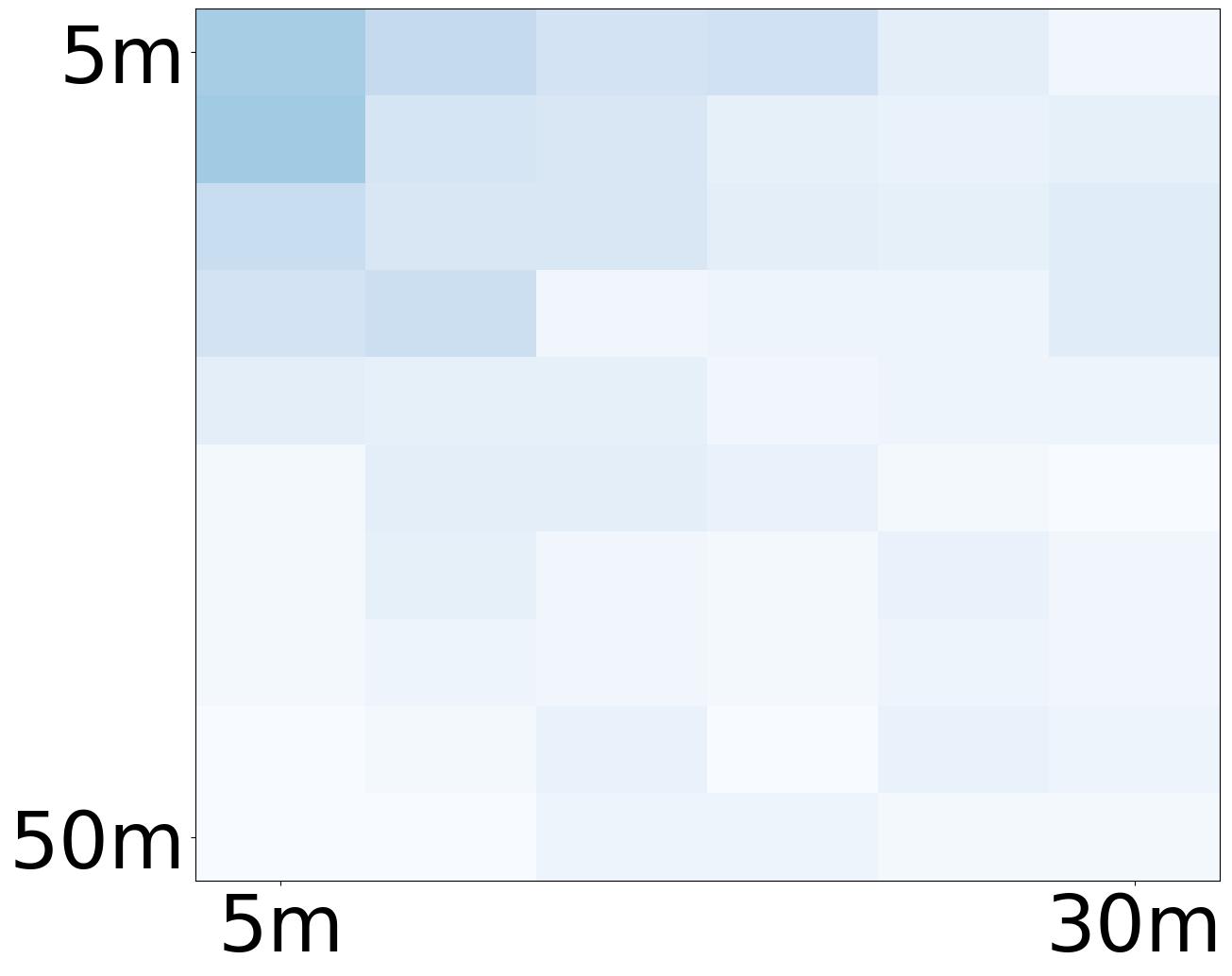} &
\includegraphics[trim=0mm 0mm 0mm 0mm,clip,width=0.185\linewidth]{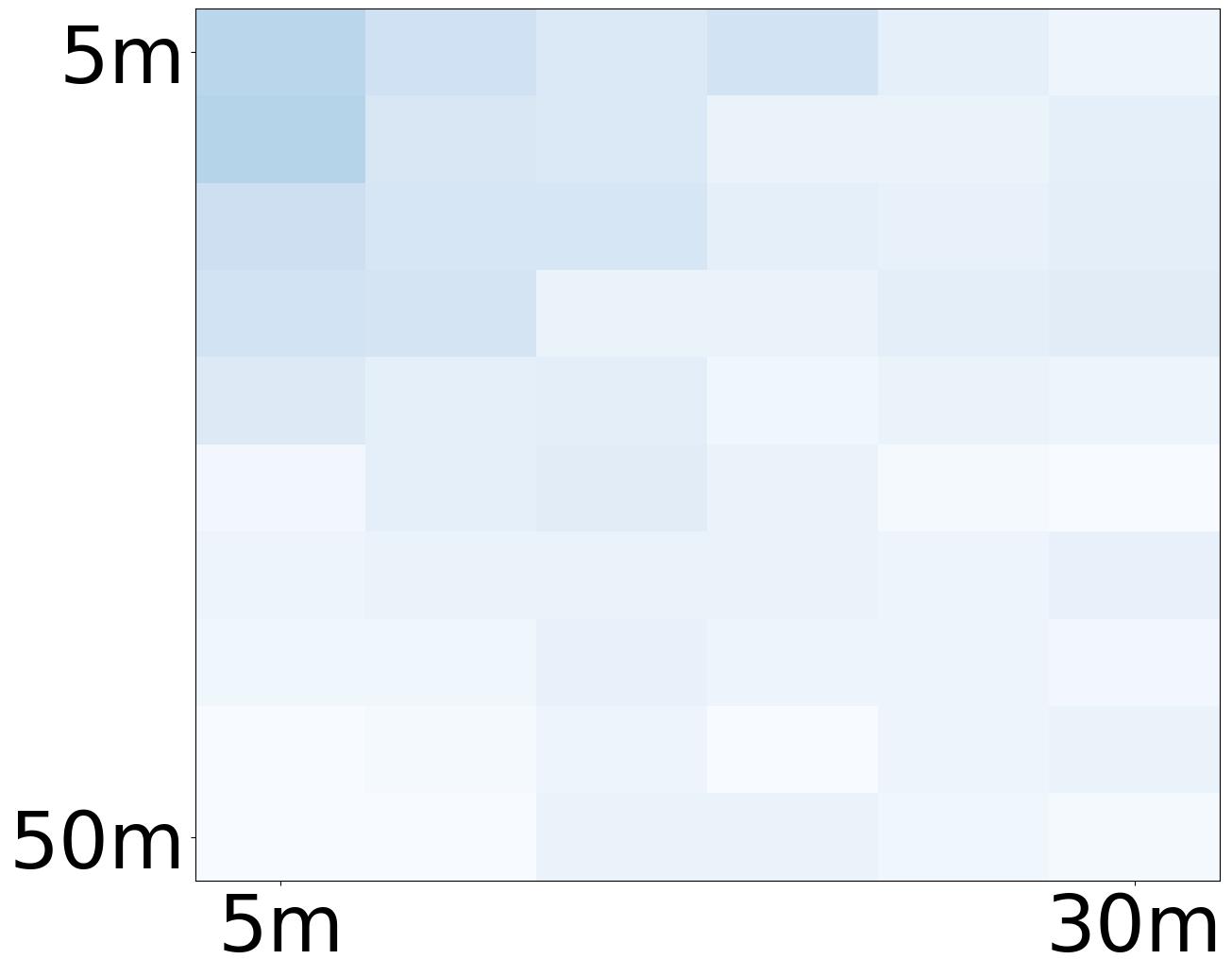} &
\includegraphics[trim=0mm 0mm 0mm 0mm,clip,width=0.185\linewidth]{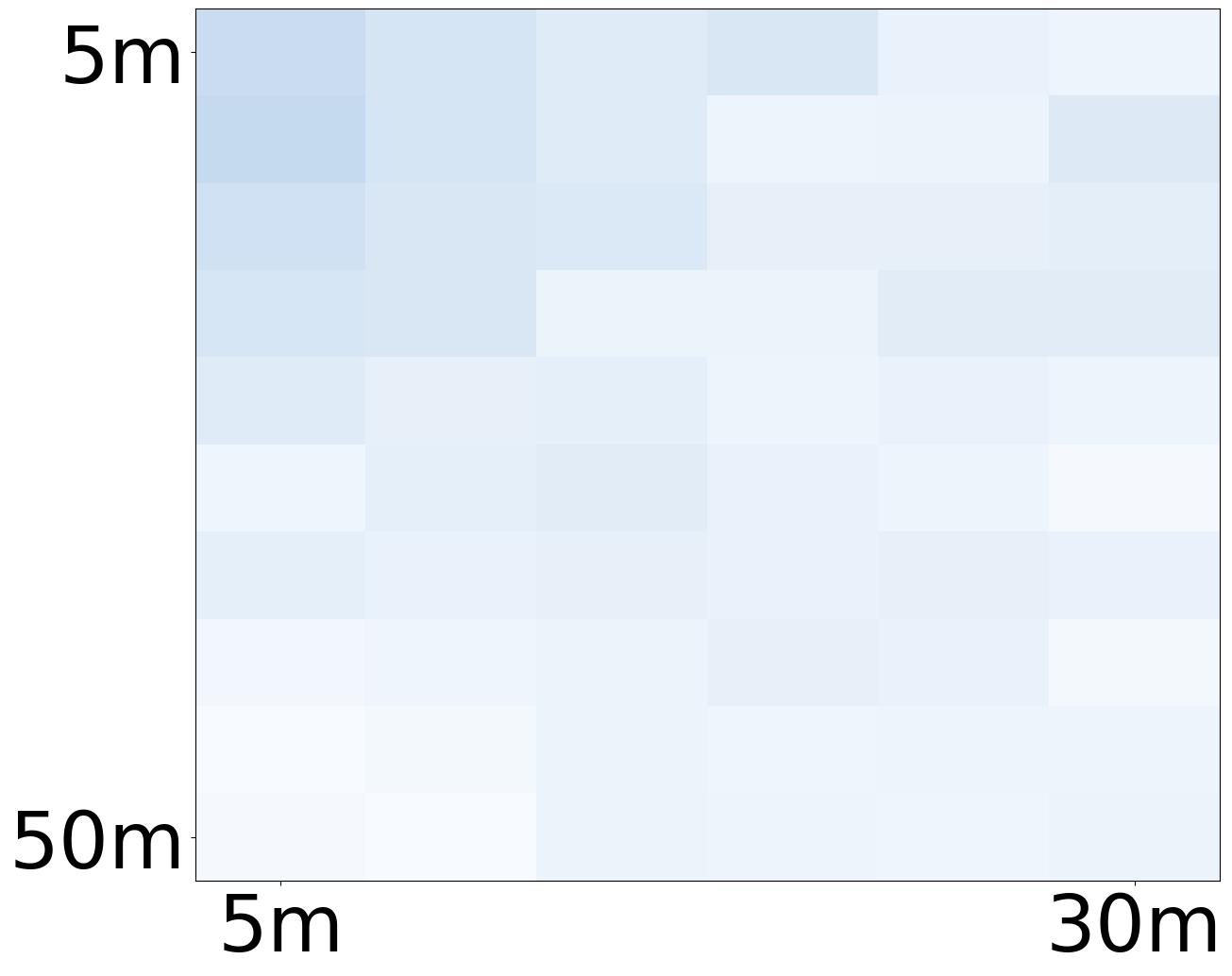} \\
\multicolumn{5}{c}{ICG, 50-img}\\
\end{tabular}
}
\caption{{\bf Accumulated distributions of transformation candidates with respect to camera locations in more setups.} This figure shows the distributions in other five experimental setups except for the setup (i.e., VisDrone, 50-img) shown in Figure 3 of the main manuscript. The $x$ and $y$ axes indicate altitude and rotation circle radius from the target human.}
\label{fig:distr_camera_location}
\end{figure*}
\begin{figure*}[t]
\centering
\resizebox{.75\linewidth}{!}{%
\setlength{\tabcolsep}{2.0pt}
\begin{tabular}{ccccc}
\scriptsize{{\bf iter. 1}} & \scriptsize{{\bf iter. 2}} & \scriptsize{{\bf iter. 3}} & \scriptsize{{\bf iter. 4}} & \scriptsize{{\bf iter. 5}}\\
\includegraphics[trim=0mm 0mm 0mm 0mm,clip,width=0.185\linewidth]{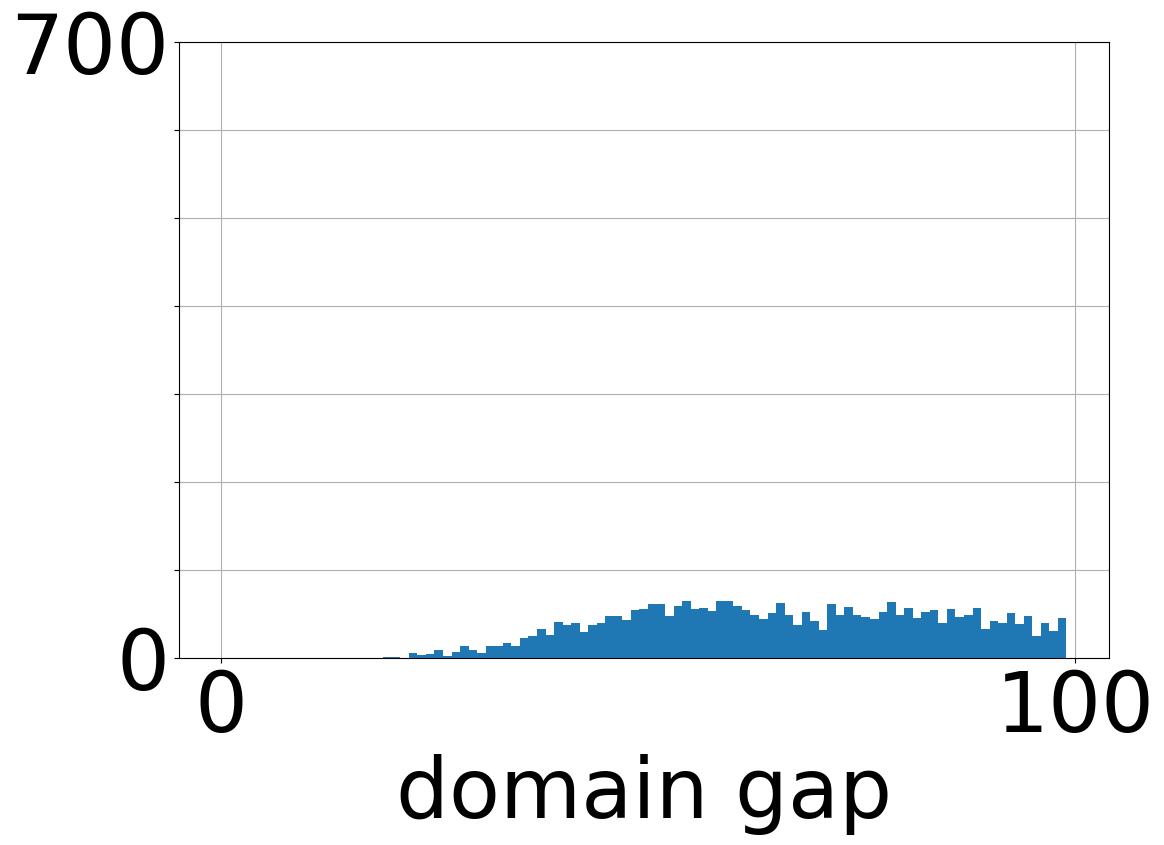} &
\includegraphics[trim=0mm 0mm 0mm 0mm,clip,width=0.185\linewidth]{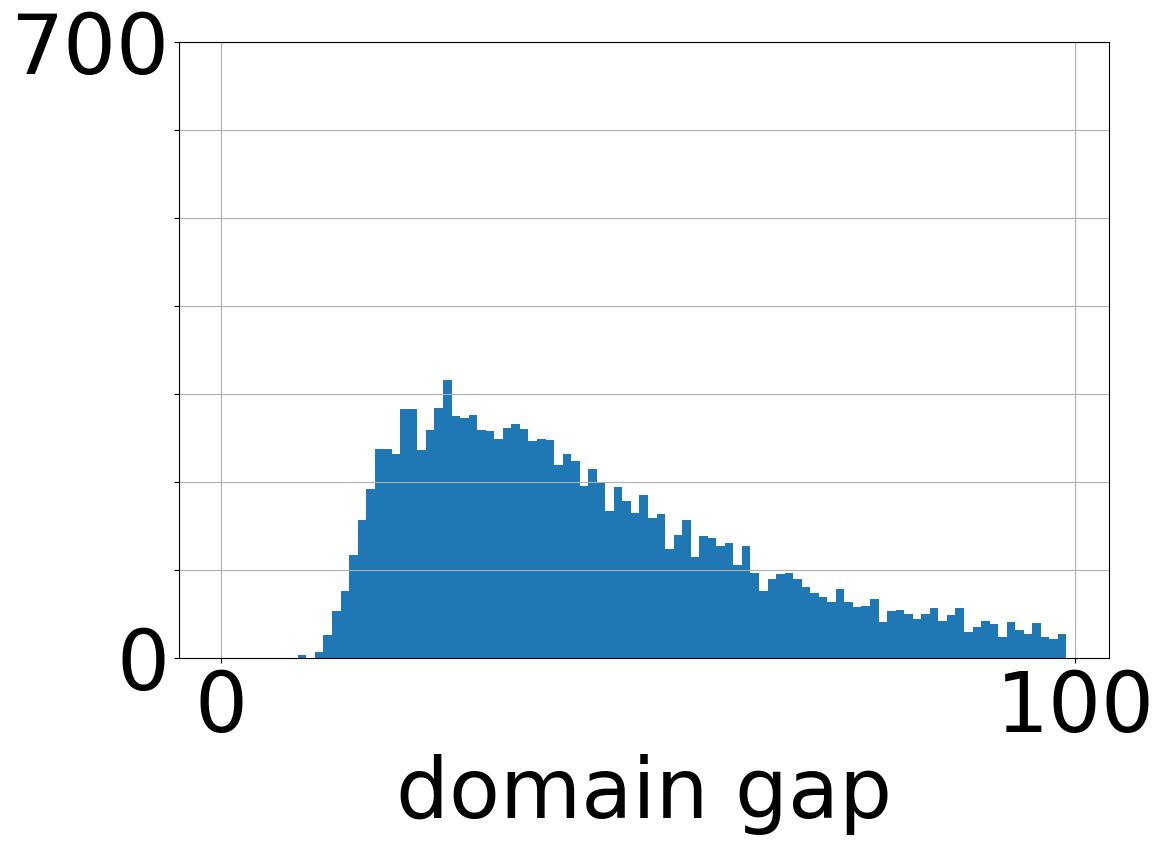} &
\includegraphics[trim=0mm 0mm 0mm 0mm,clip,width=0.185\linewidth]{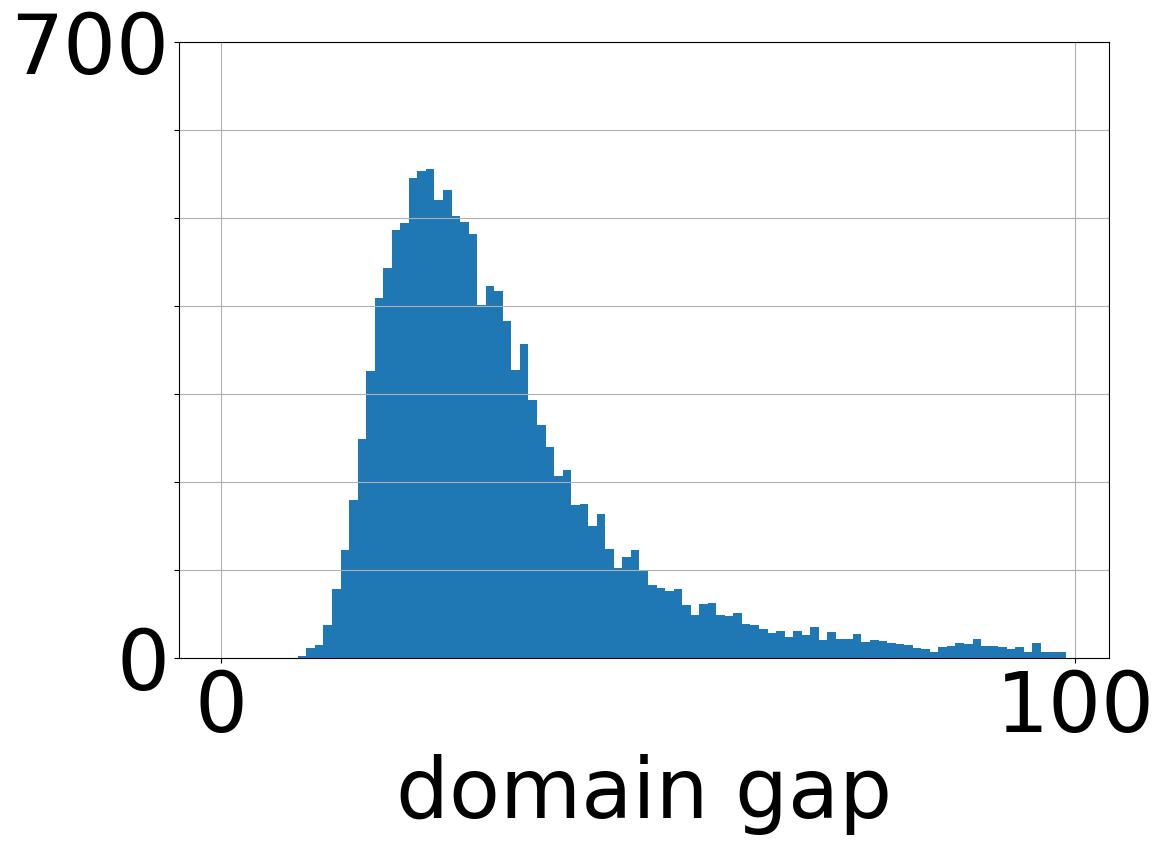} &
\includegraphics[trim=0mm 0mm 0mm 0mm,clip,width=0.185\linewidth]{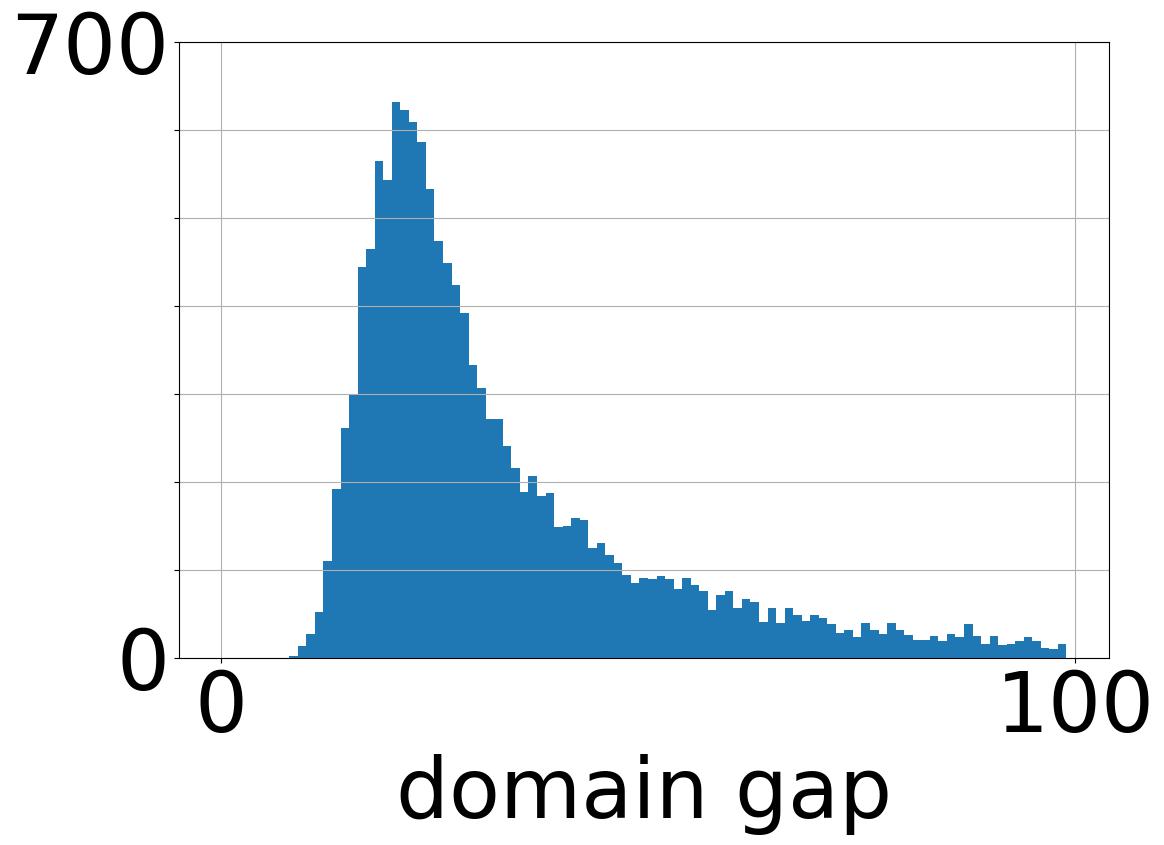} &
\includegraphics[trim=0mm 0mm 0mm 0mm,clip,width=0.185\linewidth]{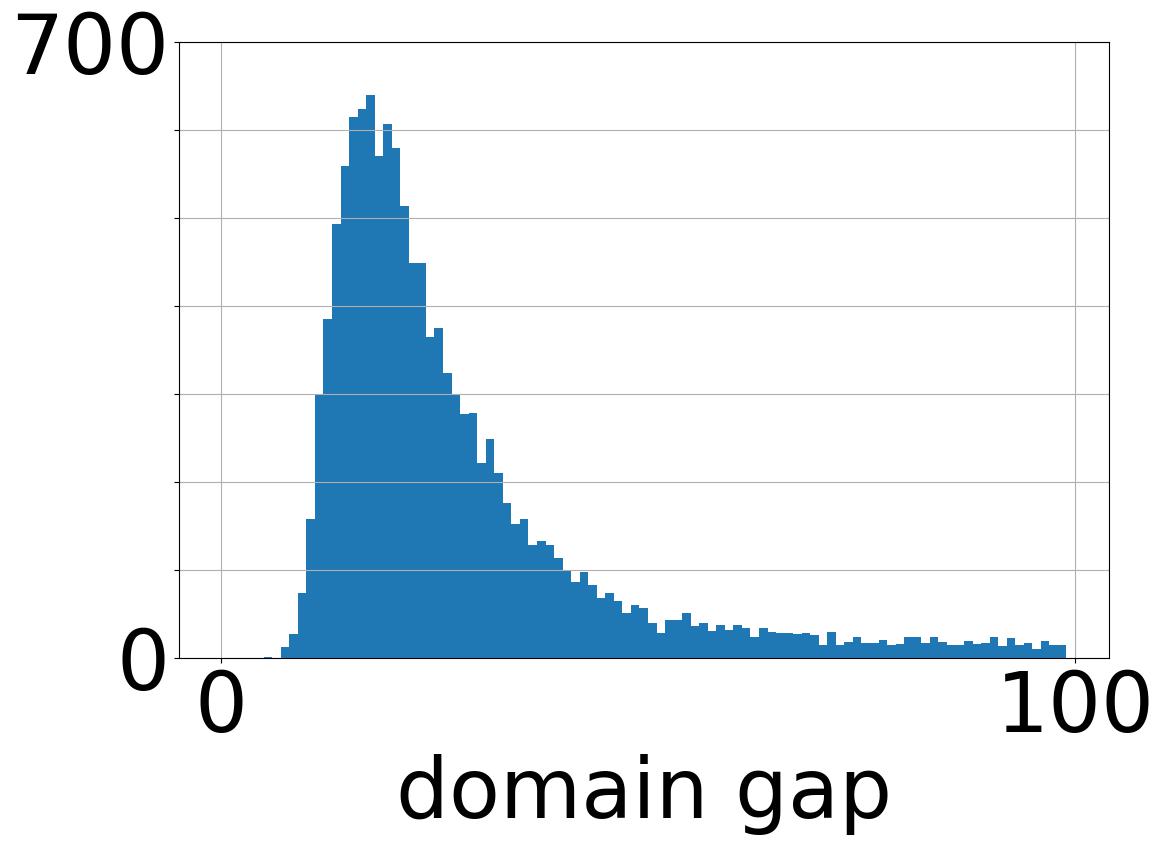} \\
\multicolumn{5}{c}{VisDrone, 20-img}\\
\scriptsize{{\bf iter. 1}} & \scriptsize{{\bf iter. 2}} & \scriptsize{{\bf iter. 3}} & \scriptsize{{\bf iter. 4}} & \scriptsize{{\bf iter. 5}}\\
\includegraphics[trim=0mm 0mm 0mm 0mm,clip,width=0.185\linewidth]{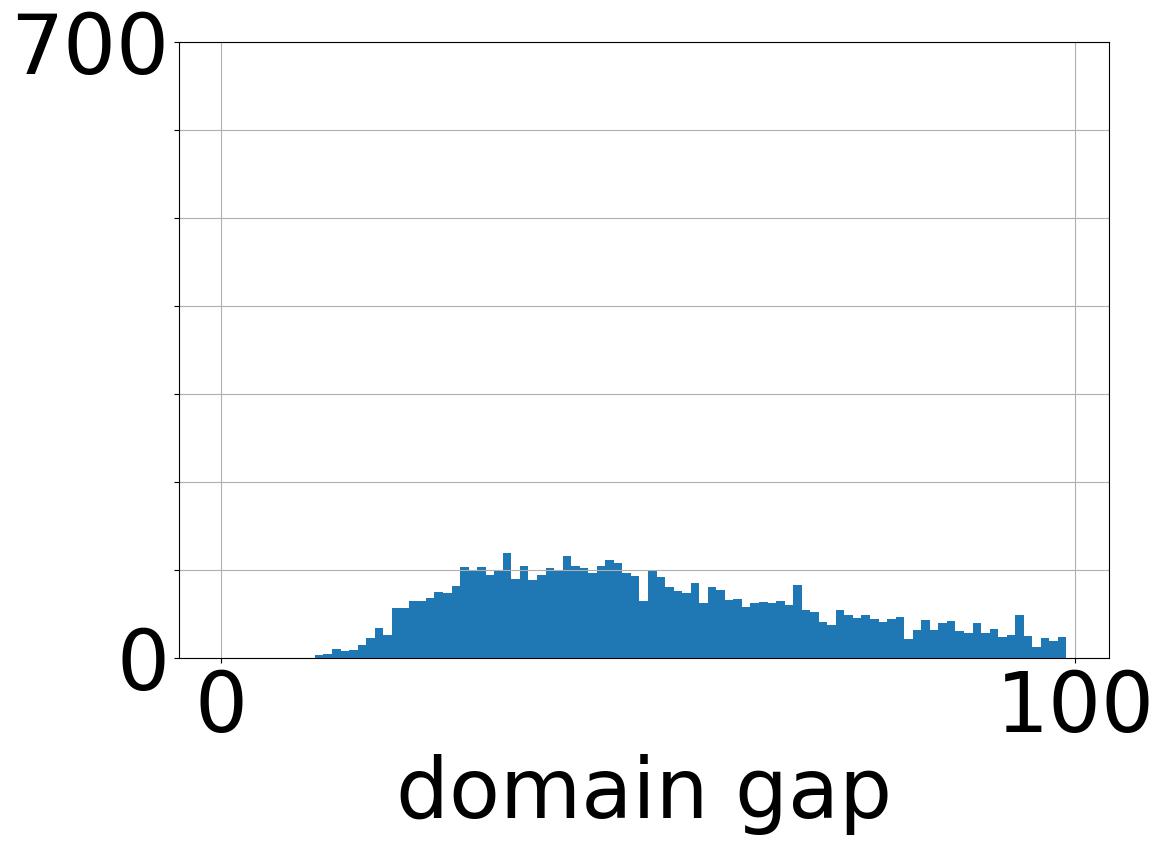} & 
\includegraphics[trim=0mm 0mm 0mm 0mm,clip,width=0.185\linewidth]{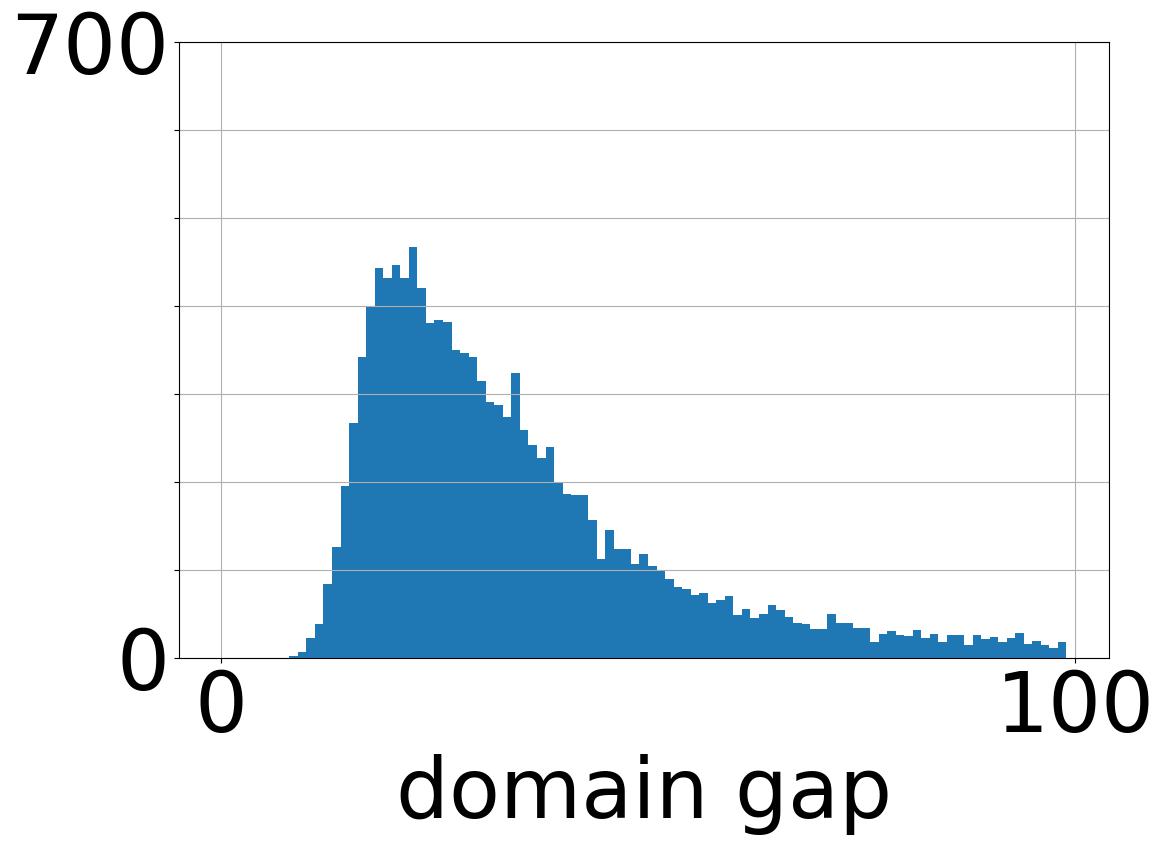} & 
\includegraphics[trim=0mm 0mm 0mm 0mm,clip,width=0.185\linewidth]{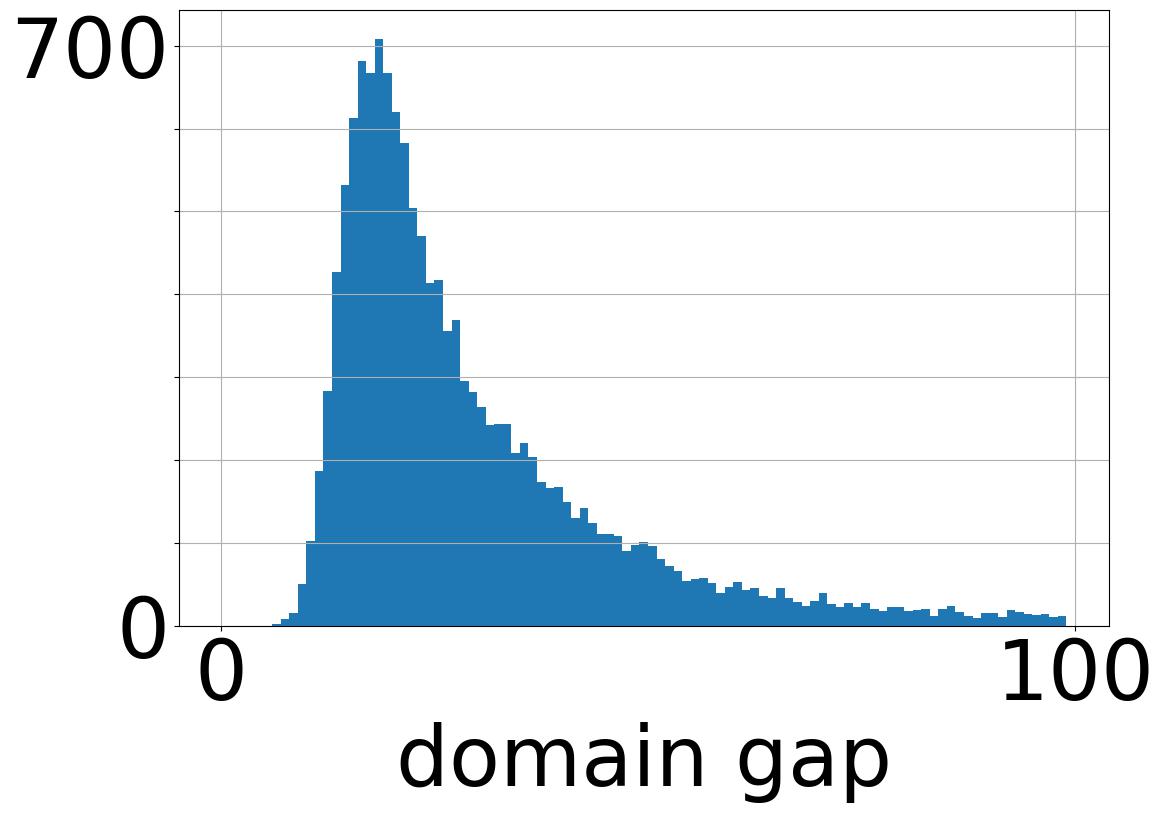} & 
\includegraphics[trim=0mm 0mm 0mm 0mm,clip,width=0.185\linewidth]{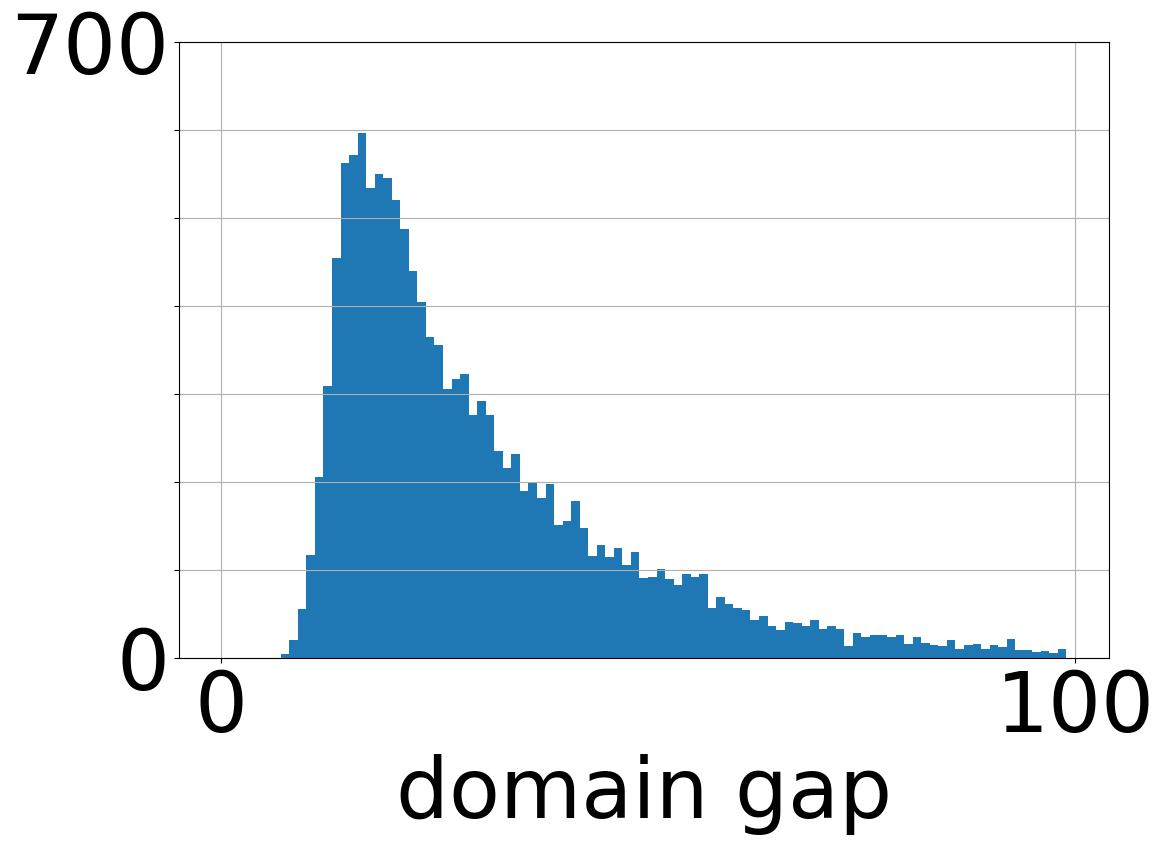} & 
\includegraphics[trim=0mm 0mm 0mm 0mm,clip,width=0.185\linewidth]{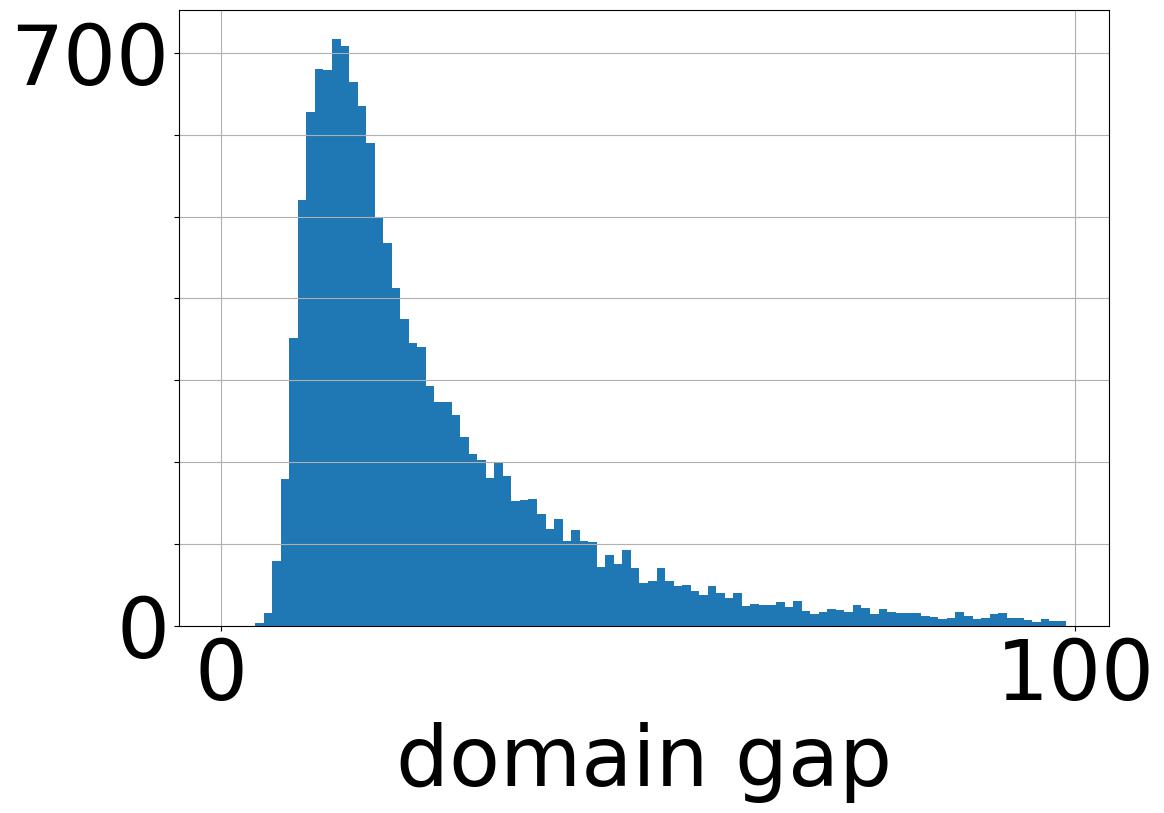}\\
\multicolumn{5}{c}{Okutama-Action, 20-img}\\
\scriptsize{{\bf iter. 1}} & \scriptsize{{\bf iter. 2}} & \scriptsize{{\bf iter. 3}} & \scriptsize{{\bf iter. 4}} & \scriptsize{{\bf iter. 5}}\\
\includegraphics[trim=0mm 0mm 0mm 0mm,clip,width=0.185\linewidth]{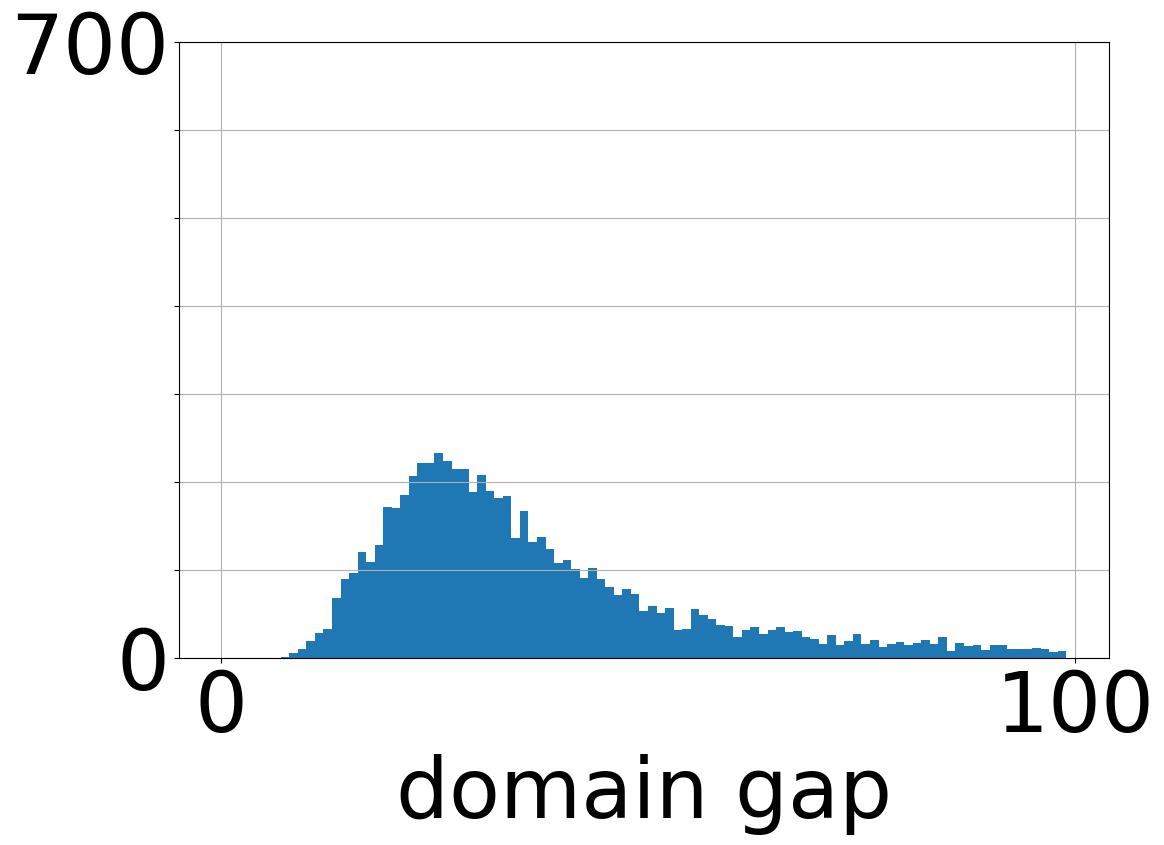} & 
\includegraphics[trim=0mm 0mm 0mm 0mm,clip,width=0.185\linewidth]{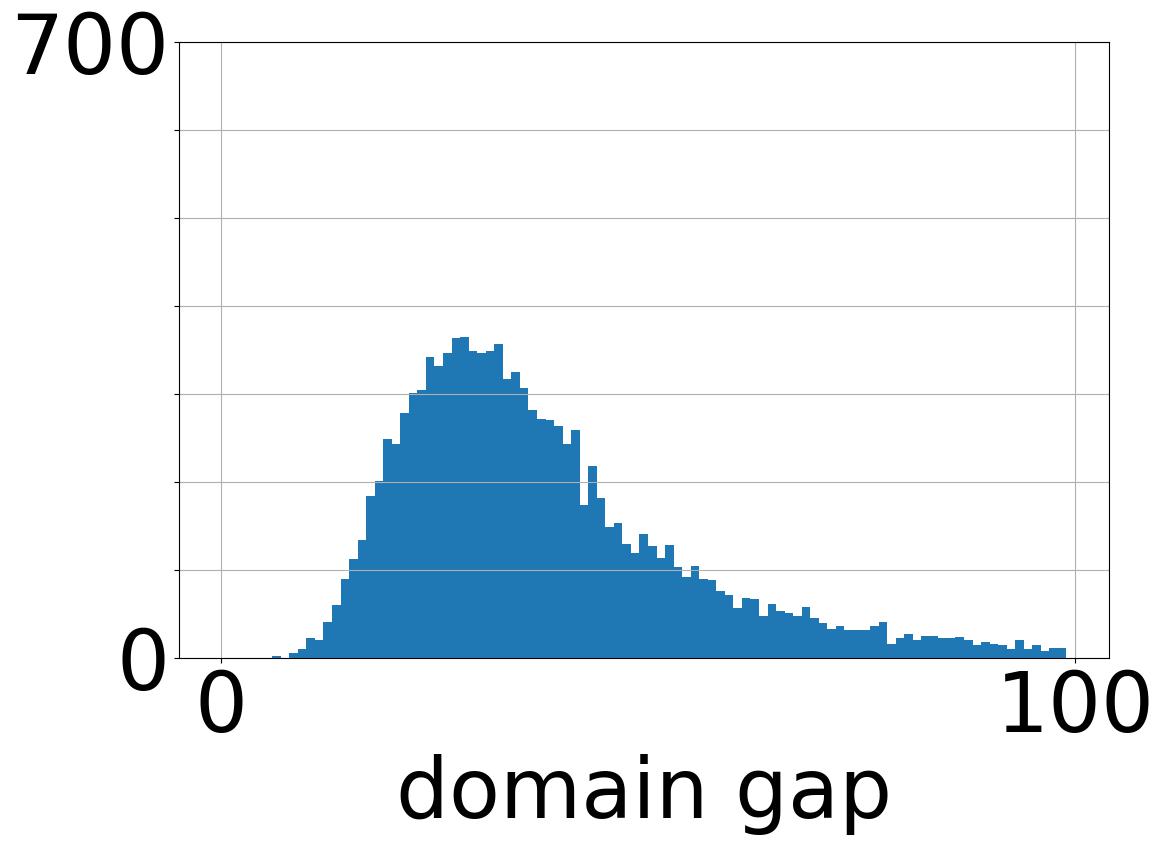} & 
\includegraphics[trim=0mm 0mm 0mm 0mm,clip,width=0.185\linewidth]{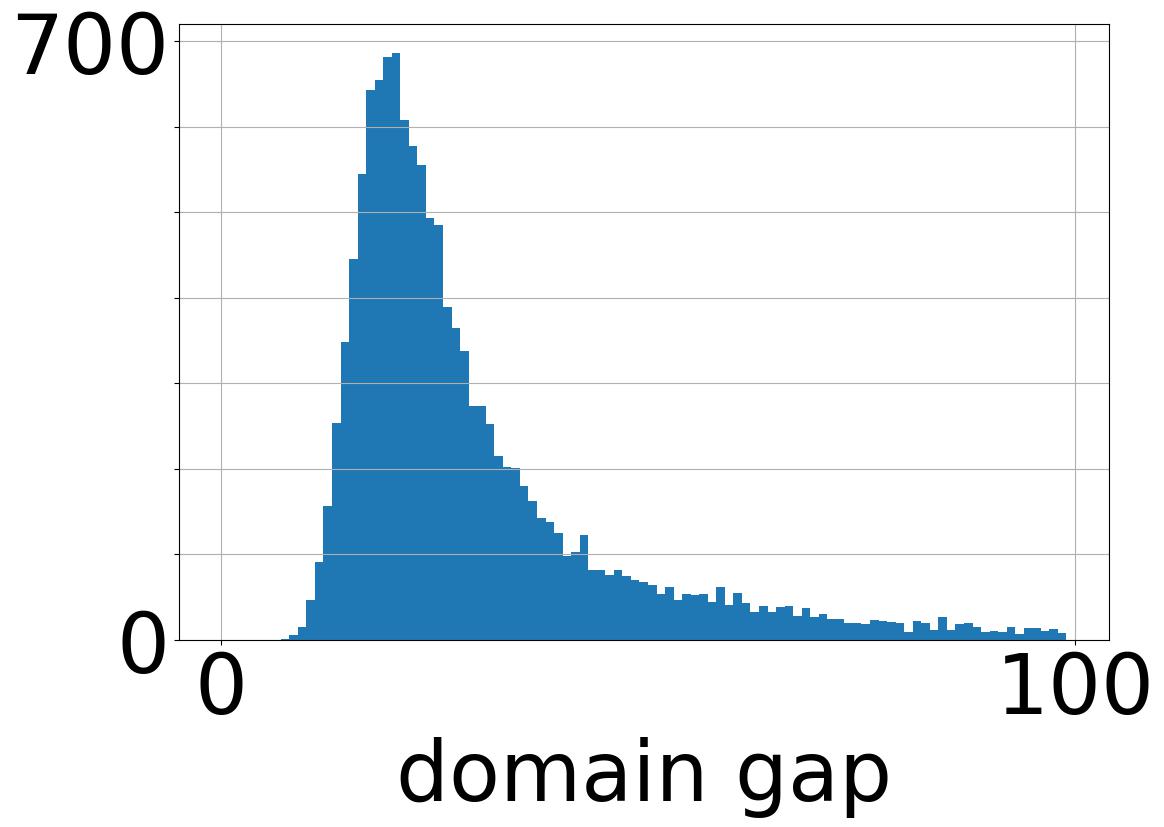} & 
\includegraphics[trim=0mm 0mm 0mm 0mm,clip,width=0.185\linewidth]{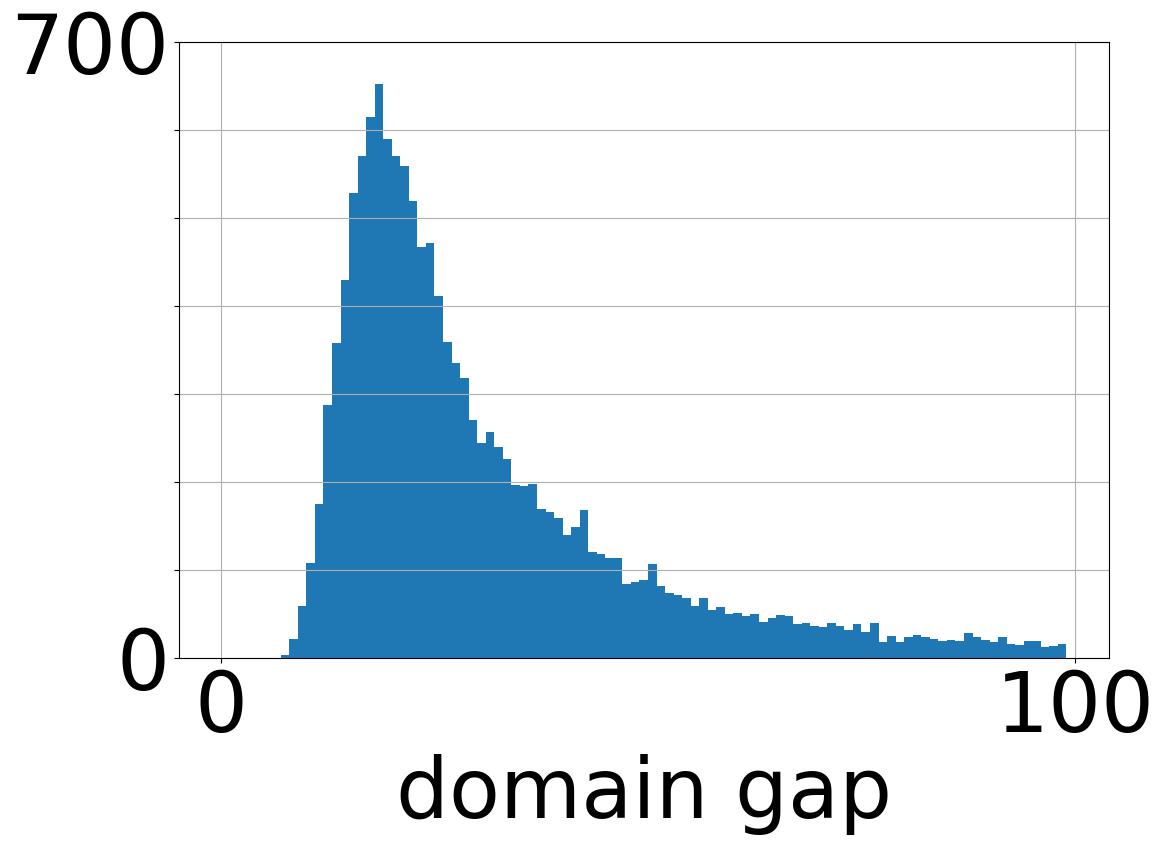} & 
\includegraphics[trim=0mm 0mm 0mm 0mm,clip,width=0.185\linewidth]{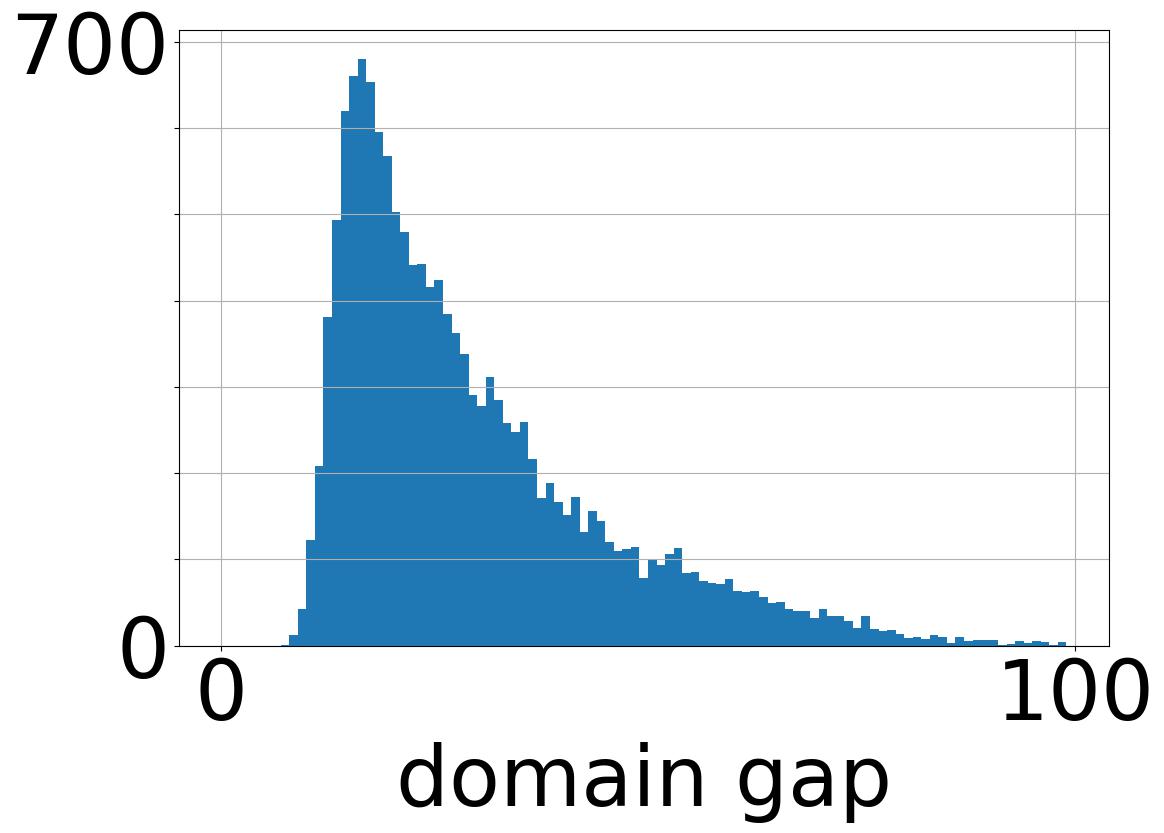} \\
\multicolumn{5}{c}{Okutama-Action, 50-img}\\
\scriptsize{{\bf iter. 1}} & \scriptsize{{\bf iter. 2}} & \scriptsize{{\bf iter. 3}} & \scriptsize{{\bf iter. 4}} & \scriptsize{{\bf iter. 5}}\\
\includegraphics[trim=0mm 0mm 0mm 0mm,clip,width=0.185\linewidth]{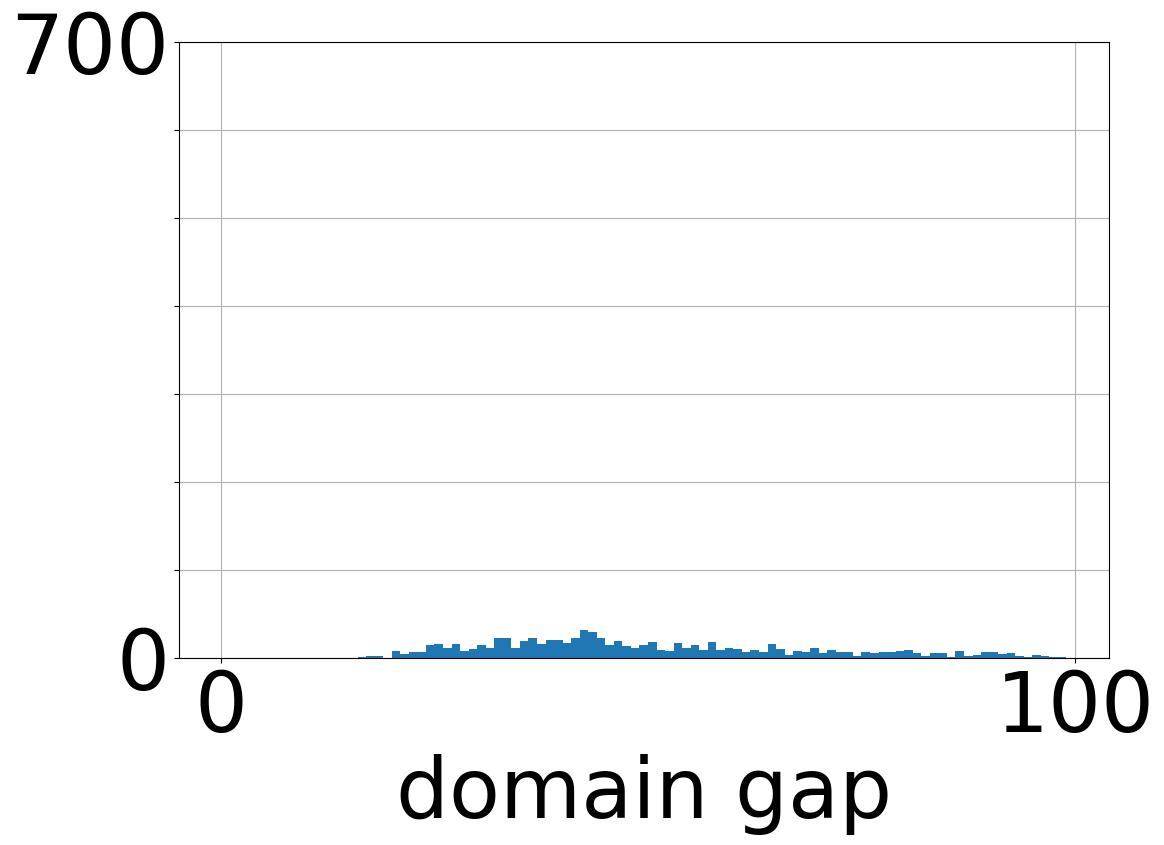} & 
\includegraphics[trim=0mm 0mm 0mm 0mm,clip,width=0.185\linewidth]{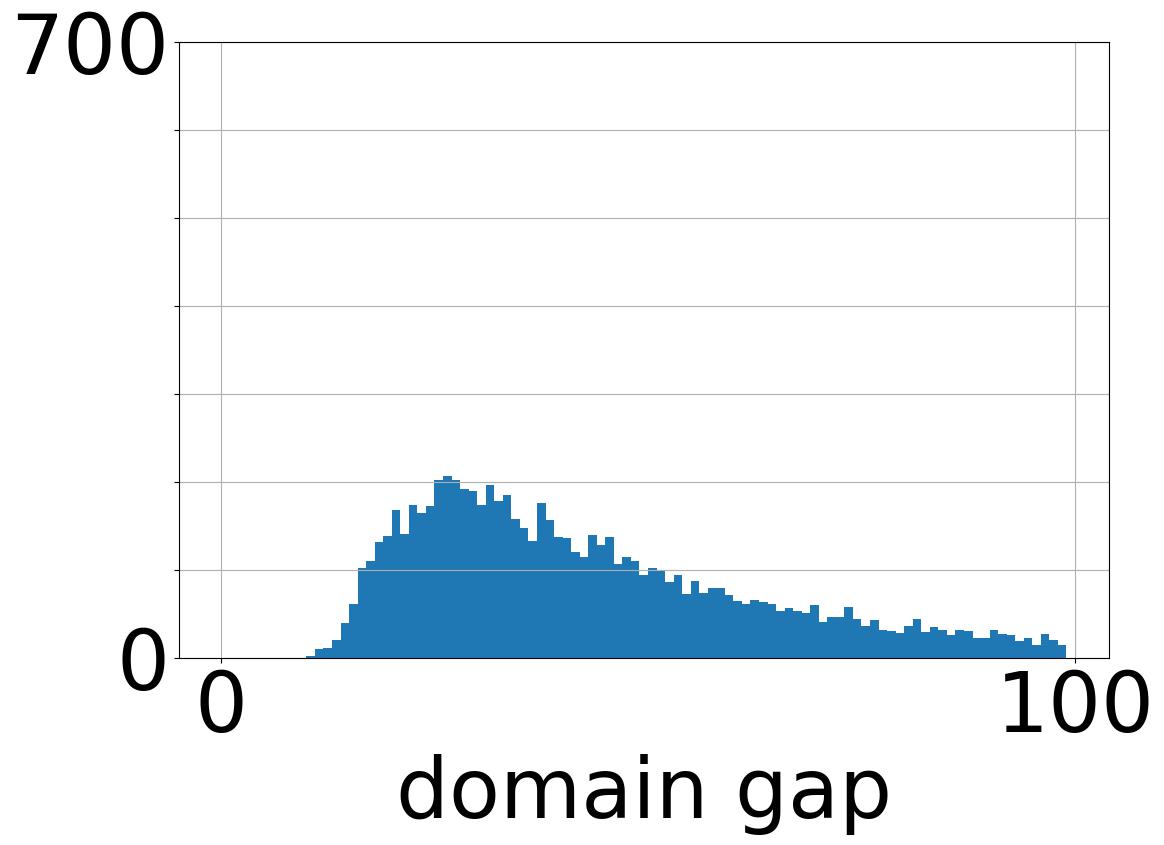} & 
\includegraphics[trim=0mm 0mm 0mm 0mm,clip,width=0.185\linewidth]{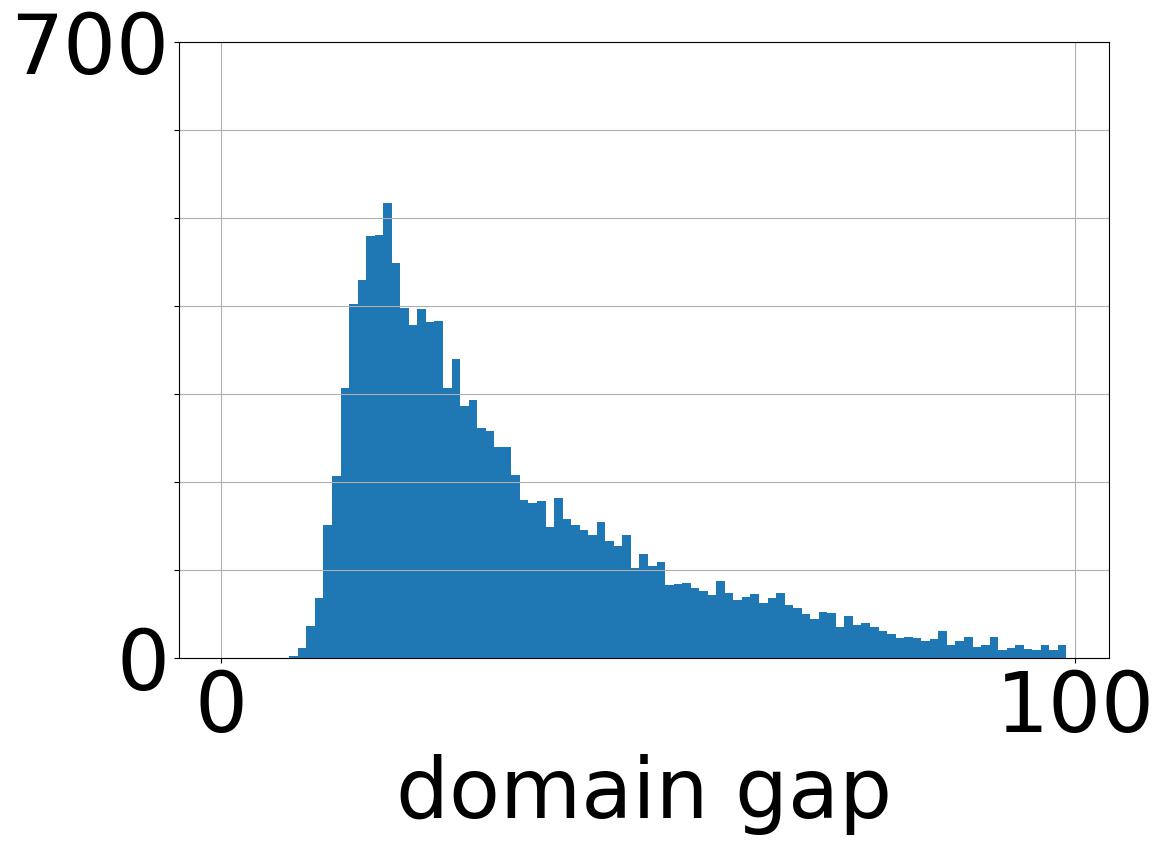} & 
\includegraphics[trim=0mm 0mm 0mm 0mm,clip,width=0.185\linewidth]{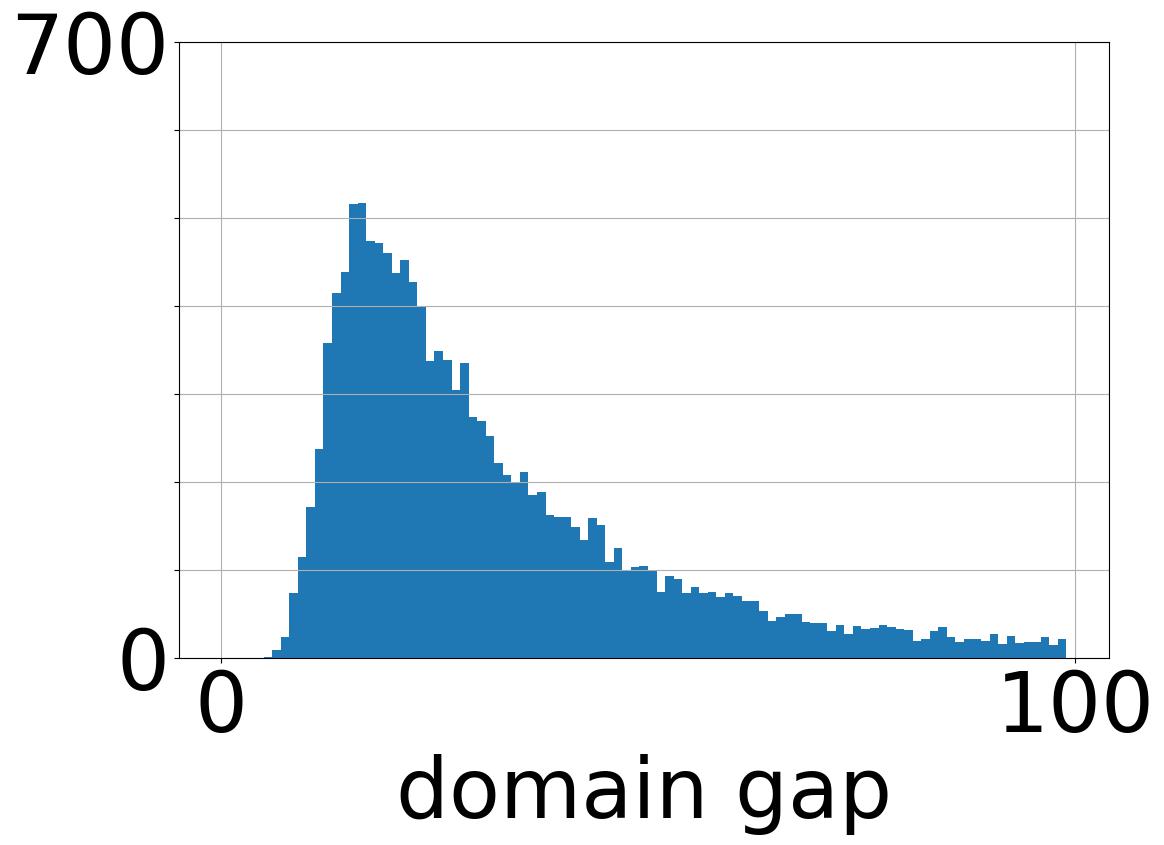} & 
\includegraphics[trim=0mm 0mm 0mm 0mm,clip,width=0.185\linewidth]{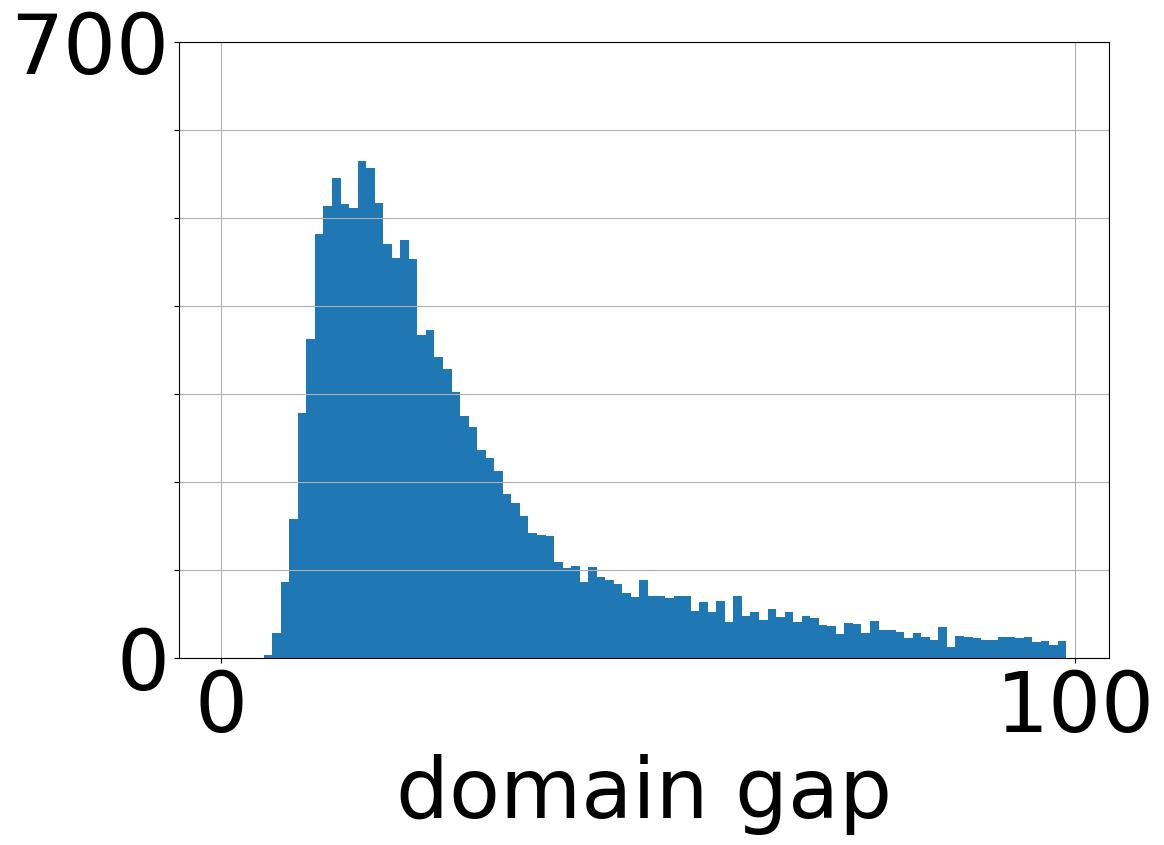} \\
\multicolumn{5}{c}{ICG, 20-img}\\
\scriptsize{{\bf iter. 1}} & \scriptsize{{\bf iter. 2}} & \scriptsize{{\bf iter. 3}} & \scriptsize{{\bf iter. 4}} & \scriptsize{{\bf iter. 5}}\\
\includegraphics[trim=0mm 0mm 0mm 0mm,clip,width=0.185\linewidth]{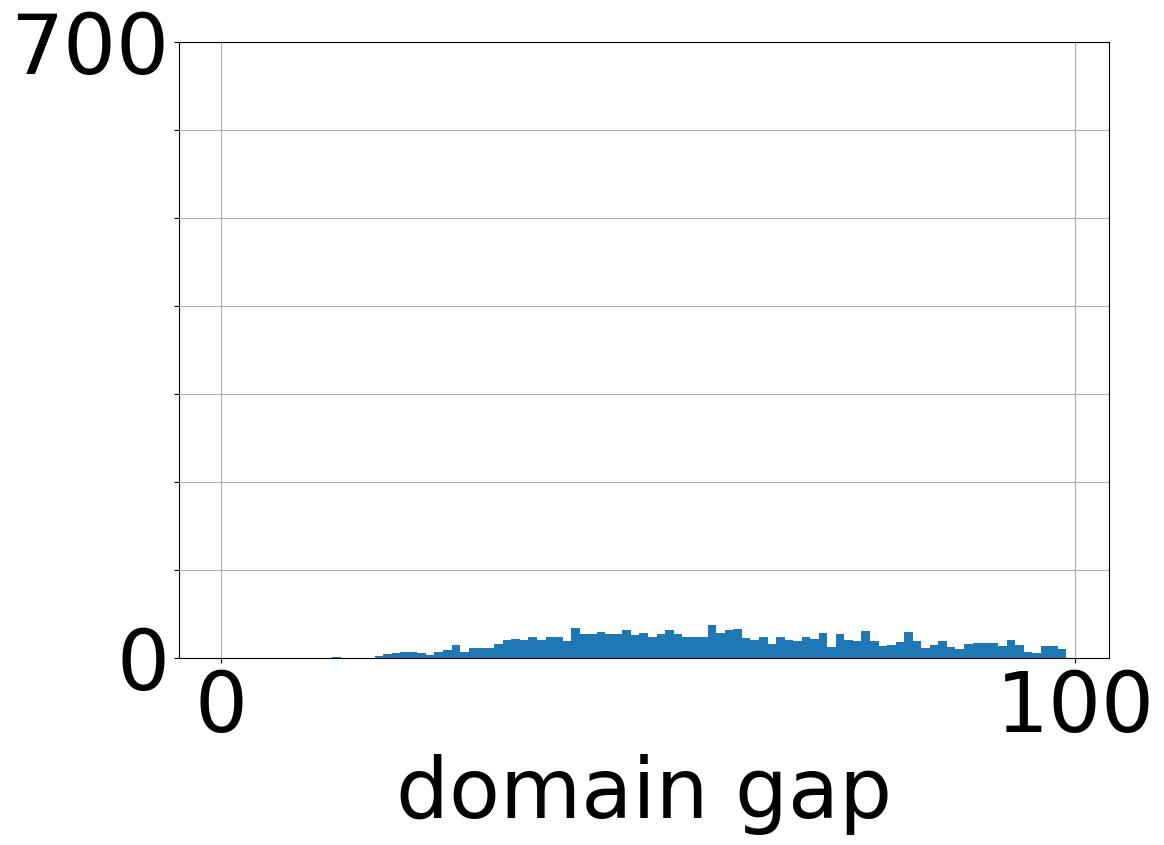} & 
\includegraphics[trim=0mm 0mm 0mm 0mm,clip,width=0.185\linewidth]{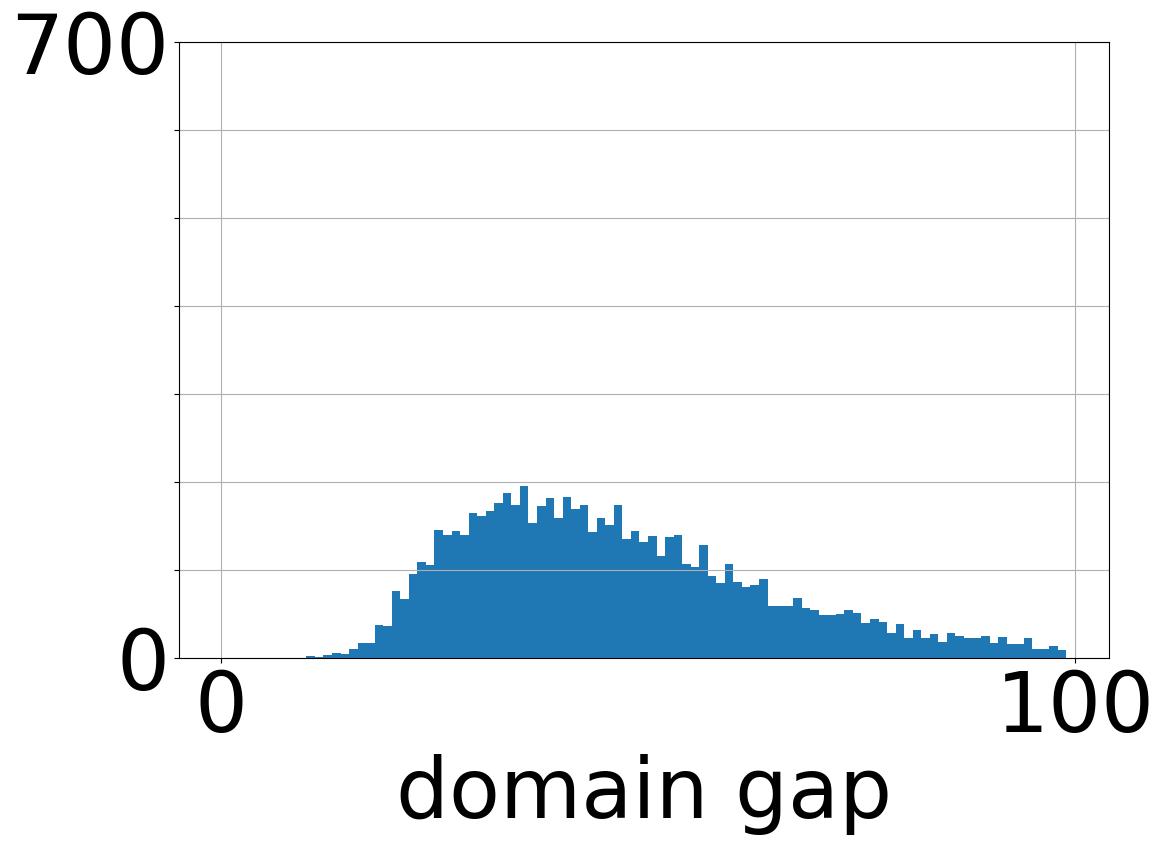} & 
\includegraphics[trim=0mm 0mm 0mm 0mm,clip,width=0.185\linewidth]{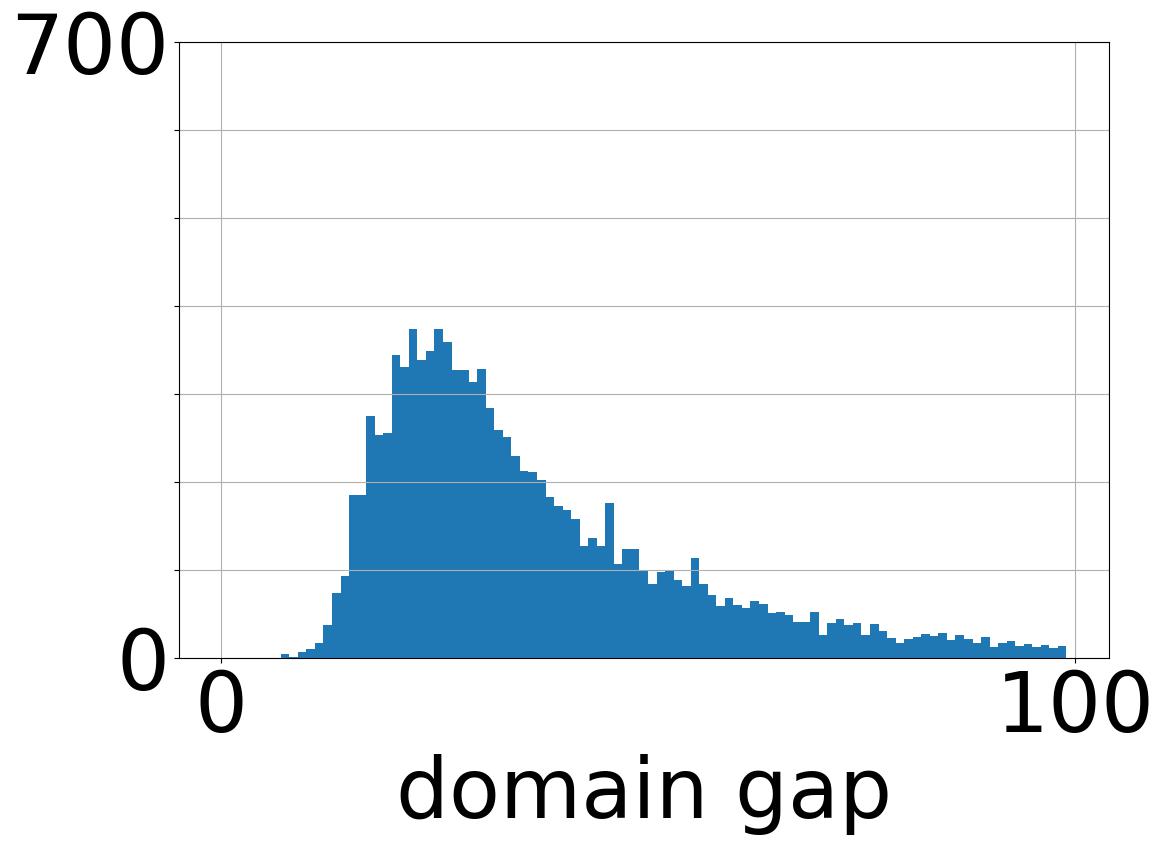} & 
\includegraphics[trim=0mm 0mm 0mm 0mm,clip,width=0.185\linewidth]{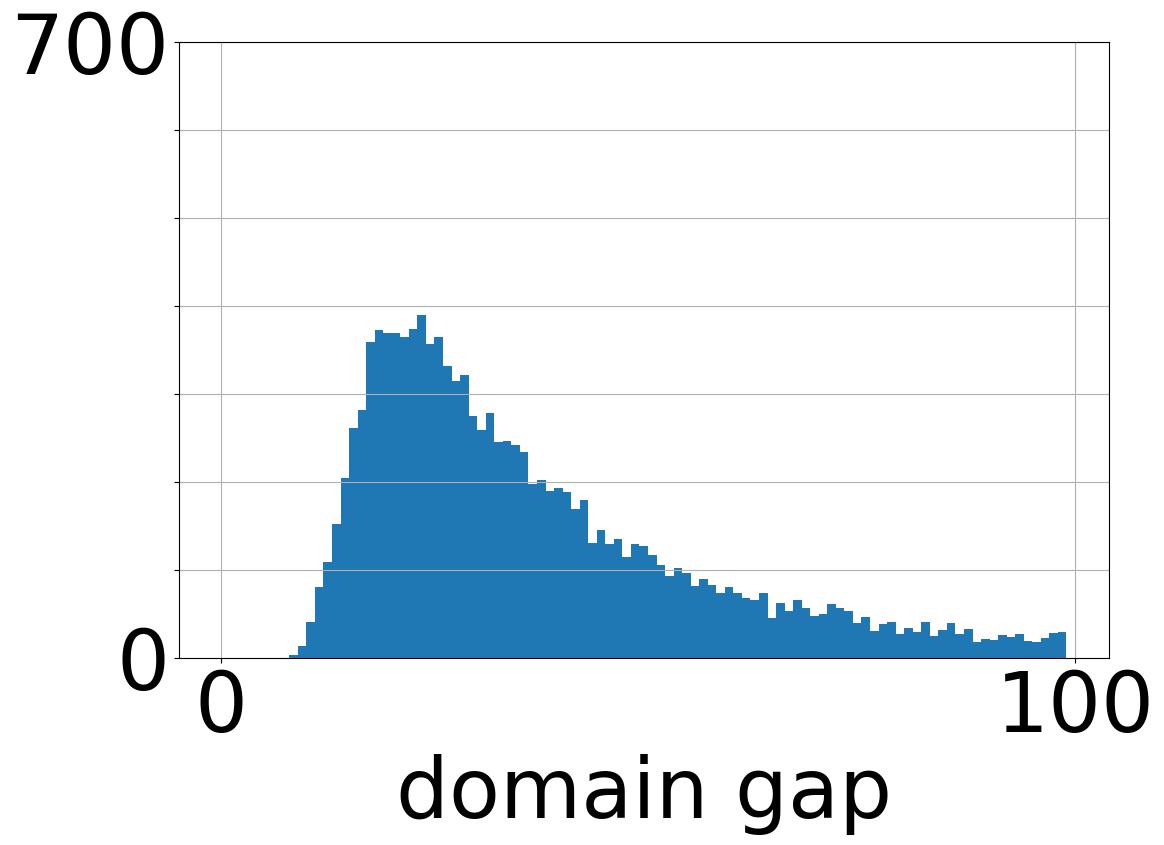} & 
\includegraphics[trim=0mm 0mm 0mm 0mm,clip,width=0.185\linewidth]{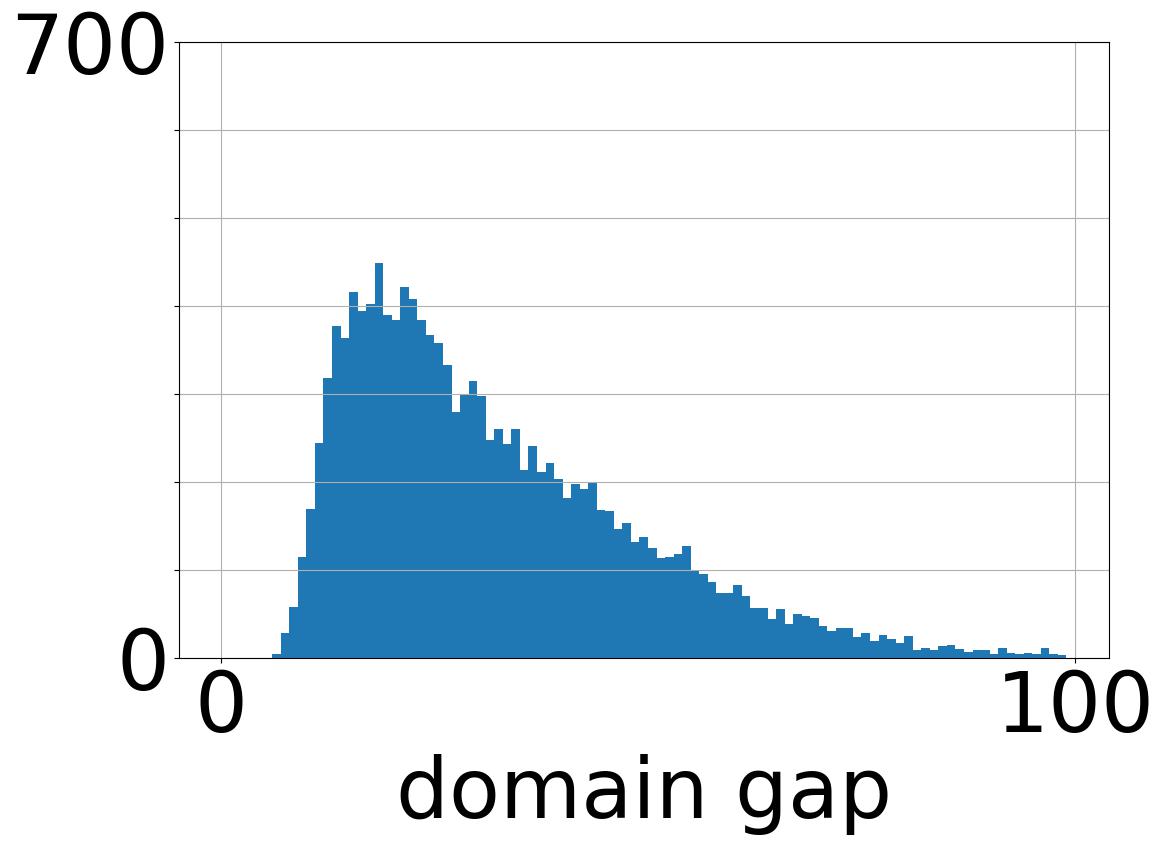}\\
\multicolumn{5}{c}{ICG, 50-img}\\
\end{tabular}
}
\caption{{\bf Distributions of virtual images with respect to the domain gap in more setups.} This figure shows the domain gap distributions of virtual images in other five experimental setups except for the setup (i.e., VisDrone, 50-img) shown in Figure 3 of the main manuscript. The $x$ axis represents the domain gap and the $y$ axis represents the corresponding number of virtual images.}
\label{fig:distr_domain_gap}
\end{figure*}

{\bf Okutama-Action} was originally created for human action detection from the aerial view. Although the main task associated with this dataset is human action detection, the dataset also includes a sub-task of pedestrian detection. To acquire images from various aerial views, a drone with an embedded camera flew freely at some altitudes between 10$m$ and 45$m$, and the camera angle was set at 45 or 90 degrees. The Okutama-Action dataset contains 43 video sequences in 4K resolution (i.e., 3840$\times$2160), where 33 video sequences are used for the \texttt{train-val} set and the remaining 10 video sequences are used for the \texttt{test} set. Since adjacent frames in the video sequence are very similar, we use every tenth frame in both the \texttt{train-val} set and the \texttt{test} set. Our model is trained on the \texttt{train-val} set and tested with the \texttt{test} set.

{\bf ICG} was collected for studying semantic understanding of urban scenes. Additionally, the ICG dataset also provides information, such as ground-truth human bounding boxes, for the human detection task. Images in the ICG dataset were captured from a camera located at some altitudes between 5-50$m$ above the ground at a resolution of 6000$\times$4000. The ICG dataset provides a \texttt{training} set of 400 images for the human detection task, of which we use the first 320 images for training and the remaining 80 images for testing.

{\bf Archangel-Synthetic} is one of the three sub-datasets included in the Archangel dataset, along with \textit{Archangel-Real} and \textit{Archangel-Mannequin}. The Archangel dataset is a hybrid UAV-based dataset captured with similar imaging conditions in real and synthetic domains. An important property of the Archangel dataset that sets it apart from other datasets is that it provides metadata about the camera positions in terms of UAV altitudes and radii of the rotation circles for each image. The Archangel-Synthetic dataset was generated by using the Unity game engine. The dataset includes eight virtual characters in three different poses captured with camera viewing angles ranging from 0$^{\circ}$ to 358$^{\circ}$ in increments of 2$^{\circ}$, UAV altitudes and rotation circle radii from 5$m$ to 80$m$ in increments of 5$m$, and four different sun angles. The total number of images in the Archangel-Synthetic dataset is 4.4M. Considering the significant difference in dataset size between the Archangel-Synthetic dataset with the other real datasets, a small subset of the entire dataset (17.6K) was used as the virtual image set in our experiments. The size of images in the Archangel-Synthetic dataset is 512$\times$512.

\section{Additional Analyses}

\noindent{\bf Training time comparison.} Training times for PTL and the baselines are shown in Tab~\ref{tab:training_time_v2}. For PTL, it is observed that the detector training time (i.e., `Dtr train' in the table) decreases as training progresses because the number of virtual images (i.e., {\it Archangel-Synthetic}) with a usually smaller image size than real images (i.e., {\it VisDrone}) increases during training. Detector training in PTL uses the same number of iterations regardless of the PTL iteration. In contrast, the CycleGAN training time (i.e., `GAN train' in the table) gradually increases due to the increased number of virtual images.


PTL is slow due to CycleGAN training, but it is still much faster than `naive merge w/ transform' (i.e., the baseline using transformation) because only a subset of virtual images is used instead of the full set for each PTL iteration. PTL can lead to scalabillity issues due to its training time when a large number of virtual images are used with longer iterations. To address this issue, we can reduce the CycleGAN training time for each PTL iteration by fine-tuning the model trained in the previous PTL iteration.\smallskip


\begin{table}[t]
\centering
\resizebox{\linewidth}{!}{%
\setlength{\tabcolsep}{5.0pt}
\renewcommand{\arraystretch}{1.1}
\begin{tabular}{l|r|l}
\multicolumn{1}{c|}{method} & \multicolumn{1}{c|}{total} & \multicolumn{1}{c}{detail} \\\hline
baseline & 40 & \\
pretrain-finetune & 21 & 4/17 (pretrain / finetune) \\
naive merge & 17 & \\
~~~~~~w/ transform &  2,777 & 2,760/17 (GAN train / Dtr train) \\\hline
{\bf PTL} & 600 & 20/iter. (domain gap calc. on virtual set) \\
&& 40/36/32/28/25/22 (Dtr train, $\sim$ $6^{th}$ iter.) \\
&& 28/41/56/69/83 (GAN train, $\sim$ $5^{th}$ iter.) \\
\end{tabular}
}
\vspace{-0.2cm}
\caption{{\bf Wall-clock training time in $mins$ (VisDrone, 20-shot).} GeForce RTX 2080 Ti GPUs are used for this comparison.}
\label{tab:training_time_v2}
\end{table}

\noindent{\bf Analysis of domain gap measurement.} To investigate the effect of using the Mahalanobis distance to measure the domain gap, we compare it to other metric available under the assumption that the representation for a certain category in a real dataset is modeled by a multivariate Gaussian distribution. Specifically, we use Euclidean distance, which depends only on the mean but not on the covariance of the distribution. As shown in Table~\ref{tab:ablation_Mdistance}, using Mahalanobis distance consistently presents better accuracy in both the in-domain and cross-domain setups.\smallskip


\begin{table}[t]
\centering
\resizebox{\linewidth}{!}{%
\setlength{\tabcolsep}{13.0pt}
\renewcommand{\arraystretch}{1.1}
\begin{tabular}{c|c||cc}
metric & Vis & Oku & ICG \\\hline
\multicolumn{1}{l|}{Euclidean} & 5.28 /1.52 & 28.80 /6.83 & 26.00 /7.06 \\
\multicolumn{1}{l|}{Mahalanobis} & {\bf 6.83} /{\bf 1.94} & {\bf 30.72} /{\bf 7.45} & {\bf 26.86} /{\bf 7.22} \\
\end{tabular}
}
\vspace{-0.2cm}
\caption{{\bf Comparison of various distance metrics (VisDrone, 20-shot)}.}
\label{tab:ablation_Mdistance}
\end{table}

\noindent{\bf Analysis of transformation candidate selection.} To investigate the effect of using weighted random sampling to select transformation candidates, we compare it with a variety of other selection strategies. We carry out experiments selecting 100 virtual images with three different ranges of domain gaps (\{close, mid, far\}) instead of using weighted random sampling (`our' in the table) for each PTL iteration in Tab~\ref{tab:ablation_domain_gap}. We observe that our selection strategy is the best in the in-domain setup without sacrificing accuracy much in the cross-domain setup.\smallskip


\begin{table}[t]
\centering
\resizebox{\linewidth}{!}{%
\setlength{\tabcolsep}{10.0pt}
\renewcommand{\arraystretch}{1.1}
\begin{tabular}{c|ccc|c}
test & close 100 & mid 100 & far 100 & our \\\hline
\multicolumn{1}{l|}{Vis} & ~~9.11/~~2.71 & ~~9.11/~~2.89 & ~~8.64/~~2.53 & ~~{\bf 9.38}/~~{\bf 2.94} \\\hline\hline
\multicolumn{1}{l|}{Oku} & 39.57/11.05 & {\bf 43.31}/{\bf 11.55} & 41.98/11.32 & 42.39/11.47 \\
\multicolumn{1}{l|}{ICG} & 29.66/~~7.40 & {\bf 34.98}/~~{\bf 8.97} & 31.24/~~8.19 & 30.01/~~7.36 \\
\end{tabular}
}
\vspace{-0.2cm}
\caption{{\bf Various transformation candidate selections (VisDrone, 50-shot).}}
\label{tab:ablation_domain_gap}
\end{table}

\noindent{\bf Further analysis of the properties of progressive learning.} In this section, we show the distributions of transformation candidates in relation to the camera position (in Figure~\ref{fig:distr_camera_location}) and the domain gap (in Figure~\ref{fig:distr_domain_gap}) for other five cases not shown in Figure 3 of the main manuscript. We intend to show that the analysis described in the main manuscript can be also applied to these cases. 

For the distributions of transformation candidates with respect to camera locations (Figure~\ref{fig:distr_camera_location}), the observation shown in the main manuscript is also applied to the other five cases. That is, as PTL progresses, the camera locations of virtual images included in the training set are gradually spread over the entire area. Therefore, the validity of the transformation candidate selection process of PTL extends to all the experimental setups considered in this paper.

Note that the distributions of humans in the real training set with respect to camera locations is likely to be similar to the distributions of transformation candidates with respect to camera locations at the first PTL iteration as a virtual image with a smaller domain gap is selected with a higher probability. Accordingly, we speculate that humans in the real training set were captured from various camera locations in the Okutama-action dataset. In contrast, in the other two datasets, most of them were taken at some similar ranges. In addition, in the ICG dataset, the camera locations where most images were taken are in the close range, which might be the reason why diversifying camera locations through PTL does not significantly improve accuracy in the cross-domain setup. 

For the distributions of virtual images with respect to the domain gap (Figure~\ref{fig:distr_domain_gap}), the observation mentioned in the main manuscript that the distribution becomes narrower and smaller as PTL progresses is still perceived in these settings. However, the speed of this distribution change is particularly slow for the ICG dataset compared to the other datasets, as shown in Figure~\ref{fig:distr_domain_gap}. This observation also implies that the ICG dataset has very different characteristics compared to the other two datasets.\smallskip

\begin{figure*}[t]
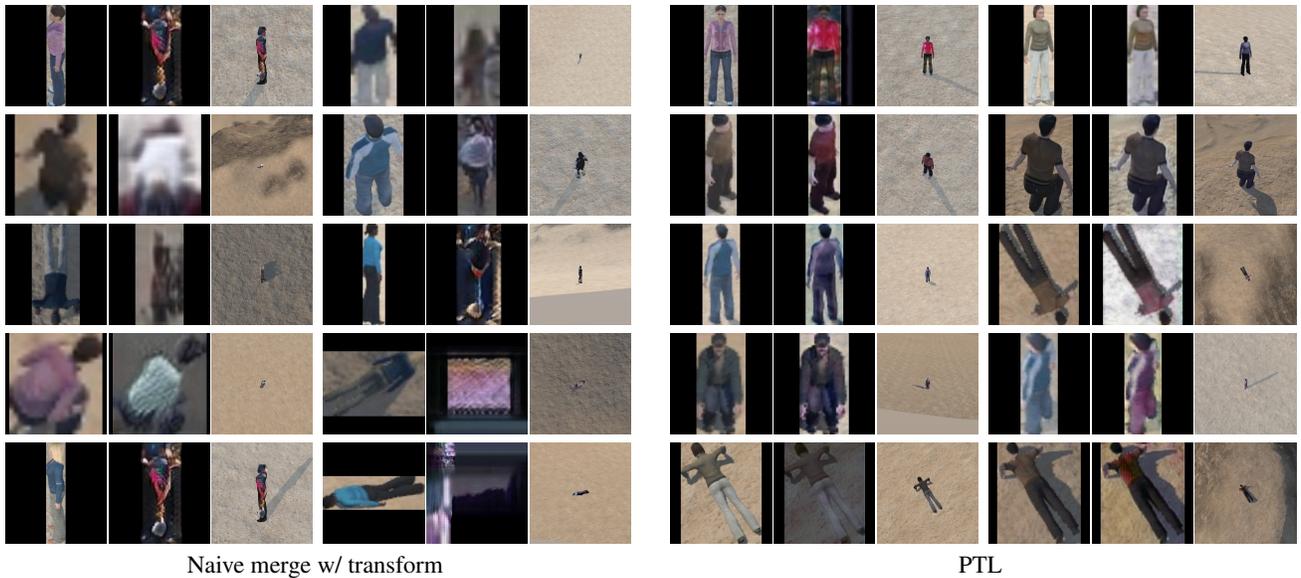

\centering
\resizebox{\linewidth}{!}{%
\setlength{\tabcolsep}{3.0pt}

}
\vspace{-0.4cm}
\caption{{\bf Sample Virtual2Real transformation output (VisDrone, 20-shot).} Each set consists of three images: original virtual image (left), transformed image (middle), and transformed image with background (right).}
\label{fig:transformation_samples_Vis_20}
\end{figure*}
\begin{figure*}[t]
\centering
\resizebox{\linewidth}{!}{%
\setlength{\tabcolsep}{3.0pt}
%
}
\vspace{-0.4cm}
\caption{{\bf Sample Virtual2Real transformation output (VisDrone, 50-shot).} Each set consists of three images: original virtual image (left), transformed image (middle), and transformed image with background (right).}
\label{fig:transformation_samples_Vis_50}
\end{figure*}
\begin{figure*}[t]
\centering
\resizebox{\linewidth}{!}{%
\setlength{\tabcolsep}{3.0pt}
%
}
\vspace{-0.4cm}
\caption{{\bf Sample Virtual2Real transformation output (Okutama-Action, 20-shot).} Each set consists of three images: original virtual image (left), transformed image (middle), and transformed image with background (right).}
\label{fig:transformation_samples_Oku_20}
\end{figure*}
\begin{figure*}[t]
\centering
\resizebox{\linewidth}{!}{%
\setlength{\tabcolsep}{3.0pt}
%
}
\vspace{-0.4cm}
\caption{{\bf Sample Virtual2Real transformation output (Okutama-Action, 50-shot).} Each set consists of three images: original virtual image (left), transformed image (middle), and transformed image with background (right).}
\label{fig:transformation_samples_Oku_50}
\end{figure*}
\begin{figure*}[t]
\centering
\resizebox{\linewidth}{!}{%
\setlength{\tabcolsep}{3.0pt}
%
}
\vspace{-0.4cm}
\caption{{\bf Sample Virtual2Real transformation output (ICG, 20-shot).} Each set consists of three images: original virtual image (left), transformed image (middle), and transformed image with background (right).}
\label{fig:transformation_samples_Icg_20}
\end{figure*}
\begin{figure*}[t]
\centering
\resizebox{\linewidth}{!}{%
\setlength{\tabcolsep}{3.0pt}
%
}
\\
\multicolumn{2}{c}{Naive merge w/ transform} &&&
\multicolumn{2}{c}{PTL} \\
\end{tabular}
}
\vspace{-0.4cm}
\caption{{\bf Sample Virtual2Real transformation output (ICG, 50-shot).} Each set consists of three images: original virtual image (left), transformed image (middle), and transformed image with background (right).}
\label{fig:transformation_samples_Icg_50}
\end{figure*}

\noindent{\bf Qualitative analysis of transformation.} Figures~\ref{fig:transformation_samples_Vis_20},~\ref{fig:transformation_samples_Vis_50},~\ref{fig:transformation_samples_Oku_20},~\ref{fig:transformation_samples_Oku_50},~\ref{fig:transformation_samples_Icg_20}, and~\ref{fig:transformation_samples_Icg_50} show several samples of transformed virtual images included in the training set using methods with virtual2real transformation (i.e., PTL and `naive merge w/ transformation') in the six experimental setups. Since the trends seen in these examples are similar to Figure 5 in the main manuscript, a qualitative analysis of the superiority of PTL over `naive merge w/ transform' and our claim that the domain gap between virtual images and real images should be considered when training the virtual2real transformation generator can also be applied to these experimental setups.


{\small
\bibliographystyle{Styles/ieee_fullname}
\bibliography{References/egbib}
}

\end{document}